\documentclass[numbers]{article}   

\usepackage[final]{neurips_2025}

\usepackage[utf8]{inputenc} 
\usepackage[T1]{fontenc}    
\usepackage{hyperref}       
\usepackage{url}            
\usepackage{booktabs}       
\usepackage{amsfonts}       
\usepackage{nicefrac}       
\usepackage{microtype}      
\usepackage{xcolor}         

\usepackage{enumitem}
\usepackage{amsmath}

\usepackage{mathtools}
\usepackage{array}
\newcolumntype{C}[1]{>{\centering\arraybackslash}p{#1}}

\usepackage{amsmath}
\usepackage{dsfont}
\usepackage{multirow}
\usepackage{textcomp}
\usepackage{siunitx}

\usepackage{tabularx}
\usepackage{makecell}
\usepackage{booktabs}

\usepackage[slc=off, subrefformat=parens]{subcaption}
\ifx\subcaptionstyle\subcaptionbelow
\else
\fi

\AtEndPreamble{
    \usepackage[capitalize]{cleveref}
    \crefname{section}{Sec.}{Secs.}
    \Crefname{section}{Section}{Sections}
    \crefname{table}{Tab.}{Tabs.}
    \Crefname{table}{Table}{Tables}
    \crefname{figure}{Fig.}{Figs.}
    \Crefname{figure}{Figure}{Figures}
}

\usepackage{glossaries}
\newacronym{RS}{RS}{remote sensing}
\newacronym{CV}{CV}{computer vision}
\newacronym{MI}{MI}{medical imaging}
\newacronym{CNNs}{CNNs}{Convolutional neural networks}
\newacronym{ViTs}{ViTs}{Vision transformers}
\newacronym{VLMs}{VLMs}{vision-language models}

\glsdisablehyper

\usepackage{makecell}

\title{ImageNet-trained CNNs are not biased towards texture: Revisiting feature reliance through controlled suppression}

\author{Tom Burgert\textsuperscript{1,2}, Oliver Stoll\textsuperscript{1,2}, Paolo Rota\textsuperscript{3}, Begüm Demir\textsuperscript{1,2} \\
BIFOLD\textsuperscript{1}, TU Berlin\textsuperscript{2},  University of Trento\textsuperscript{3} \\
{\tt\small \{t.burgert,o.stoll,demir\}@tu-berlin.de}, {\tt\small paolo.rota@unitn.it}
}

\begin{document}

\maketitle

\begin{abstract}
The hypothesis that Convolutional Neural Networks (CNNs) are inherently texture-biased has shaped much of the discourse on feature use in deep learning. We revisit this hypothesis by examining limitations in the cue-conflict experiment by Geirhos et al. To address these limitations, we propose a domain-agnostic framework that quantifies feature reliance through systematic suppression of shape, texture, and color cues, avoiding the confounds of forced-choice conflicts. By evaluating humans and neural networks under controlled suppression conditions, we find that CNNs are not inherently texture-biased but predominantly rely on local shape features. Nonetheless, this reliance can be substantially mitigated through modern training strategies or architectures (ConvNeXt, ViTs). We further extend the analysis across computer vision, medical imaging, and remote sensing, revealing that reliance patterns differ systematically: computer vision models prioritize shape, medical imaging models emphasize color, and remote sensing models exhibit a stronger reliance on texture. Code is available at \mbox{https://github.com/tomburgert/feature-reliance}.
\end{abstract}

\maketitle  
\section{Introduction}\label{sec:intro}

\gls{CNNs} have played a central role in the development of deep learning models for visual recognition \cite{krizhevsky_imagenet_2012}, \cite{he_deep_2016}, \cite{tan_efficientnet_2019}, \cite{liu_convnet_2022}. Their success across a range of \gls{CV} benchmarks has contributed to the perception that they acquire perceptual representations resembling those of humans \cite{lecun_deep_2015}, \cite{kubilius_deep_2016}, \cite{ritter_cognitive_2017}. However, a growing body of work suggests that \gls{CNNs} may process visual information in fundamentally different ways \cite{geirhos_imagenet-trained_2019}, \cite{baker_deep_2018}, \cite{brendel_approximating_2019}. One of the most influential claims in this direction is that \gls{CNNs} trained on ImageNet are inherently biased towards texture \cite{geirhos_imagenet-trained_2019}, in contrast to humans who predominantly rely on shape cues \cite{landau_importance_1988}. This claim, first formalized by Geirhos et al. \cite{geirhos_imagenet-trained_2019} through their cue-conflict experiment, has since shaped much of the discourse on how to evaluate and interpret the use of features in deep neural networks. 

In the cue-conflict experiment, images are synthesized by combining the shape of one object class with the texture of another, using neural style transfer techniques \cite{gatys_image_2016}. Models and humans are then presented with these hybrid images, and their predictions are analyzed to infer which visual cues they rely on. The observed divergence, with \gls{CNNs} favoring texture and humans favoring shape, has become a dominant narrative for understanding human–machine perceptual differences and has inspired a wide range of follow-up studies \cite{hermann_origins_2020}, \cite{hermann_what_2020}, \cite{islam_shape_2021}, \cite{gavrikov_can_2024}, \cite{jain_combining_2022}.

Although influential, the cue-conflict experiment is based on assumptions that may limit the generalizability and clarity of its findings. Conceptually, it reduces feature reliance to a binary choice between shape and texture, overlooking other potentially informative cues such as color, and tends to link salience with reliance implicitly. Methodologically, the generated stimuli entangle unintentionally multiple features, introduce texture cues across the image in a spatially unbalanced manner, and rely on shape-based response interfaces that may bias human judgments. As discussed further in \Cref{sec:rethinking_texture_bias}, these conceptual and methodological limitations complicate conclusions about the feature use of models and humans.

\begin{figure*}[t!]
    \centering
    \includegraphics[width=1.0\linewidth]{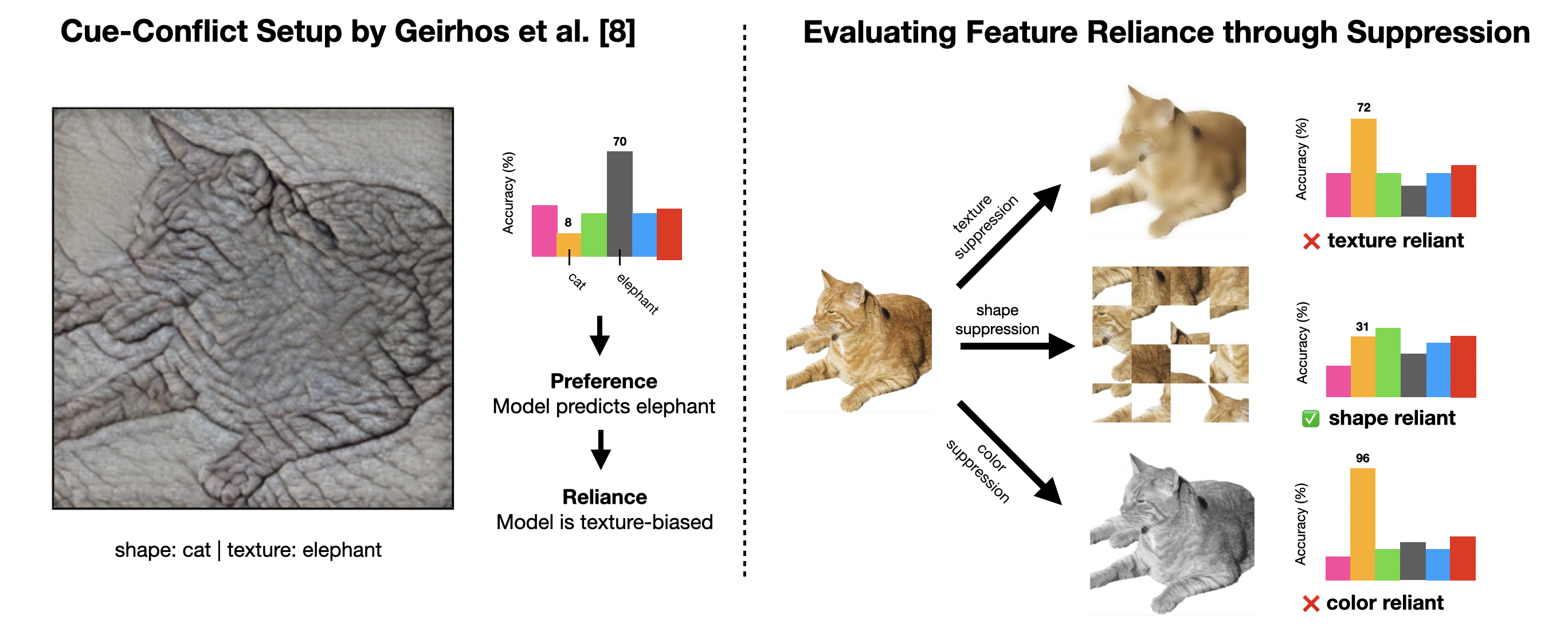}
    \caption{Comparison of cue-conflict setup \cite{geirhos_imagenet-trained_2019} (left) and our suppression-based framework (right). While Geirhos et al. infer reliance through preference on hybrid images, our framework directly quantifies reliance by measuring accuracy under systematic suppression of texture, shape, or color.}
    \label{fig:method_overview}
\end{figure*}

In this work, we argue for a conceptual shift: from analyzing feature bias through forced-choice conflicts to assessing feature reliance through targeted suppression. This conceptual distinction reframes how feature preferences and reliance should be evaluated. A model may prefer a certain cue in conflict, not because it is more predictive, but because it is more salient. Conversely, a model may rely heavily on a feature in natural settings, even if it does not dominate in cue-conflict scenarios. To address the aforementioned limitations, we propose a new domain-agnostic evaluation framework that quantifies performance degradation under systematic suppression of individual feature types (e.g., shape, texture, and color), enabling empirical measurement of reliance. The proposed framework does not rely on adversarial inputs or neural style transfer, but instead uses direct feature-suppressing transformations. By isolating individual feature contributions, our framework offers a more reliable basis for interpreting model decisions and comparing representational strategies, both between humans and neural networks, and across model architectures and domains.

Our main contributions are as follows:

\begin{enumerate}[label=(\arabic*)]
    \item We present a re-examination of Geirhos et al.’s cue-conflict experiment \cite{geirhos_imagenet-trained_2019}, highlighting aspects in their evaluation protocol that may limit its generalizability.
    \item We introduce a domain-agnostic framework for evaluating feature reliance through targeted feature suppression, enabling cleaner measurement of model dependence on individual visual cues without requiring conflicting cue setups.
    \item Using the proposed framework, we systematically compare human and model feature reliance under controlled conditions. Our results challenge the texture bias hypothesis \cite{geirhos_imagenet-trained_2019} by showing that \gls{CNNs} are not inherently texture-biased; instead, they only exhibit a pronounced sensitivity to local shape, which can be mitigated through modern training strategies. Notably, models trained with vision-language supervision most closely match human behavior.
    \item We apply the same framework to assess domain-specific differences in feature reliance, showing that models trained on \gls{CV}, \gls{RS}, and \gls{MI} datasets prioritize distinct visual cues depending on domain characteristics.
\end{enumerate}

\section{Related Work}\label{sec:related_work}

Understanding which features deep neural networks rely on for image classification has been a long-standing research question. While early interpretations of \gls{CNNs} assumed a hierarchical buildup from low-level edges to complex shape representations \cite{lecun_deep_2015}, \cite{kubilius_deep_2016}, \cite{ritter_cognitive_2017} more recent studies have challenged this view, suggesting that \gls{CNNs} often rely disproportionately on local texture rather than global shape \cite{baker_deep_2018}, \cite{brendel_approximating_2019}, \cite{geirhos_imagenet-trained_2019}, \cite{subramanian_spatial-frequency_2023}. Geirhos et al. \cite{geirhos_imagenet-trained_2019} formalized this observation as the texture bias hypothesis, using a cue-conflict protocol to reveal divergent feature preferences between humans and \gls{CNNs}.

Subsequent work investigated factors shaping feature reliance beyond architecture. Hermann et al. \cite{hermann_origins_2020} showed that texture bias in \gls{CNNs} arises primarily from training objectives and augmentations, with techniques like blurring and cropping increasing shape bias more than architectural changes. Although shape features are present in deeper layers \cite{islam_shape_2021}, \cite{hermann_what_2020}, they are not consistently used during classification. Transformer-based models and vision-language models have shifted this discussion. \gls{ViTs} exhibit lower texture bias due to their global attention mechanism \cite{naseer_intriguing_2021}, \cite{tuli_are_2021}, and vision-language models show improved alignment with human-like shape use \cite{gavrikov_can_2025}.

Various methods have attempted to enforce shape bias or suppress texture cues for improved robustness, including anisotropic filtering \cite{mishra_learning_2022}, edge encoding \cite{nazari_role_2022}, style disentanglement \cite{nam_reducing_2021}, \cite{kashyap_towards_2022}, and shape-focused augmentations \cite{lee_improving_2022}, \cite{tripathi_edges_2023}. However, stylization alone may improve robustness independent of shape bias \cite{mummadi_does_2021}, and neither shape nor texture bias reliably predicts generalization \cite{gavrikov_can_2024}. These findings have motivated integrative approaches that combine diverse feature biases. Joint supervision \cite{li_shape-texture_2021}, ensembles \cite{co_universal_2021}, and adaptive recombination \cite{qiu_shape-biased_2024} aim to harness complementary features. Ge et al. \cite{ge_contributions_2022} and Jain et al. \cite{jain_combining_2022} show that disentangling and combining shape, texture, and color improve robustness and interpretability. Nonetheless, Lucieri et al. \cite{lucieri_revisiting_2022} caution that in domains like \gls{MI}, cue entanglement is essential and biasing towards shape may be counterproductive.

Efforts to increase shape bias are often motivated by the broader goal of human-model alignment. Geirhos et al. \cite{geirhos_generalisation_2018}, \cite{geirhos_partial_2021} show that even robust models exhibit error patterns that diverge from humans, revealing a persistent consistency gap. Muttenthaler et al. \cite{muttenthaler_human_2023} further argue that alignment with human conceptual structure depends more on training signals than model scale, indicating that robustness and shape bias alone are insufficient proxies for human-like perception.

\section{Rethinking Texture Bias: A Critical Look at Cue-Conflict Evaluation}\label{sec:rethinking_texture_bias}

\begin{figure*}[h]
    \def\subfigwidth{.29}
    \centering
    \begin{subfigure}{\subfigwidth\linewidth}
        {\includegraphics[width=\linewidth]{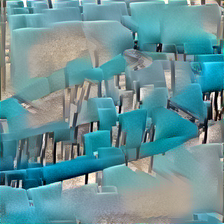}}
        {\caption{}\label{subfigure:limitations_a}}
    \end{subfigure}
    \hfill
    \begin{subfigure}{\subfigwidth\linewidth}
        {\includegraphics[width=\linewidth]{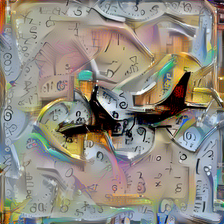}}
        {\caption{}\label{subfigure:limitations_b}}
    \end{subfigure}
    \hfill
    \begin{subfigure}{\subfigwidth\linewidth}
        {\includegraphics[width=\linewidth]{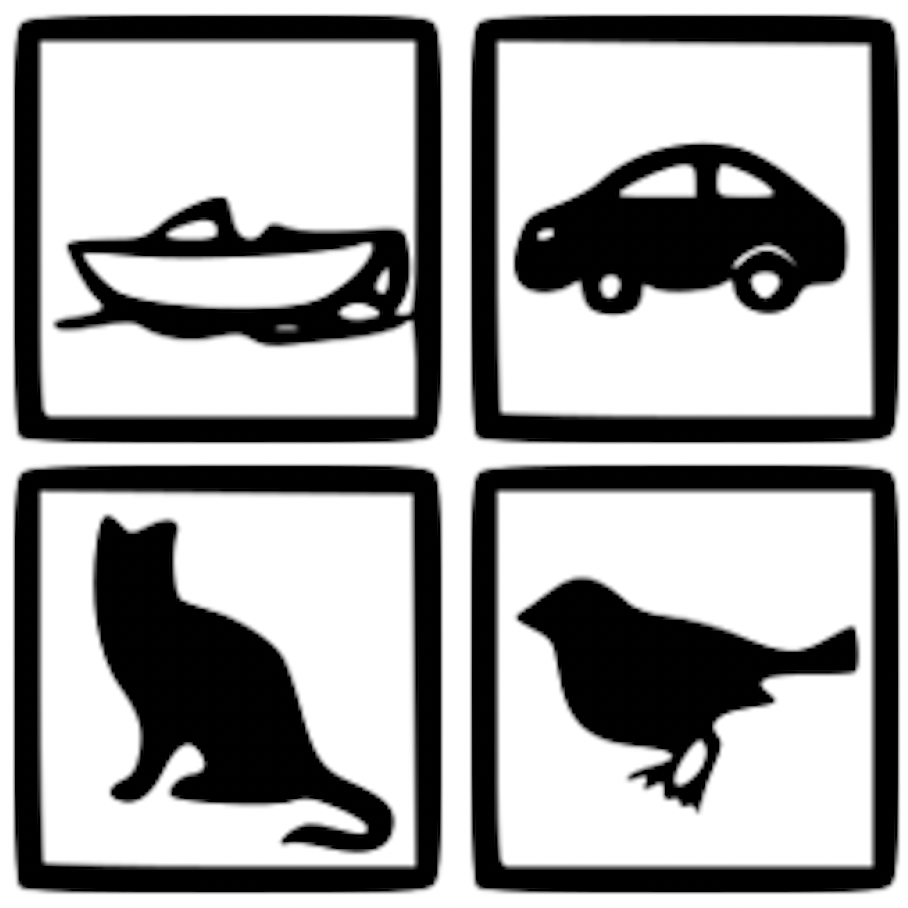}}
        {\caption{}\label{subfigure:limitations_c}}
    \end{subfigure}
    \label{fig:limitations_cue-conflict}
    \caption{
    Example images taken from the cue-conflict dataset \cite{geirhos_imagenet-trained_2019}. \textbf{(a)} Boat shape cue merged with chair texture cue. \textbf{(b)} Airplane shape cue merged with clock texture cue. \textbf{(c)} Icons of the human interface to select classes.
    }
\end{figure*}

The hypothesis that \gls{CNNs} trained on ImageNet are biased towards texture was popularized by Geirhos et al. \cite{geirhos_imagenet-trained_2019}, who introduced a cue-conflict evaluation protocol. In this protocol, images were generated by neural style transfer \cite{gatys_image_2016}, combining the shape content (cue) of one class with the texture content (cue) of another. Predictions from both humans and \gls{CNNs} on these images were then used to infer whether classification decisions were driven more by shape or texture features. Over time, the cue-conflict evaluation protocol has become a de facto standard for assessing feature bias in deep neural networks. While impactful, this protocol introduced several assumptions and limitations that have received limited attention. Conceptually, the protocol frames feature reliance as a binary shape-or-texture choice, which may overlook other cues such as color and conflates preference with dependence. In addition, the stylized stimuli constrain the evaluation of feature bias to naturalistic images with a similar set of classes and cannot be generalized across datasets (e.g., flower classification) or domains (e.g., \gls{RS}, \gls{MI}). Beyond these conceptual limitations, the cue-conflict protocol exhibits three methodological concerns in its design and implementation:

\begin{enumerate}[label=(\roman*)]
    \item \textbf{Lack of Feature Isolation.} The texture cues within the cue-conflict images also preserved information beyond texture, including color and local shape structures (e.g., contours and parts of silhouettes). As a result, the synthesized texture cue was not a pure representation of texture but a composite of multiple features, making it difficult to attribute classification behavior to texture alone. An example can be seen in \Cref{subfigure:limitations_a}.
    \item \textbf{Overloaded Texture Class Signals.} The protocol consistently inserted texture cues not only into the object region but also into the image background. Since \gls{CNNs} aggregate local statistics across spatial positions, this broad spatial distribution increases the signal strength of the texture class relative to the shape class. This spatial imbalance systematically biases \gls{CNNs} towards texture-based decisions, not because of an intrinsic preference but due to the dominant spatial availability of the texture signal. An example can be seen in  \Cref{subfigure:limitations_b}.
    \item \textbf{Human Interface Bias Towards Shape.} Participants in the human experiments selected the image class by clicking on buttons labeled with icons representing each category. These icons represented global shape characteristics (i.e., silhouettes), potentially guiding participants towards matching shape features in the cue-conflict image with the icon. This response format potentially introduces bias towards shape decisions, especially when participants were unsure which feature to prioritize. The used icons are visualized in  \Cref{subfigure:limitations_c}.
\end{enumerate}

These design choices may inadvertently influence \gls{CNNs} towards texture-driven decisions and humans towards shape-based decisions and complicate drawing definitive conclusions about the actual feature reliance of models.

\section{A Domain-Agnostic Framework for Feature Reliance}\label{sec:feature_reliance_protocol}

Accurately assessing how deep neural networks rely on different visual features remains a central challenge in understanding their behavior. While the cue-conflict evaluation protocol \cite{geirhos_imagenet-trained_2019} introduced a reliance test based on feature bias, it imposes conceptual and methodological constraints that limit its generalizability. Rather than forcing models to choose between shape and texture, we propose to assess their reliance on individual feature types by systematically suppressing them and measuring the resulting impact on classification performance. This shift enables a more flexible, generalizable, and semantically grounded analysis of feature use in neural networks.

To evaluate the reliance of deep neural networks on individual visual features, we employ a set of image transformations that selectively suppress shape, texture, or color information while minimally affecting the remaining features. Each transformation is chosen for its ability to target a specific feature class. We define three feature types:

\begin{itemize}
    \item \textbf{Shape} refers to information carried by spatial arrangement and structural contours, including both global (object outline) and local (part-level) shape.
    \item \textbf{Texture} is defined by repetitive patterns, high-frequency local variations, and fine-grained surface details.
    \item \textbf{Color} denotes chromatic information independent of spatial layout or texture.
\end{itemize}

For each feature type, we include two complementary transformations that differ in their suppression mechanisms and preservation profiles, offering distinct but comparable perspectives on the targeted feature. The transformations are summarized in Table 1 and briefly described in the following. Patch Shuffle \cite{luo_defective_2019}, \cite{mummadi_does_2021} and Patch Rotation disrupt shape by modifying non-overlapping image patches: Shuffle randomizes spatial positions, while Rotation preserves locality of patches but breaks edge continuity. Both affect global or local shape, depending on the grid size. Bilateral Filtering \cite{tomasi_bilateral_1998} and Gaussian Blur reduce texture by smoothing high-frequency details, with the former preserving edges more effectively. Grayscale removes chromatic cues entirely, while Channel Shuffle disrupts color correlations without altering intensity. In the following, we validate the suppression effects of these transformations using quantitative metrics.

\begin{table}[h]
\small
\centering
\caption{Feature suppression transformations used in this work. Each feature is suppressed using two transformations with differing strengths.}
\vspace{8pt}
\begin{tabular}{lll}
\toprule
\textbf{Feature Type} & \textbf{Transformation 1} & \textbf{Transformation 2} \\
\midrule
Shape & Patch Shuffle & Patch Rotation \\
Texture & Bilateral Filter & Gaussian Blur \\
Color & Grayscale & Channel Shuffle \\
\bottomrule
\end{tabular}
\label{tab:transforms}
\end{table}

\subsection{Quantitative Validation of Suppression Transformations}\label{subsec:quantitative_justification_suppression}

While the individual transformations used in this work are not novel, their selection for targeted feature suppression requires empirical justification. To validate that each transformation suppresses the intended visual feature (e.g., texture, shape) while preserving others, we quantify their effects using four metrics: Local Variance (LV) \cite{haralick_textural_1973} and High-Frequency Energy (HFE) \cite{gonzales_digital_1987} to assess texture suppression, and Edge-SSIM (ESSIM) \cite{wang_image_2004} and Gradient Correlation (GC) to measure shape preservation. All metrics are normalized to the range $[0,1]$ by dividing by the scores of the unsuppressed (i.e., original) image. Higher values of ESSIM and GC indicate better preservation of edge and structural information, while lower values of LV and HFE reflect stronger suppression of texture features. Further, we compute a harmonic mean across the two texture metrics (Texture) and the two shape metrics (Shape) for each transformation.

We test the effectiveness of the feature suppression transformations across 800 sampled images from the ImageNet validation set. For each transformation, we evaluate a representative parameter setting chosen to balance suppression of the target feature and preservation of others. The respective parameters, such as kernel size or smoothing strength, are indexed by Param IDs in \Cref{tab:feature_suppression_metrics}, with details listed below the table. A full ablation of different parameter settings is provided in the supplemental material (see \Cref{sec:ablation_supp_effects}). In addition to our selected texture suppression transformations, we also compare common alternatives such as Non-Local Means Denoising \cite{buades_non-local_2005}, Box blur, and Median filtering \cite{huang_fast_1979} to ensure a fair comparison across standard smoothing techniques. Among texture-suppressing methods, bilateral filtering yields the most balanced trade-off between reducing texture (LV: 0.54, HFE: 0.49) and preserving shape (ESSIM: 0.74, GC: 0.85). Gaussian Blur suppresses texture more uniformly but leads to a greater loss of shape information. Box blur and median filtering remove texture strongly, but at a substantial cost to shape preservation. For shape suppression, we evaluate Patch Shuffle and Patch Rotation with a grid size of 6. These transformations preserve texture but substantially disrupt structural contours, making them suitable for assessing shape reliance. To complement the quantitative evaluation, qualitative visual examples of the suppression effects are provided in the supplemental material (see \Cref{sec:visual_examples}).

\newcolumntype{Y}{>{\centering\arraybackslash}X} 

\begin{table}[t]
\small
\centering
\caption{Quantitative validation of suppression transformations across 800 images of ImageNet. Each transformation is used with a fixed parameter setting (see Param ID legend below). Values report normalized metric scores. Arrows indicate desired direction: ↑ higher is better, ↓ lower is better.}
\label{tab:feature_suppression_metrics}
\vspace{6pt}
\renewcommand{\arraystretch}{1.2}
\setlength{\tabcolsep}{3pt}
\begin{tabularx}{\linewidth}{l c *{2}{>{\centering\arraybackslash}X}|*{4}{>{\centering\arraybackslash}X}}
\toprule
\textbf{Transformation} & \textbf{Param ID} & \textbf{Texture↓} & \textbf{Shape↑} &
\textbf{LV↓} & \textbf{HFE↓} & \textbf{ESSIM↑} & \textbf{GC↑} \\
\midrule
\multicolumn{8}{l}{\textit{Texture-Suppressing}} \\
Bilateral Filter & A & 0.521 & 0.796 & 0.548 & 0.493 & 0.737 & 0.855 \\
Box Blur         & B & 0.193 & 0.363 & 0.237 & 0.148 & 0.436 & 0.289 \\
Gaussian Blur    & C & 0.349 & 0.662 & 0.392 & 0.306 & 0.744 & 0.579 \\
Median Filter    & D & 0.357 & 0.506 & 0.399 & 0.316 & 0.584 & 0.429 \\
NLMeans Denoising & E & 0.706 & 0.797 & 0.723 & 0.690 & 0.730 & 0.864 \\
\midrule
\textbf{Transformation} & \textbf{Param ID} & \textbf{Texture↑} & \textbf{Shape↓} &
\textbf{LV↑} & \textbf{HFE↑} & \textbf{ESSIM↓} & \textbf{GC↓} \\
\midrule
\multicolumn{8}{l}{\textit{Shape-Suppressing}} \\
Patch Shuffle     & F & 1.000 & 0.176 & 1.000 & 1.000 & 0.205 & 0.147 \\
Patch Rotation    & F & 1.000 & 0.293 & 1.000 & 1.000 & 0.339 & 0.247 \\
\bottomrule
\end{tabularx}

\vspace{4pt}
\begin{minipage}{\linewidth}
\footnotesize
\textbf{Legend:}  
A: $d$=11, $\sigma_c$=170, $\sigma_s$=75;  
B: $k$=11;  
C: $k$=11, $\sigma$=2.0;  
D: $k$=11;  
E: $h$=20, $tws$=11, $sws$=11;  
F: grid=6.
\end{minipage}
\end{table}

\section{Experiments}\label{sec:human_vs_cnns}

\subsection{Experiment I: Human vs. CNNs Feature Reliance}

\textbf{Experimental Setup}. To compare human and model reliance on different visual features, we designed a controlled experiment inspired by Geirhos et al. \cite{geirhos_generalisation_2018}, \cite{geirhos_imagenet-trained_2019}. We constructed an ImageNet16-like dataset by selecting 50 representative images for each of 16 entry-level categories derived from the WordNet hierarchy \cite{miller_wordnet_1995} (see \cite{geirhos_generalisation_2018} for details). Images were selected based on the most confidently predicted samples in the ImageNet validation set \cite{deng_imagenet_2009} by a ResNet50 \cite{he_deep_2016} pretrained on ImageNet1k, ensuring balanced subclass coverage. For categories with insufficient confident predictions (airplane, knife, oven), additional samples were manually added. All images were resized to \( 224 \times 224 \) pixels.

Humans were presented with image stimuli in randomized order under one of five conditions: original, global shape suppression, local shape suppression, texture suppression, or color suppression. Each feature was suppressed via a single transformation with fixed hyperparameters: Patch Shuffle with grid size 3 (global shape), grid size 6 (local shape), bilateral filtering with $d$=12, $\sigma_{\text{color}}$=170 and $\sigma_{\text{space}}$=75 (texture), and grayscale conversion (color). See \Cref{subsec:quantitative_justification_suppression} for justification. Each participant saw only one randomly chosen version of each image to avoid learning effects. The five suppression conditions of one image were split across groups of five participants to ensure balanced coverage. Twenty participants completed the study. Following Geirhos et al.\cite{geirhos_imagenet-trained_2019}, each trial included a \SI{300}{\milli\second} fixation square, \SI{200}{\milli\second} image presentation, and \SI{200}{\milli\second} pink noise mask (1/f spectral shape) to minimize feedback processing. Participants selected one of 16 categories via a \( 4 \times 4 \) grid of alphabetically sorted class names. An additional “not clear” button was available for unrecognizable stimuli. Attention checks were administered every 100 trials, and failed trials were excluded. Additional details and interface screenshots can be found in the supplemental material.

Model evaluation mirrored the human protocol, evaluating their performance under the same five suppression conditions using the identical image set shown to humans. For each image, the class prediction was computed by summing softmax outputs over all ImageNet subclasses mapping to the same entry-level category. Only predictions above the threshold of 0.5 were considered correct. This procedure was chosen heuristically, complementary results using argmax to define class predictions are reported in the supplemental material and show nearly identical reliance profiles. 

We evaluated several architectures: ResNet50-standard, trained from scratch with basic augmentations, and ResNet50-sota, trained with a modern recipe \cite{wightman_resnet_2021}. Additional CNNs include MobileNetV3 \cite{howard_searching_2019}, EfficientNet \cite{tan_efficientnet_2019}, EfficientNetV2 \cite{tan_efficientnetv2_2021}, ConvMixer \cite{trockman_patches_2023}, ConvNeXt \cite{liu_convnet_2022}, and ConvNeXtV2 \cite{woo_convnext_2023}. Transformer-based models include ViT \cite{dosovitskiy_image_2021}, DeiT  \cite{touvron_training_2021}, SwinTransformer \cite{liu_swin_2021}, and CLIP ViT \cite{radford_learning_2021}. All models except ResNet50-standard were obtained as pretrained checkpoints from the \texttt{timm} library \cite{wightman_pytorch_2019}. The detailed training procedures can be found in the supplemental material.

\textbf{Results.} \Cref{fig:humans_vs_cnns} presents a comparative overview of the performance of humans and \gls{CNNs} under feature suppression, plotted as the relative accuracy (i.e., accuracy under suppression divided by baseline accuracy on original images). Separate subplots show results for each suppressed feature type. We highlight three representative \gls{CNNs}: ResNet50-standard, ResNet50-sota, and ConvNeXtV2 alongside human performance. The results show that \gls{CNNs} are not strongly reliant on texture: under texture suppression, ResNet50-standard retains 80\% of its original performance, close to performance under global shape suppression (83\%). The highest vulnerability is observed under local shape suppression, where accuracy drops to just 28\%. Humans exhibit a similar reliance profile with local shape suppression being most disruptive, but show higher robustness to it (76\% retained accuracy). Interestingly, modern training strategies substantially mitigate this effect: the ResNet50-sota reaches 62\% under local shape suppression, and ConvNeXtV2 improves further to 65\%. These results suggest that the heavy reliance on local shape observed in earlier \gls{CNNs} is not architectural in nature but can be alleviated through better training regimes. A likely contributing factor is the inclusion of stronger regularization, improved data augmentations, and more extensive training schedules in the modern setup, which may encourage broader feature utilization beyond local patterns. Statistical significance tests confirming these differences are reported in the supplemental material.

Broadening the analysis to a wider range of architectures (\Cref{tab:feature_reliance}), we observe that several models trained with state-of-the-art recipes exhibit a more balanced reliance profile. However, this trend is not universal: ConvMixer, EfficientNet, and MobileNet variants retain a strong dependence on local shape, indicating that improved training alone does not guarantee human-like feature use and that architectural inductive biases or capacity limitations may still play a role. Among transformer-based models, the ViT demonstrates a feature reliance profile similar to ResNet50-sota across all suppression conditions, challenging the notion that transformers are inherently more shape-oriented than \gls{CNNs}. Notably, the CLIP VIT model most closely matches human performance across all feature suppression conditions, suggesting that vision-language supervision encourages more human-aligned representations. This may reflect the effect of contrastive vision-language training, which prioritizes alignment with high-level semantic concepts over low-level visual cues.

These findings challenge the texture bias hypothesis popularized by Geirhos et al. \cite{geirhos_imagenet-trained_2019} as a fixed inductive bias of \gls{CNNs}. Instead, the observed behavior in the cue-conflict experiment may have reflected a dominant reliance on local shape features, rather than an inherent texture bias. 

\begin{figure}[t!]
    \def\subfigwidth{.23}
    \centering

    \makebox[\linewidth][c]{%
        \begin{minipage}{\subfigwidth\linewidth}
            \centering\scriptsize\textbf{(a) Global Shape}
        \end{minipage}
        \begin{minipage}{\subfigwidth\linewidth}
            \centering\scriptsize\textbf{(b) Local Shape}
        \end{minipage}
        \begin{minipage}{\subfigwidth\linewidth}
            \centering\scriptsize\textbf{(c) Texture}
        \end{minipage}
        \begin{minipage}{\subfigwidth\linewidth}
            \centering\scriptsize\textbf{(d) Color}
        \end{minipage}
    }

    \vspace{2pt}

    \begin{subfigure}{\subfigwidth\linewidth}
        \includegraphics[width=\linewidth]{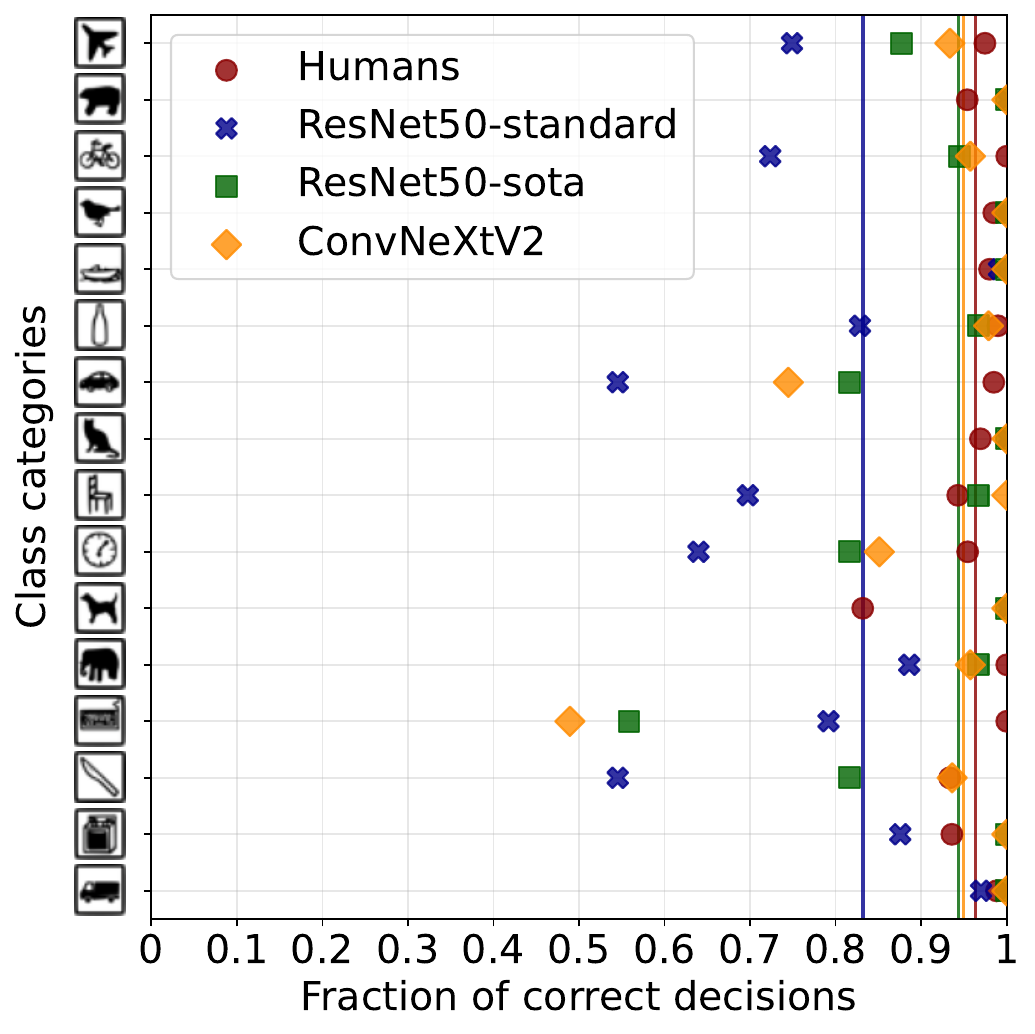}
        \label{subfigure:humans_vs_cnns_global_shape}
    \end{subfigure}
    \begin{subfigure}{\subfigwidth\linewidth}
        \includegraphics[width=\linewidth]{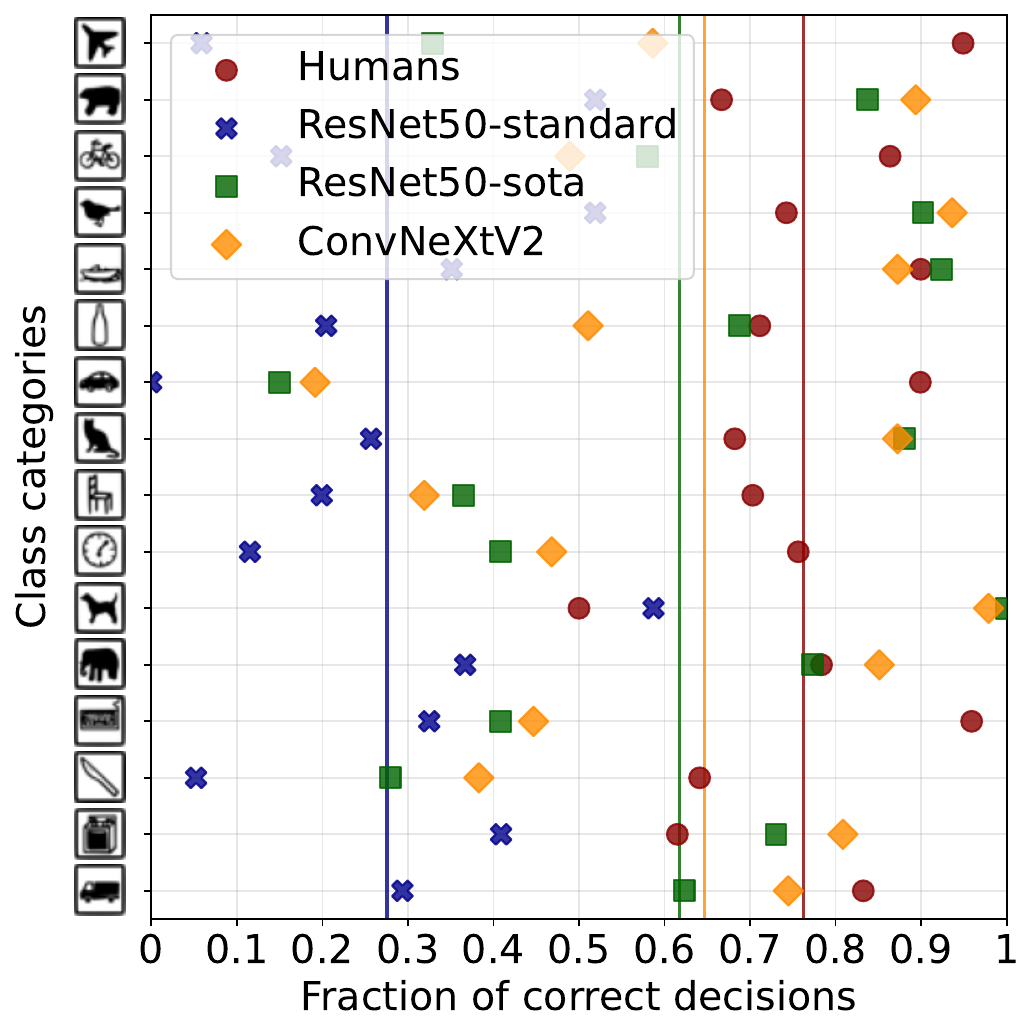}
        \label{subfigure:humans_vs_cnns_local_shape}
    \end{subfigure}
    \begin{subfigure}{\subfigwidth\linewidth}
        \includegraphics[width=\linewidth]{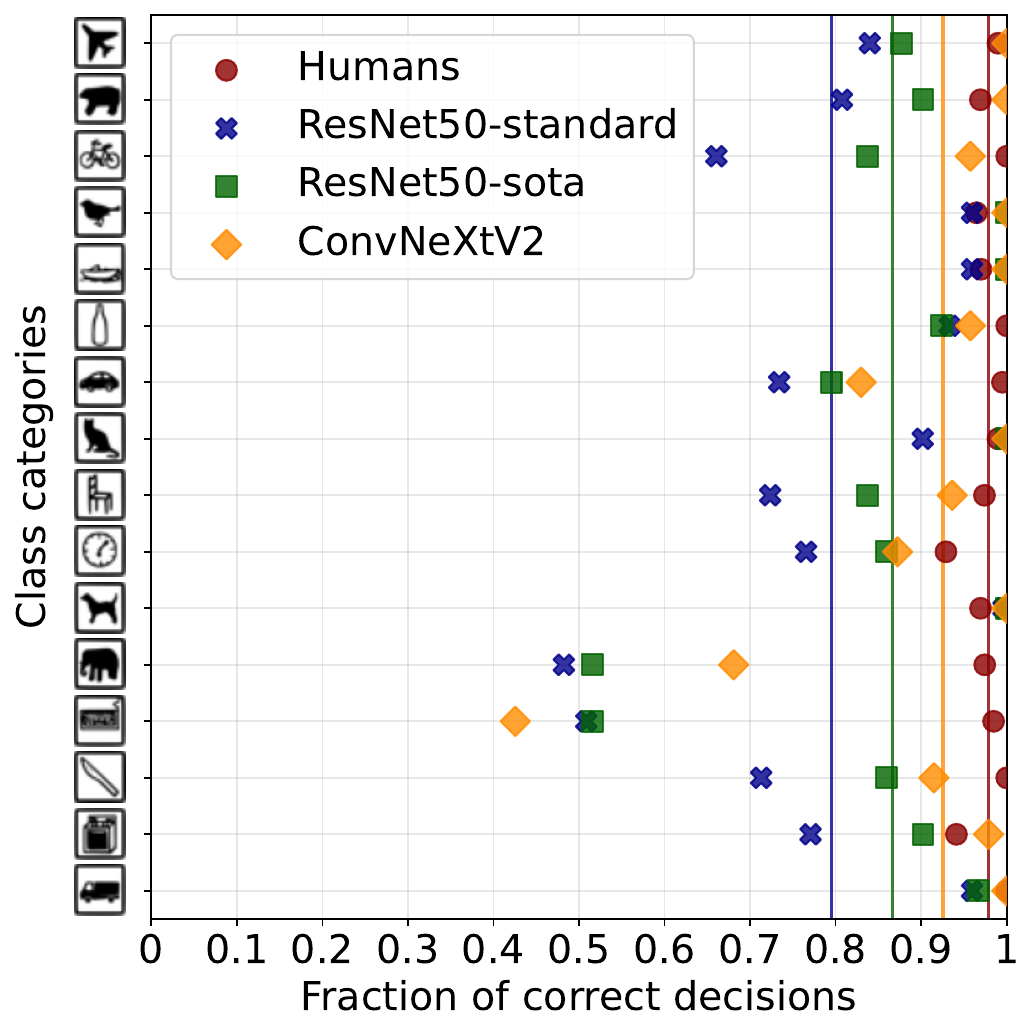}
        \label{subfigure:humans_vs_cnns_texture}
    \end{subfigure}
    \begin{subfigure}{\subfigwidth\linewidth}
        \includegraphics[width=\linewidth]{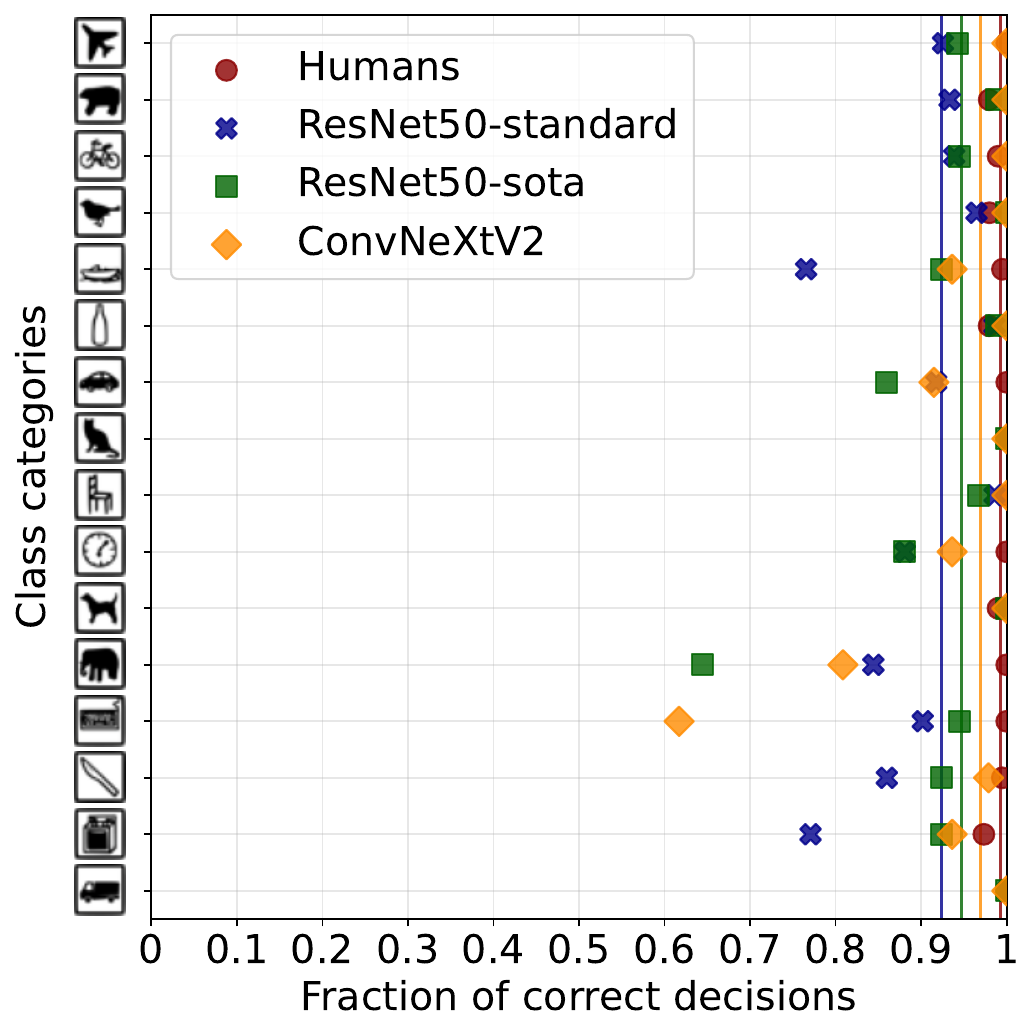}
        \label{subfigure:humans_vs_cnns_color}
    \end{subfigure}
    \vspace{-6pt}

    \caption{Relative accuracy under feature suppression for human observers and three \gls{CNNs} ResNet50-standard, ResNet50-sota, ConvNeXtV2 on the curated ImageNet16 dataset. Each subplot shows performance under suppression of a specific feature: \textbf{(a)} global shape via Patch Shuffle (grid=3); \textbf{(b)} local shape via Patch Shuffle (grid=6); \textbf{(c)} texture via bilateral filtering; and \textbf{(d)} color via grayscale.}
    \label{fig:humans_vs_cnns}
\end{figure}

\begin{table*}[t]
\small
\centering
\caption{Relative accuracy (accuracy under suppression divided by accuracy on original images) for each feature suppression type across models and human observers.}
\label{tab:feature_reliance}
\renewcommand{\tabularxcolumn}[1]{>{\centering\arraybackslash}m{#1}} 
\setlength{\tabcolsep}{4pt} 
\begin{tabularx}{\textwidth}{l*{4}{X}|X X}
\toprule
\textbf{Architecture} & \textbf{Global Shape} & \textbf{Local Shape} & \textbf{Texture} & \textbf{Color} & \textbf{Original} & \textbf{\#Params} \\
\midrule
Humans & 0.965 & 0.763 & 0.979 & 0.999 & 0.969 & -- \\
\midrule
ResNet50-standard \cite{he_deep_2016} & 0.832 & 0.276 & 0.795 & 0.924 & 0.954 & 25.6M \\
ResNet50-sota \cite{wightman_resnet_2021} & 0.943 & 0.618 & 0.867 & 0.948 & 0.931 & 25.6M \\
ConvNeXt \cite{liu_convnet_2022} & 0.938 & 0.606 & 0.910 & 0.961 & 0.934 & 28.6M \\
ConvNeXtV2 \cite{woo_convnext_2023} & 0.949 & 0.647 & 0.925 & 0.969 & 0.940 & 28.6M \\
EfficientNet \cite{tan_efficientnet_2019} & 0.870 & 0.240 & 0.892 & 0.987 & 0.856 & 30.0M \\
EfficientNetV2 \cite{tan_efficientnetv2_2021} & 0.926 & 0.423 & 0.897 & 0.957 & 0.932 & 24.0M \\
MobileNetV3 \cite{howard_searching_2019} & 0.795 & 0.217 & 0.761 & 0.859 & 0.881 & 5.4M \\
ConvMixer \cite{trockman_patches_2023} & 0.920 & 0.437 & 0.815 & 0.891 & 0.874 & 21.1M \\
\midrule
ViT \cite{dosovitskiy_image_2021} & 0.930 & 0.636 & 0.921 & 0.977 & 0.929 & 86.6M \\
DeiT \cite{touvron_training_2021} & 0.938 & 0.730 & 0.926 & 0.969 & 0.932 & 86.6M \\
Swin \cite{liu_swin_2021} & 0.924 & 0.713 & 0.906 & 0.941 & 0.945 & 87.8M \\
CLIP ViT \cite{radford_learning_2021} & 0.959 & 0.758 & 0.949 & 0.984 & 0.936 & 86.6M \\
\bottomrule
\end{tabularx}
\end{table*}

\subsection{Experiment II: Domain-specific Feature Reliance}\label{sec:experiment2}

While Section 5.1 focuses on comparing feature reliance between humans and \gls{CNNs} on a fixed benchmark, this section explores how reliance on shape, texture, and color varies across domains. The same suppression-based framework introduced earlier is applied to three representative visual domains: \gls{CV}, \gls{MI}, and \gls{RS}. In each case, we fix the architecture to a ResNet50 and apply the standard training protocol, including only the data augmentation techniques random resized crop and horizontal flip. For \gls{CV} datasets, we either train from scratch or initialize models with ImageNet-pretrained weights (standard training protocol) and then fine-tune on the respective datasets. For \gls{MI} and \gls{RS}, we train from scratch to allow a disentangled comparison across domains. Additional results for \gls{MI} and \gls{RS} with pretrained models to simulate operational scenarios can be found in the supplemental material. Details about the hyperparameter, as well as an overview of the corresponding validation accuracies, are provided in the supplemental material. In contrast to the previous experiment, in this experiment, suppression strength is treated as a continuous hyperparameter and systematically varied to obtain suppression curves that characterize feature reliance across domains. To reduce redundancy, we report results using one representative suppression technique per feature type in the main paper. Results using alternative suppression methods per feature type are included in the supplemental material and exhibit qualitatively similar patterns across domains.

\begin{figure}[t!]
    \def\subfigwidth{.31}
    \centering

    \makebox[\linewidth][c]{%
        \begin{minipage}{\subfigwidth\linewidth}
            \centering\scriptsize\textbf{(a) CV Shape Suppression}
        \end{minipage}
        \hspace{0.1cm}
        \begin{minipage}{\subfigwidth\linewidth}
            \centering\scriptsize\textbf{(b) CV Texture Suppression}
        \end{minipage}
        \hspace{0.1cm}
        \begin{minipage}{\subfigwidth\linewidth}
            \centering\scriptsize\textbf{(c) CV Color Suppression}
        \end{minipage}
    }

    \vspace{2pt}

    \begin{subfigure}{\subfigwidth\linewidth}
        \includegraphics[width=\linewidth]{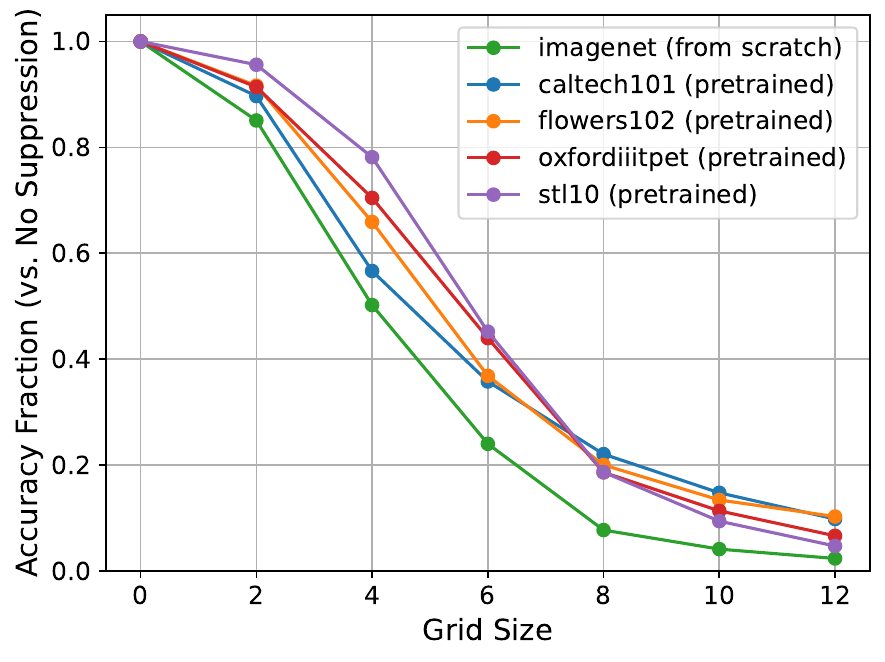}
        \label{subfigure:cv_pretrained_shape}
    \end{subfigure}
    \hspace{0.1cm}
    \begin{subfigure}{\subfigwidth\linewidth}
        \includegraphics[width=\linewidth]{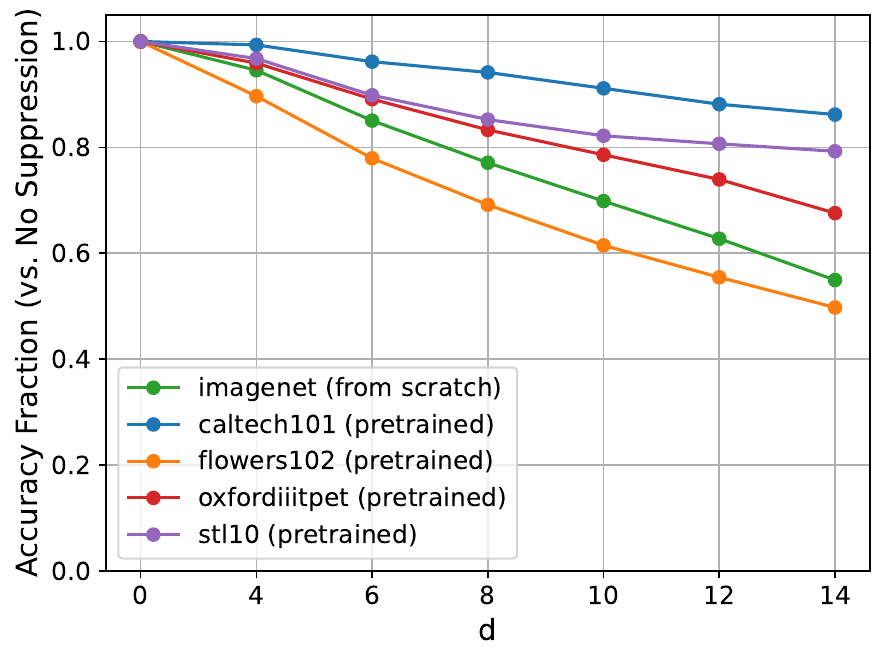}
        \label{subfigure:cv_pretrained_patch_texture}
    \end{subfigure}
    \hspace{0.1cm}
    \begin{subfigure}{\subfigwidth\linewidth}
        \includegraphics[width=\linewidth]{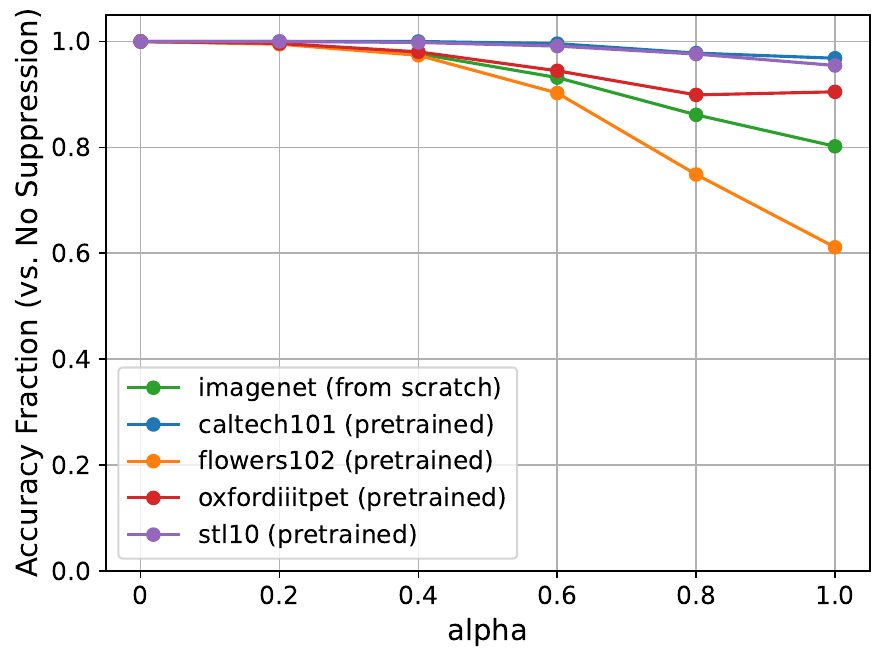}
        \label{subfigure:cv_pretrained_color}
    \end{subfigure}

  \vspace{-10pt}
  
    \makebox[\linewidth][c]{%
        \begin{minipage}{\subfigwidth\linewidth}
            \centering\scriptsize\textbf{(d) MI Shape Suppression}
        \end{minipage}
        \hspace{0.1cm}
        \begin{minipage}{\subfigwidth\linewidth}
            \centering\scriptsize\textbf{(e) MI Texture Suppression}
        \end{minipage}
        \hspace{0.1cm}
        \begin{minipage}{\subfigwidth\linewidth}
            \centering\scriptsize\textbf{(f) MI Color Suppression}
        \end{minipage}
    }

    \vspace{2pt}

    \begin{subfigure}{\subfigwidth\linewidth}
        \includegraphics[width=\linewidth]{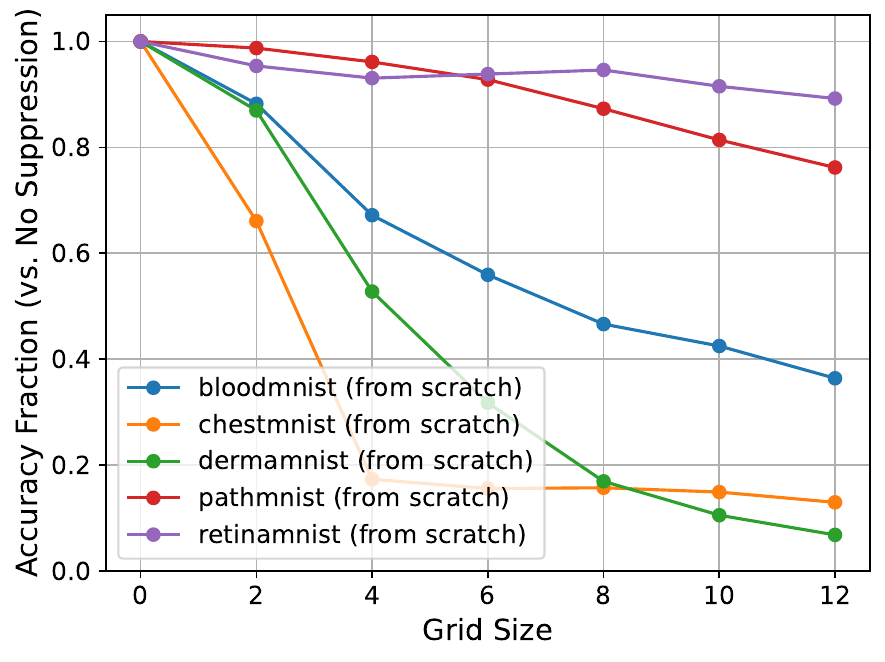}
        \label{subfigure:mi_shape}
    \end{subfigure}
    \hspace{0.1cm}
    \begin{subfigure}{\subfigwidth\linewidth}
        \includegraphics[width=\linewidth]{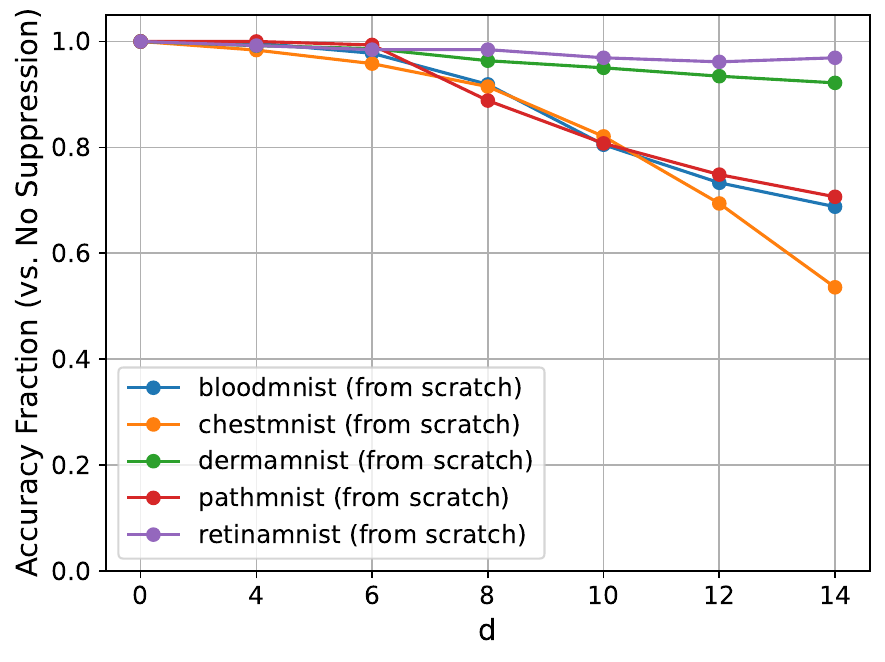}
        \label{subfigure:mi_patch_texture}
    \end{subfigure}
    \hspace{0.1cm}
    \begin{subfigure}{\subfigwidth\linewidth}
        \includegraphics[width=\linewidth]{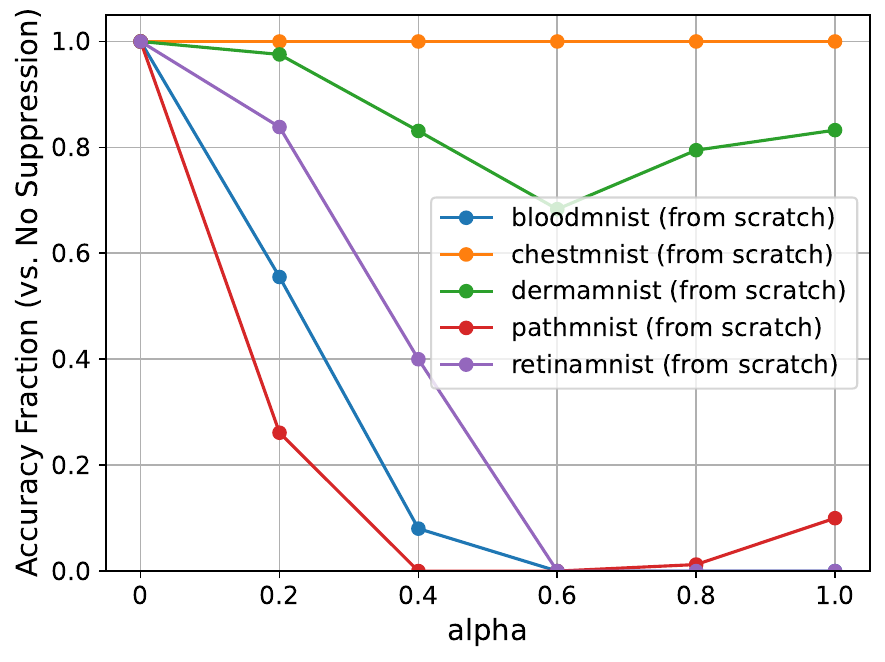}
        \label{subfigure:mi_color}
    \end{subfigure}

   \vspace{-10pt}

    \makebox[\linewidth][c]{%
        \begin{minipage}{\subfigwidth\linewidth}
            \centering\scriptsize\textbf{(g) RS Shape Suppression}
        \end{minipage}
        \hspace{0.1cm}
        \begin{minipage}{\subfigwidth\linewidth}
            \centering\scriptsize\textbf{(h) RS Texture Suppression}
        \end{minipage}
        \hspace{0.1cm}
        \begin{minipage}{\subfigwidth\linewidth}
            \centering\scriptsize\textbf{(i) RS Color Suppression}
        \end{minipage}
    }

    \vspace{2pt}

    \begin{subfigure}{\subfigwidth\linewidth}
        \includegraphics[width=\linewidth]{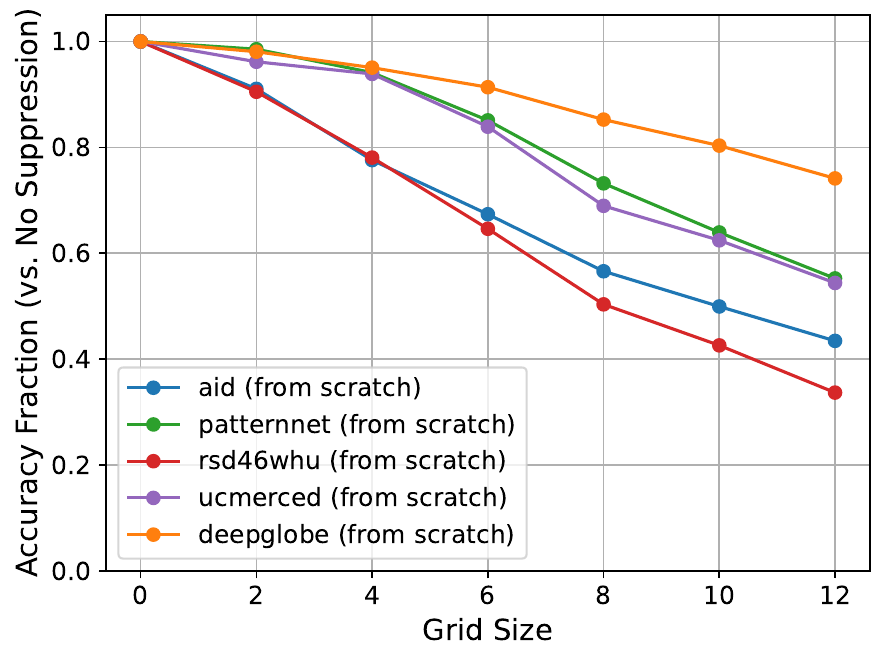}
        \label{subfigure:rs_shape}
    \end{subfigure}
    \hspace{0.1cm}
    \begin{subfigure}{\subfigwidth\linewidth}
        \includegraphics[width=\linewidth]{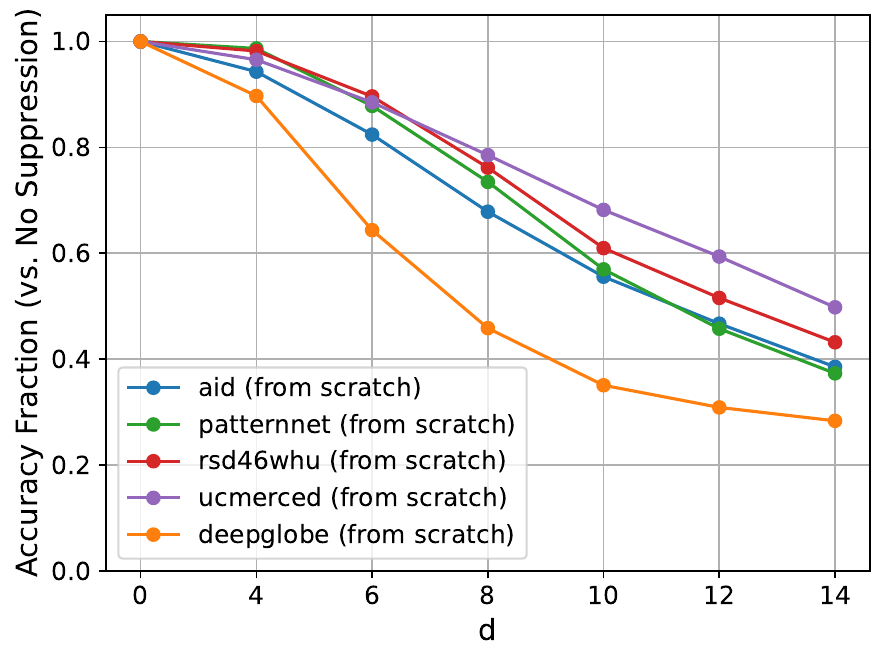}
        \label{subfigure:rs_patch_texture}
    \end{subfigure}
    \hspace{0.1cm}
    \begin{subfigure}{\subfigwidth\linewidth}
        \includegraphics[width=\linewidth]{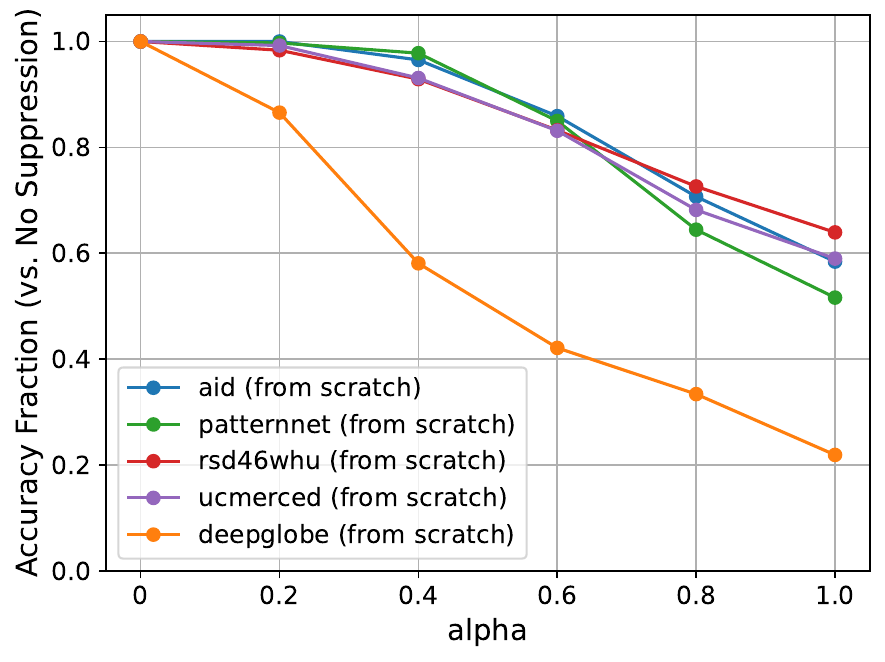}
        \label{subfigure:rs_color}
    \end{subfigure}
    \vspace{-5pt}
    \caption{Feature suppression results across three domains. 
\textbf{Top row (a–c)}: ResNet50 pretrained on ImageNet and fine-tuned on \gls{CV} datasets. 
\textbf{Middle row (d–f)}: ResNet50 trained from scratch on MI datasets from MedMNIST-v2. 
\textbf{Bottom row (g–i)}: ResNet50 trained from scratch on high-resolution \gls{RS} datasets. 
Columns correspond to: \textbf{(a, d, g)} shape suppression (Patch Shuffle), \textbf{(b, e, h)} texture suppression (Bilateral Filter), and \textbf{(c, f, i)} color suppression (Grayscale).}
    \label{fig:suppression_grid}
\end{figure}

\begin{figure}[t!]
    \def\subfigwidth{.31}
    \centering

    \makebox[\linewidth][c]{%
        \begin{minipage}{\subfigwidth\linewidth}
            \centering\scriptsize\textbf{(a) Shape Suppression}
        \end{minipage}
        \hspace{0.1cm}
        \begin{minipage}{\subfigwidth\linewidth}
            \centering\scriptsize\textbf{(b) Texture Suppression}
        \end{minipage}
        \hspace{0.1cm}
        \begin{minipage}{\subfigwidth\linewidth}
            \centering\scriptsize\textbf{(c) Color Suppression}
        \end{minipage}
    }

    \vspace{2pt}

    \begin{subfigure}{\subfigwidth\linewidth}
        \includegraphics[width=\linewidth]{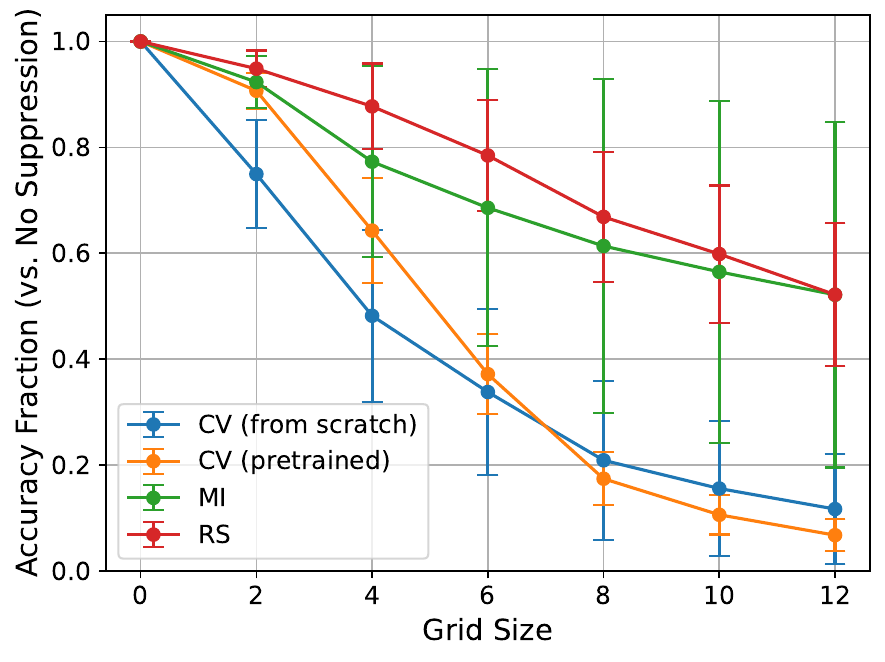}
        \label{subfigure:domain_comparison_shape}
    \end{subfigure}
    \hspace{0.1cm}
    \begin{subfigure}{\subfigwidth\linewidth}
        \includegraphics[width=\linewidth]{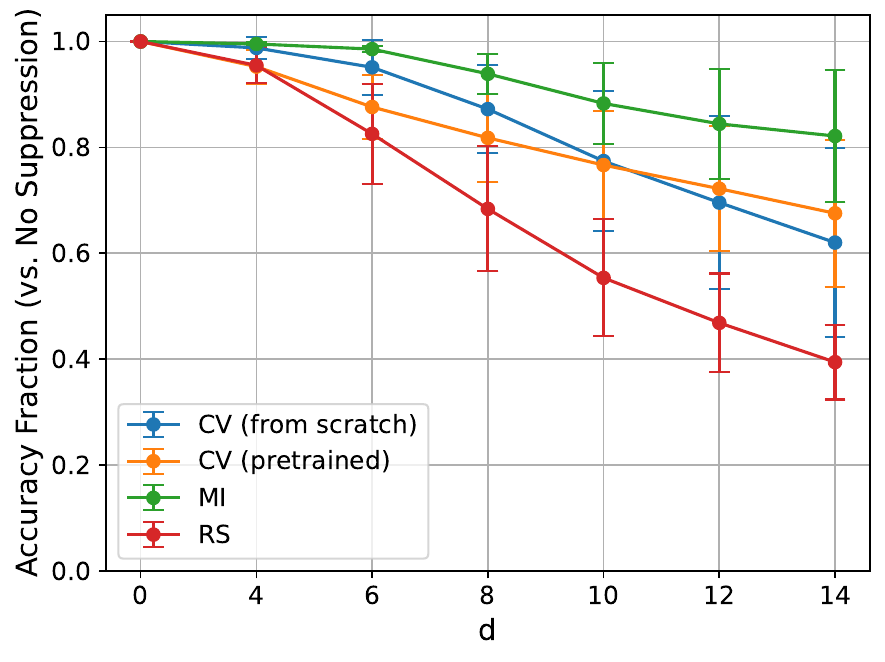}
        \label{subfigure:domain_comparison_patch_texture}
    \end{subfigure}
    \hspace{0.1cm}
    \begin{subfigure}{\subfigwidth\linewidth}
        \includegraphics[width=\linewidth]{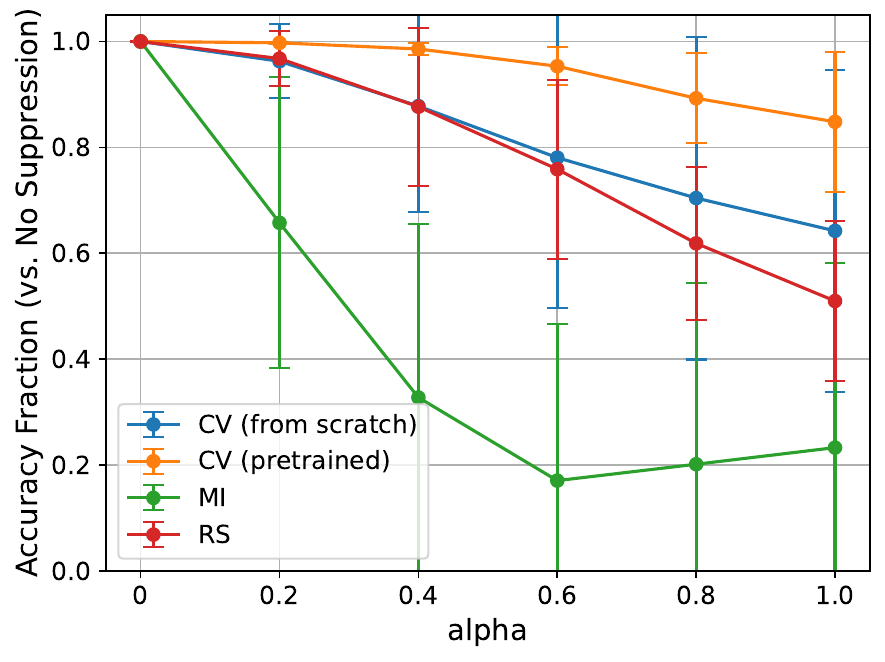}
        \label{subfigure:domain_comparison_color}
    \end{subfigure}
    \vspace{-5pt}
    \caption{Domain-averaged feature suppression curves for CV, MI, and RS. \textbf{(a)} Shape suppression via Patch Shuffle. \textbf{(b)} Texture suppression via bilateral filtering. \textbf{(c)} Color suppression via grayscale.}
    \label{fig:domain_comparison}
\end{figure}

To visualize domain-specific suppression sensitivity, we present a composite figure of per-domain results in \Cref{fig:suppression_grid}, showing the effect of suppressing shape, texture, and color for datasets from each domain. To ensure comparability across datasets with different numbers of classes and baseline accuracies, we standardize performance by rescaling: chance-level accuracy is mapped to 0, and baseline accuracy (i.e., accuracy on original images) is mapped to 1. Relative accuracy under suppression is then expressed on this normalized scale, facilitating direct comparison of feature reliance across domains and datasets. Finally, to synthesize the findings, we aggregate suppression curves in a domain-level comparison (\Cref{fig:domain_comparison}) by averaging results across datasets within each domain.

\textbf{Computer Vision (CV).} \Cref{fig:suppression_grid}a–c shows suppression results for five standard \gls{CV} benchmarks (ImageNet \cite{deng_imagenet_2009}, Caltech101 \cite{fei-fei_learning_2004}, Flowers102 \cite{nilsback_automated_2008}, Oxford-IIIT-Pet \cite{parkhi_cats_2012}, STL10 \cite{coates_analysis_2011}). Across datasets, we observe that shape suppression induces the strongest performance degradation, especially as the patch shuffle grid size increases. This confirms a pronounced reliance on local shape information in pretrained \gls{CNNs}. In contrast, texture suppression via bilateral filtering has minimal effect, and color suppression through grayscale conversion yields only minor degradation, indicating that \gls{CNNs} fine-tuned on these datasets are largely robust to the removal of texture and color cues. These results are consistent with our human comparison study and suggest that local shape continues to dominate feature reliance in natural image classification tasks. For completeness, the supplemental material includes results for models trained from scratch as well as a class-wise analysis for ImageNet, which confirms that the global reliance patterns are consistent across categories.

\textbf{Medical Imaging (MI).} \Cref{fig:suppression_grid}d–f summarizes results on five datasets from the MedMNIST-v2 collection \cite{yang_medmnist_2023}: PathMNIST, RetinaMNIST, BloodMNIST, DermaMNIST, and ChestMNIST. We use the standardized \( 224 \times 224 \) pixels version to ensure consistency with the experimental setup. Across these datasets, suppression effects are more heterogeneous than in \gls{CV}. While shape suppression degrades performance, the impact is generally less pronounced, and texture suppression yields moderate performance drops in datasets such as PathMNIST and BloodMNIST, but relatively little effect in RetinaMNIST and DermaMNIST. By contrast, color suppression induces a substantial decline in classification accuracy for most datasets, reflecting the strong diagnostic role of chromatic cues, except in ChestMNIST, which contains only grayscale images. Taken together, these results suggest that feature reliance in \gls{MI} varies substantially across datasets, with a common trend towards greater dependence on color information.

\textbf{Remote Sensing (RS).} \Cref{fig:suppression_grid}g–i reports suppression curves for five very-high-resolution RGB datasets: UCMerced \cite{yang_bag--visual-words_2010}, RSD46-WHU \cite{long_accurate_2017}, DeepGlobe \cite{demir_deepglobe_2018}, PatternNet \cite{zhou_patternnet_2018}, and AID \cite{xia_aid_2017}. As in MI, shape suppression impacts performance, but the degradation is less pronounced than in the \gls{CV} domain, indicating lower reliance on local shape. In contrast to \gls{CV} and MI, texture suppression leads to substantial performance degradation across all datasets, suggesting that fine-grained surface patterns are critical for \gls{RS} classification. Surprisingly, color suppression also results in notable performance drops, despite the use of RGB imagery only. This likely reflects strong correlations between chromatic cues and semantic land cover categories. Overall, \gls{RS} models exhibit a pronounced reliance on texture and color, and comparatively less dependence on local shape, reflecting the distinct statistical structure and spatial semantics of \gls{RS} imagery.

\textbf{Cross-Domain Comparison.} To synthesize these observations, \Cref{fig:domain_comparison} presents the domain-averaged suppression curves for each feature, including 1-sigma error bars. Three clear trends emerge. First, \gls{CV} models are most reliant on local shape, especially when trained from scratch, while ImageNet pretraining induces slightly greater robustness. Second, \gls{MI} models exhibit stronger dependence on color, consistent with the nature of some medical tasks (e.g., in dermatology, histopathology), which often require interpreting chromatic cues. Third, \gls{RS} models exhibit the highest texture reliance among the three tested domains. This may reflect the nature of many \gls{RS} classes that are defined by texture-like patterns (e.g., fields, residential areas), rather than by distinct global contours. These patterns confirm that feature reliance is shaped not only by architecture and training regime, but also by the visual and semantic properties of the task or domain.

Finally, to validate the observed feature reliance patterns, we conduct complementary experiments on \gls{CV} datasets with simultaneous suppression of two features (see \Cref{sec:joint_suppression} in the supplemental material). Results confirm the trends of single-feature suppression: performance is highest when only shape is preserved, reduced when only texture remains, and nearly lost when only color is available. In summary, the findings highlight that domain characteristics, alongside architecture and training regime, play a crucial role in shaping feature reliance. While prior work emphasized architecture-induced biases, our results suggest that data properties equally govern the perceptual strategies that models adopt. 
\newpage
\section{Conclusion}\label{sec:conclusion} This paper revisited the widely cited claim that \gls{CNNs} trained on ImageNet are inherently biased towards texture. We identify critical conceptual and methodological limitations in the cue-conflict experiment popularized by Geirhos et al. \cite{geirhos_imagenet-trained_2019} that support this hypothesis. Further, we propose a new framework for evaluating feature reliance based on targeted suppression rather than forced-choice preference. Using this framework, we find no evidence for an inherent texture bias in \gls{CNNs}, but instead observed a pronounced reliance on local shape features. Nonetheless, we show that this reliance can be substantially mitigated through modern training strategies. Across domains, we find that feature reliance varies substantially: \gls{CV} models prioritize shape, \gls{MI} models rely more evenly on color, and \gls{RS} models exhibit strong texture sensitivity. These findings challenge the notion of fixed architectural biases and instead position feature reliance as a flexible property shaped by optimization objectives and domain-specific semantics, offering new directions for designing models that better align with human perceptual strategies. At the same time, the relative contributions of architectural components and training strategies to these reliance patterns remain to be systematically evaluated.

\textbf{Limitations.} Our framework relies on operational definitions of shape, texture, and color based on specific transformations, but features are continuous and interdependent, limiting perfect isolation. In practice, suppression only reduces rather than eliminates features: texture suppression can leave residual low-level features perceptible as texture, while shape suppression does not fully remove all shape cues. This reflects the inherent trade-off of reducing one feature while preserving others, making absolute removal unattainable. The applied suppression techniques may also introduce artifacts that affect model behavior independently of the targeted features (e.g., block-like structures from Patch Shuffle, smoothing from filtering). Further, the results obtained with pretrained models may reflect effects of similarities between suppression transformations and augmentation techniques (e.g., Cutout and Patch Shuffle). Finally, our human experiments employed a controlled forced-choice design with brief exposures and a limited set of categories to ensure comparability. While necessary for experimental control, these constraints may not fully reflect the richness and adaptability of human visual perception in real-world settings.
\section*{Acknowledgements}

We thank Johanna Vielhaben and Genc Hoxha for suggesting the addition of quantitative experiments to validate the feature suppression techniques, Liliann Lehrke for valuable guidance on the design of the human study, and Christopher Olk for helpful discussions in selecting the title. This work was partly supported by EU Horizon projects ELIAS (No. 101120237) and ELLIOT (No. 101214398).
{
    \bibliographystyle{unsrtnat}  
    \bibliography{main}
}




\newpage
\appendix
\clearpage

\section{Notation of Feature Suppression Metrics}

To quantify the effect of feature suppression, we define a transformation \( T_\tau: \mathbb{R}^{H \times W} \rightarrow \mathbb{R}^{H \times W} \) that maps an original grayscale image \( \mathbf{x} \) to its feature-suppressed version \( \hat{\mathbf{x}} = T_\tau(\mathbf{x}) \), where \( \tau \in \{\text{texture}, \text{shape}\} \). We then compare \( \mathbf{x} \) and \( \hat{\mathbf{x}} \) using a set of similarity metrics \( \phi(\cdot) \) that target either texture or shape characteristics.

\textbf{(1) Local Variance (LV).}  
This metric quantifies local contrast variability and serves as a proxy for fine-grained texture information. We compute the mean variance over non-overlapping windows $\mathbf{w}_{i,j}$ of size $k \times k$:

\begin{align}
    \label{eq:lv1}
\phi_{\mathrm{var}}(\mathbf{x}) = \frac{1}{N} \sum_{i,j} \mathrm{Var}(\mathbf{w}_{i,j}),
\end{align}

\begin{align}
    \label{eq:lv2}
    \mathrm{LV}(\mathbf{x}, \hat{\mathbf{x}}) = \min\left(1, \frac{\phi_{\mathrm{var}}(\hat{\mathbf{x}})}{\phi_{\mathrm{var}}(\mathbf{x})} \right).
\end{align}

\textbf{(2) High-Frequency Energy Ratio (HFE).}  
This metric captures the spectral energy in high frequencies and is used to measure texture preservation. Using the 2D Fourier transform $\mathcal{F}(\cdot)$, we compute:

\begin{align}
    \label{eq:hf1}
\phi_{\mathrm{hf}}(\mathbf{x}) = \frac{\sum_{(u,v) \in \mathcal{H}} |\mathcal{F}(\mathbf{x})_{u,v}|^2}{\sum_{(u,v)} |\mathcal{F}(\mathbf{x})_{u,v}|^2},
\end{align}

\begin{align}
    \label{eq:hf2}
\mathrm{HFE}(\mathbf{x}, \hat{\mathbf{x}}) = \min\left(1, \frac{\phi_{\mathrm{hf}}(\hat{\mathbf{x}})}{\phi_{\mathrm{hf}}(\mathbf{x})} \right),
\end{align}

where $\mathcal{H}$ is the set of frequency components with distance greater than radius $r$ from the center.

\textbf{(3) Edge Structural Similarity (ESSIM).}  
This metric evaluates the similarity of edge structures, capturing shape information. Sobel gradients are computed with kernel size $k$:

\begin{align}
    \label{eq:essim1}
\phi_{\mathrm{sobel}}(\mathbf{x}) = \sqrt{(\partial_x \mathbf{x})^2 + (\partial_y \mathbf{x})^2},
\end{align}

\begin{align}
    \label{eq:essim2}
\mathrm{ESSIM}(\mathbf{x}, \hat{\mathbf{x}}) = \mathrm{SSIM}\left(\phi_{\mathrm{sobel}}(\mathbf{x}), \phi_{\mathrm{sobel}}(\hat{\mathbf{x}})\right).
\end{align}

\textbf{(4) Gradient Correlation (GC).}  
This metric compares first-order gradients along both axes and targets shape preservation. It is defined as:
\begin{align}
    \label{eq:gc1}
g_x(\mathbf{x}) = \frac{\partial \mathbf{x}}{\partial x}, \quad g_y(\mathbf{x}) = \frac{\partial \mathbf{x}}{\partial y},
\end{align}

\begin{align}
    \label{eq:gc2}
\mathrm{GC}(\mathbf{x}, \hat{\mathbf{x}}) = \frac{1}{2} \left[ \mathrm{corr}(g_x(\mathbf{x}), g_x(\hat{\mathbf{x}})) + \mathrm{corr}(g_y(\mathbf{x}), g_y(\hat{\mathbf{x}})) \right].
\end{align}

\textbf{Averaged Texture and Shape Metrics.}  
We aggregate individual metrics into a Texture score and a Shape score using the \textit{arithmetic mean}:
\begin{align}
    \label{eq:texture_score}
\mathrm{Texture} = \frac{1}{2} \left( \mathrm{LVS} + \mathrm{HF} \right),
\end{align}
\begin{align}
    \label{eq:shape_score}
\mathrm{Shape} = \frac{1}{2} \left( \mathrm{EdgeSSIM} + \mathrm{GC} \right).
\end{align}

All metrics are bounded in $[0, 1]$, where higher values indicate greater similarity to the original image and hence lower suppression of the respective feature.  
We set all hyperparameters to $w = r = k = 11$.

\section{Notation for Relative Accuracy}

To ensure comparability across datasets with different numbers of classes and baseline accuracies, we standardize performance using a linear rescaling:

\begin{align}
    \label{eq:rel_acc}
\mathrm{RelativeAccuracy} = \frac{A_{\text{sup}} - A_{\text{chance}}}{A_{\text{orig}} - A_{\text{chance}}},
\end{align}

where \( A_{\text{sup}} \) denotes the accuracy under feature suppression, \( A_{\text{orig}} \) the accuracy on the original (unsuppressed) images, and \( A_{\text{chance}} = \frac{1}{C} \) the chance-level accuracy for \( C \) classes in the dataset. This mapping assigns 0 to chance-level accuracy and 1 to original accuracy, allowing direct comparison of suppression sensitivity across datasets and domains. For multi-label classification tasks, we estimate \( A_{\text{chance}} \) by simulating random predictions and computing the expected mean average precision.

\section{Visual Examples for Suppression Transformations}\label{sec:visual_examples}

\begin{figure}[t!]
    \def\subfigwidth{.31}
    \centering

    \makebox[\linewidth][c]{%
        \begin{minipage}{\subfigwidth\linewidth}
            \centering\scriptsize\textbf{(a) Original}
        \end{minipage}
        \hspace{0.1cm}
        \begin{minipage}{\subfigwidth\linewidth}
            \centering\scriptsize\textbf{(b) Global Shape (Patch Shuffle)}
        \end{minipage}
        \hspace{0.1cm}
        \begin{minipage}{\subfigwidth\linewidth}
            \centering\scriptsize\textbf{(c) Local Shape (Patch Rotation)}
        \end{minipage}
    }

    \vspace{2pt}

    \begin{subfigure}{\subfigwidth\linewidth}
        \includegraphics[width=\linewidth]{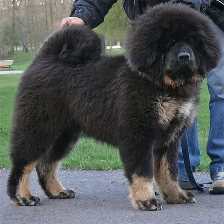}
        \label{subfigure:imagenet_idx_539_original1}
    \end{subfigure}
    \hspace{0.1cm}
    \begin{subfigure}{\subfigwidth\linewidth}
        \includegraphics[width=\linewidth]{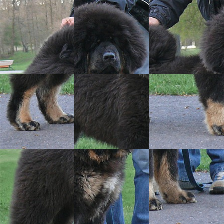}
        \label{subfigure:imagenet_idx_539_patch_shuffle}
    \end{subfigure}
    \hspace{0.1cm}
    \begin{subfigure}{\subfigwidth\linewidth}
        \includegraphics[width=\linewidth]{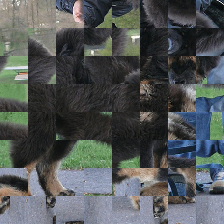}
        \label{subfigure:imagenet_idx_539_patch_rotation}
    \end{subfigure}

  \vspace{-10pt}
  
    \makebox[\linewidth][c]{%
        \begin{minipage}{\subfigwidth\linewidth}
            \centering\scriptsize\textbf{(d) Original}
        \end{minipage}
        \hspace{0.1cm}
        \begin{minipage}{\subfigwidth\linewidth}
            \centering\scriptsize\textbf{(e) Texture (Bilateral Filter)}
        \end{minipage}
        \hspace{0.1cm}
        \begin{minipage}{\subfigwidth\linewidth}
            \centering\scriptsize\textbf{(f) Texture (Gaussian Blur)}
        \end{minipage}
    }

    \vspace{2pt}

    \begin{subfigure}{\subfigwidth\linewidth}
        \includegraphics[width=\linewidth]{images/visual_examples/imagenet_idx_539_original.png}
    \label{subfigure:imagenet_idx_539_original2}
    \end{subfigure}
    \hspace{0.1cm}
    \begin{subfigure}{\subfigwidth\linewidth}
        \includegraphics[width=\linewidth]{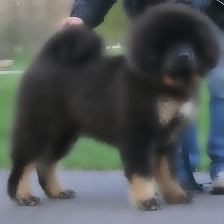}
        \label{subfigure:imagenet_idx_539_bilateral}
    \end{subfigure}
    \hspace{0.1cm}
    \begin{subfigure}{\subfigwidth\linewidth}
        \includegraphics[width=\linewidth]{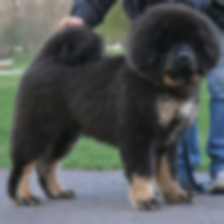}
        \label{subfigure:imagenet_idx_539_gaussian}
    \end{subfigure}

   \vspace{-10pt}

    \makebox[\linewidth][c]{%
        \begin{minipage}{\subfigwidth\linewidth}
            \centering\scriptsize\textbf{(g) Original}
        \end{minipage}
        \hspace{0.1cm}
        \begin{minipage}{\subfigwidth\linewidth}
            \centering\scriptsize\textbf{(h) Color (Grayscale)}
        \end{minipage}
        \hspace{0.1cm}
        \begin{minipage}{\subfigwidth\linewidth}
            \centering\scriptsize\textbf{(i) Color (Channel Shuffle)}
        \end{minipage}
    }

    \vspace{2pt}

    \begin{subfigure}{\subfigwidth\linewidth}
        \includegraphics[width=\linewidth]{images/visual_examples/imagenet_idx_539_original.png}
        \label{subfigure:imagenet_idx_539_original3}
    \end{subfigure}
    \hspace{0.1cm}
    \begin{subfigure}{\subfigwidth\linewidth}
        \includegraphics[width=\linewidth]{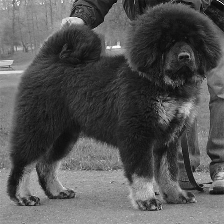}
        \label{subfigure:imagenet_idx_539_grayscale}
    \end{subfigure}
    \hspace{0.1cm}
    \begin{subfigure}{\subfigwidth\linewidth}
        \includegraphics[width=\linewidth]{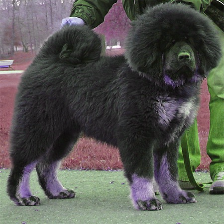}
        \label{subfigure:imagenet_idx_539_channel_shuffle}
    \end{subfigure}
    \vspace{-5pt}
    \caption{Visual illustration of feature suppression transformations applied to a sample image from the ImageNet validation set belonging to the entry-level category dog. \textbf{(a, d, g)} Show the original image. \textbf{(b)} Global shape suppression via Patch Shuffle with grid size 3. \textbf{(c)} Local shape suppression via Patch Rotation with grid size 8. \textbf{(e)} Texture suppression using Bilateral Filtering with \( d = 12 \), \( \sigma_{\text{color}} = 170 \), and \( \sigma_{\text{space}} = 75 \). \textbf{(f)} Texture suppression using Gaussian Blur with kernel size \( k = 11 \) and standard deviation \( \sigma = 2.0 \). \textbf{(h)} Color suppression via grayscale conversion. \textbf{(i)} Color suppression via random channel shuffle.}
    \label{fig:visual_example_1}
\end{figure}

\begin{figure}[t!]
    \def\subfigwidth{.31}
    \centering

    \makebox[\linewidth][c]{%
        \begin{minipage}{\subfigwidth\linewidth}
            \centering\scriptsize\textbf{(a) Original}
        \end{minipage}
        \hspace{0.1cm}
        \begin{minipage}{\subfigwidth\linewidth}
            \centering\scriptsize\textbf{(b) Global Shape (Patch Shuffle)}
        \end{minipage}
        \hspace{0.1cm}
        \begin{minipage}{\subfigwidth\linewidth}
            \centering\scriptsize\textbf{(c) Local Shape (Patch Rotation)}
        \end{minipage}
    }

    \vspace{2pt}

    \begin{subfigure}{\subfigwidth\linewidth}
        \includegraphics[width=\linewidth]{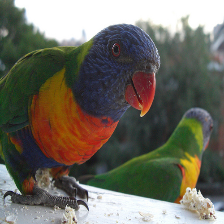}
        \label{subfigure:imagenet_idx_171_original1}
    \end{subfigure}
    \hspace{0.1cm}
    \begin{subfigure}{\subfigwidth\linewidth}
        \includegraphics[width=\linewidth]{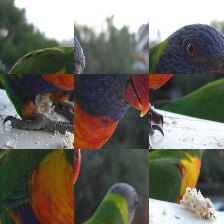}
        \label{subfigure:imagenet_idx_171_patch_shuffle}
    \end{subfigure}
    \hspace{0.1cm}
    \begin{subfigure}{\subfigwidth\linewidth}
        \includegraphics[width=\linewidth]{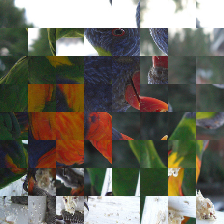}
        \label{subfigure:imagenet_idx_171_patch_rotation}
    \end{subfigure}

  \vspace{-10pt}
  
    \makebox[\linewidth][c]{%
        \begin{minipage}{\subfigwidth\linewidth}
            \centering\scriptsize\textbf{(d) Original}
        \end{minipage}
        \hspace{0.1cm}
        \begin{minipage}{\subfigwidth\linewidth}
            \centering\scriptsize\textbf{(e) Texture (Bilateral Filter)}
        \end{minipage}
        \hspace{0.1cm}
        \begin{minipage}{\subfigwidth\linewidth}
            \centering\scriptsize\textbf{(f) Texture (Gaussian Blur)}
        \end{minipage}
    }

    \vspace{2pt}

    \begin{subfigure}{\subfigwidth\linewidth}
        \includegraphics[width=\linewidth]{images/visual_examples/imagenet_idx_171_original.png}
    \label{subfigure:imagenet_idx_171_original2}
    \end{subfigure}
    \hspace{0.1cm}
    \begin{subfigure}{\subfigwidth\linewidth}
        \includegraphics[width=\linewidth]{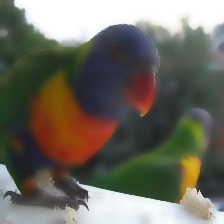}
        \label{subfigure:imagenet_idx_171_bilateral}
    \end{subfigure}
    \hspace{0.1cm}
    \begin{subfigure}{\subfigwidth\linewidth}
        \includegraphics[width=\linewidth]{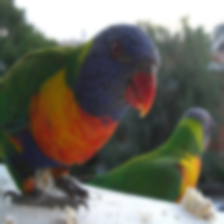}
        \label{subfigure:imagenet_idx_171_gaussian}
    \end{subfigure}

   \vspace{-10pt}

    \makebox[\linewidth][c]{%
        \begin{minipage}{\subfigwidth\linewidth}
            \centering\scriptsize\textbf{(g) Original}
        \end{minipage}
        \hspace{0.1cm}
        \begin{minipage}{\subfigwidth\linewidth}
            \centering\scriptsize\textbf{(h) Color (Grayscale)}
        \end{minipage}
        \hspace{0.1cm}
        \begin{minipage}{\subfigwidth\linewidth}
            \centering\scriptsize\textbf{(i) Color (Channel Shuffle)}
        \end{minipage}
    }

    \vspace{2pt}

    \begin{subfigure}{\subfigwidth\linewidth}
        \includegraphics[width=\linewidth]{images/visual_examples/imagenet_idx_171_original.png}
        \label{subfigure:imagenet_idx_171_original3}
    \end{subfigure}
    \hspace{0.1cm}
    \begin{subfigure}{\subfigwidth\linewidth}
        \includegraphics[width=\linewidth]{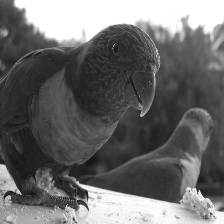}
        \label{subfigure:imagenet_idx_171_grayscale}
    \end{subfigure}
    \hspace{0.1cm}
    \begin{subfigure}{\subfigwidth\linewidth}
        \includegraphics[width=\linewidth]{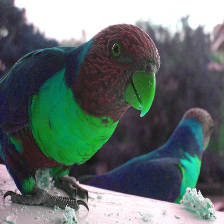}
        \label{subfigure:imagenet_idx_171_channel_shuffle}
    \end{subfigure}
    \vspace{-5pt}
    \caption{Visual illustration of feature suppression transformations applied to a sample image from the ImageNet validation set belonging to the entry-level category bird. \textbf{(a, d, g)} Show the original image. \textbf{(b)} Global shape suppression via Patch Shuffle with grid size 3. \textbf{(c)} Local shape suppression via Patch Rotation with grid size 8. \textbf{(e)} Texture suppression using Bilateral Filtering with \( d = 12 \), \( \sigma_{\text{color}} = 170 \), and \( \sigma_{\text{space}} = 75 \). \textbf{(f)} Texture suppression using Gaussian Blur with kernel size \( k = 11 \) and standard deviation \( \sigma = 2.0 \). \textbf{(h)} Color suppression via grayscale conversion. \textbf{(i)} Color suppression via random channel shuffle.}
    \label{fig:visual_example_2}
\end{figure}

\begin{figure}[t!]
    \def\subfigwidth{.31}
    \centering

    \makebox[\linewidth][c]{%
        \begin{minipage}{\subfigwidth\linewidth}
            \centering\scriptsize\textbf{(a) Original}
        \end{minipage}
        \hspace{0.1cm}
        \begin{minipage}{\subfigwidth\linewidth}
            \centering\scriptsize\textbf{(b) Global Shape (Patch Shuffle)}
        \end{minipage}
        \hspace{0.1cm}
        \begin{minipage}{\subfigwidth\linewidth}
            \centering\scriptsize\textbf{(c) Local Shape (Patch Rotation)}
        \end{minipage}
    }

    \vspace{2pt}

    \begin{subfigure}{\subfigwidth\linewidth}
        \includegraphics[width=\linewidth]{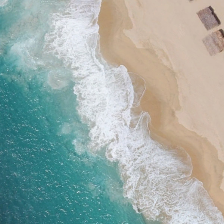}
        \label{subfigure:aidml_idx_67_original1}
    \end{subfigure}
    \hspace{0.1cm}
    \begin{subfigure}{\subfigwidth\linewidth}
        \includegraphics[width=\linewidth]{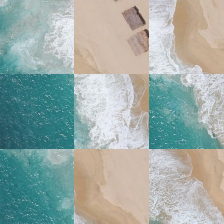}
        \label{subfigure:aidml_idx_67_patch_shuffle}
    \end{subfigure}
    \hspace{0.1cm}
    \begin{subfigure}{\subfigwidth\linewidth}
        \includegraphics[width=\linewidth]{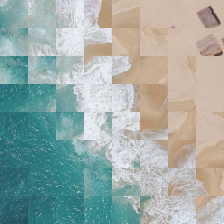}
        \label{subfigure:aidml_idx_67_patch_rotation}
    \end{subfigure}

  \vspace{-10pt}
  
    \makebox[\linewidth][c]{%
        \begin{minipage}{\subfigwidth\linewidth}
            \centering\scriptsize\textbf{(d) Original}
        \end{minipage}
        \hspace{0.1cm}
        \begin{minipage}{\subfigwidth\linewidth}
            \centering\scriptsize\textbf{(e) Texture (Bilateral Filter)}
        \end{minipage}
        \hspace{0.1cm}
        \begin{minipage}{\subfigwidth\linewidth}
            \centering\scriptsize\textbf{(f) Texture (Gaussian Blur)}
        \end{minipage}
    }

    \vspace{2pt}

    \begin{subfigure}{\subfigwidth\linewidth}
        \includegraphics[width=\linewidth]{images/visual_examples/aidml_idx_67_original.png}
    \label{subfigure:aidml_idx_67_original2}
    \end{subfigure}
    \hspace{0.1cm}
    \begin{subfigure}{\subfigwidth\linewidth}
        \includegraphics[width=\linewidth]{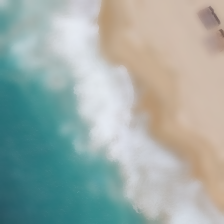}
        \label{subfigure:aidml_idx_67_bilateral}
    \end{subfigure}
    \hspace{0.1cm}
    \begin{subfigure}{\subfigwidth\linewidth}
        \includegraphics[width=\linewidth]{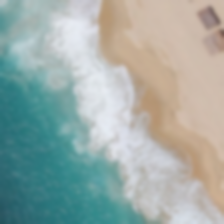}
        \label{subfigure:aidml_idx_67_gaussian}
    \end{subfigure}

   \vspace{-10pt}

    \makebox[\linewidth][c]{%
        \begin{minipage}{\subfigwidth\linewidth}
            \centering\scriptsize\textbf{(g) Original}
        \end{minipage}
        \hspace{0.1cm}
        \begin{minipage}{\subfigwidth\linewidth}
            \centering\scriptsize\textbf{(h) Color (Grayscale)}
        \end{minipage}
        \hspace{0.1cm}
        \begin{minipage}{\subfigwidth\linewidth}
            \centering\scriptsize\textbf{(i) Color (Channel Shuffle)}
        \end{minipage}
    }

    \vspace{2pt}

    \begin{subfigure}{\subfigwidth\linewidth}
        \includegraphics[width=\linewidth]{images/visual_examples/aidml_idx_67_original.png}
        \label{subfigure:aidml_idx_67_original3}
    \end{subfigure}
    \hspace{0.1cm}
    \begin{subfigure}{\subfigwidth\linewidth}
        \includegraphics[width=\linewidth]{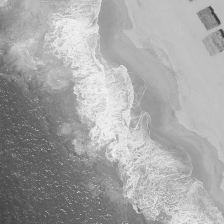}
        \label{subfigure:aidml_idx_67_grayscale}
    \end{subfigure}
    \hspace{0.1cm}
    \begin{subfigure}{\subfigwidth\linewidth}
        \includegraphics[width=\linewidth]{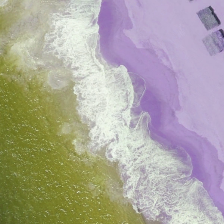}
        \label{subfigure:aidml_idx_67_channel_shuffle}
    \end{subfigure}
    \vspace{-5pt}
    \caption{Visual illustration of feature suppression transformations applied to a sample image from the AID training set belonging to the class farmland. \textbf{(a, d, g)} Show the original image. \textbf{(b)} Global shape suppression via Patch Shuffle with grid size 3. \textbf{(c)} Local shape suppression via Patch Rotation with grid size 8. \textbf{(e)} Texture suppression using Bilateral Filtering with \( d = 12 \), \( \sigma_{\text{color}} = 170 \), and \( \sigma_{\text{space}} = 75 \). \textbf{(f)} Texture suppression using Gaussian Blur with kernel size \( k = 11 \) and standard deviation \( \sigma = 2.0 \). \textbf{(h)} Color suppression via grayscale conversion. \textbf{(i)} Color suppression via random channel shuffle.}
    \label{fig:visual_example_3}
\end{figure}

\begin{figure}[t!]
    \def\subfigwidth{.31}
    \centering

    \makebox[\linewidth][c]{%
        \begin{minipage}{\subfigwidth\linewidth}
            \centering\scriptsize\textbf{(a) Original}
        \end{minipage}
        \hspace{0.1cm}
        \begin{minipage}{\subfigwidth\linewidth}
            \centering\scriptsize\textbf{(b) Global Shape (Patch Shuffle)}
        \end{minipage}
        \hspace{0.1cm}
        \begin{minipage}{\subfigwidth\linewidth}
            \centering\scriptsize\textbf{(c) Local Shape (Patch Rotation)}
        \end{minipage}
    }

    \vspace{2pt}

    \begin{subfigure}{\subfigwidth\linewidth}
        \includegraphics[width=\linewidth]{images/visual_examples/aidml_idx_27_original.png}
        \label{subfigure:aidml_idx_27_original1}
    \end{subfigure}
    \hspace{0.1cm}
    \begin{subfigure}{\subfigwidth\linewidth}
        \includegraphics[width=\linewidth]{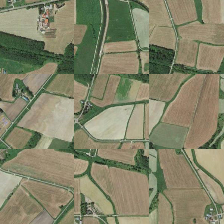}
        \label{subfigure:aidml_idx_27_patch_shuffle}
    \end{subfigure}
    \hspace{0.1cm}
    \begin{subfigure}{\subfigwidth\linewidth}
        \includegraphics[width=\linewidth]{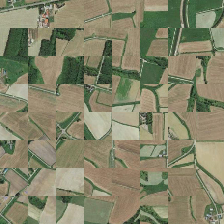}
        \label{subfigure:aidml_idx_27_patch_rotation}
    \end{subfigure}

  \vspace{-10pt}
  
    \makebox[\linewidth][c]{%
        \begin{minipage}{\subfigwidth\linewidth}
            \centering\scriptsize\textbf{(d) Original}
        \end{minipage}
        \hspace{0.1cm}
        \begin{minipage}{\subfigwidth\linewidth}
            \centering\scriptsize\textbf{(e) Texture (Bilateral Filter)}
        \end{minipage}
        \hspace{0.1cm}
        \begin{minipage}{\subfigwidth\linewidth}
            \centering\scriptsize\textbf{(f) Texture (Gaussian Blur)}
        \end{minipage}
    }

    \vspace{2pt}

    \begin{subfigure}{\subfigwidth\linewidth}
        \includegraphics[width=\linewidth]{images/visual_examples/aidml_idx_27_original.png}
    \label{subfigure:aidml_idx_27_original2}
    \end{subfigure}
    \hspace{0.1cm}
    \begin{subfigure}{\subfigwidth\linewidth}
        \includegraphics[width=\linewidth]{images/visual_examples/aidml_idx_27_bilateral_d12_c170.png}
        \label{subfigure:aidml_idx_27_bilateral}
    \end{subfigure}
    \hspace{0.1cm}
    \begin{subfigure}{\subfigwidth\linewidth}
        \includegraphics[width=\linewidth]{images/visual_examples/aidml_idx_27_gaussian_k11_s2.0.png}
        \label{subfigure:aidml_idx_27_gaussian}
    \end{subfigure}

   \vspace{-10pt}

    \makebox[\linewidth][c]{%
        \begin{minipage}{\subfigwidth\linewidth}
            \centering\scriptsize\textbf{(g) Original}
        \end{minipage}
        \hspace{0.1cm}
        \begin{minipage}{\subfigwidth\linewidth}
            \centering\scriptsize\textbf{(h) Color (Grayscale)}
        \end{minipage}
        \hspace{0.1cm}
        \begin{minipage}{\subfigwidth\linewidth}
            \centering\scriptsize\textbf{(i) Color (Channel Shuffle)}
        \end{minipage}
    }

    \vspace{2pt}

    \begin{subfigure}{\subfigwidth\linewidth}
        \includegraphics[width=\linewidth]{images/visual_examples/aidml_idx_27_original.png}
        \label{subfigure:aidml_idx_27_original3}
    \end{subfigure}
    \hspace{0.1cm}
    \begin{subfigure}{\subfigwidth\linewidth}
        \includegraphics[width=\linewidth]{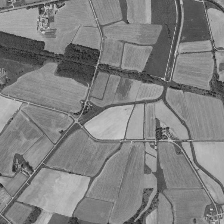}
        \label{subfigure:aidml_idx_27_grayscale}
    \end{subfigure}
    \hspace{0.1cm}
    \begin{subfigure}{\subfigwidth\linewidth}
        \includegraphics[width=\linewidth]{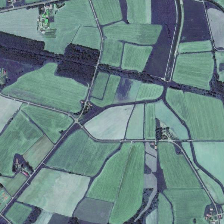}
        \label{subfigure:aidml_idx_27_channel_shuffle}
    \end{subfigure}
    \vspace{-5pt}
    \caption{Visual illustration of feature suppression transformations applied to a sample image from the AID training set belonging to the class beach. \textbf{(a, d, g)} Show the original image. \textbf{(b)} Global shape suppression via Patch Shuffle with grid size 3. \textbf{(c)} Local shape suppression via Patch Rotation with grid size 8. \textbf{(e)} Texture suppression using Bilateral Filtering with \( d = 12 \), \( \sigma_{\text{color}} = 170 \), and \( \sigma_{\text{space}} = 75 \). \textbf{(f)} Texture suppression using Gaussian Blur with kernel size \( k = 11 \) and standard deviation \( \sigma = 2.0 \). \textbf{(h)} Color suppression via grayscale conversion. \textbf{(i)} Color suppression via random channel shuffle.}
    \label{fig:visual_example_4}
\end{figure}

\begin{figure}[t!]
    \def\subfigwidth{.31}
    \centering

    \makebox[\linewidth][c]{%
        \begin{minipage}{\subfigwidth\linewidth}
            \centering\scriptsize\textbf{(a) Original}
        \end{minipage}
        \hspace{0.1cm}
        \begin{minipage}{\subfigwidth\linewidth}
            \centering\scriptsize\textbf{(b) Global Shape (Patch Shuffle)}
        \end{minipage}
        \hspace{0.1cm}
        \begin{minipage}{\subfigwidth\linewidth}
            \centering\scriptsize\textbf{(c) Local Shape (Patch Rotation)}
        \end{minipage}
    }

    \vspace{2pt}

    \begin{subfigure}{\subfigwidth\linewidth}
        \includegraphics[width=\linewidth]{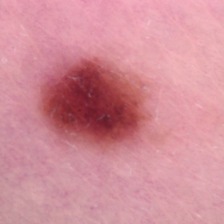}
        \label{subfigure:dermamnist_idx_2_original1}
    \end{subfigure}
    \hspace{0.1cm}
    \begin{subfigure}{\subfigwidth\linewidth}
        \includegraphics[width=\linewidth]{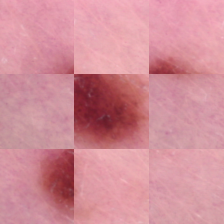}
        \label{subfigure:dermamnist_idx_2_patch_shuffle}
    \end{subfigure}
    \hspace{0.1cm}
    \begin{subfigure}{\subfigwidth\linewidth}
        \includegraphics[width=\linewidth]{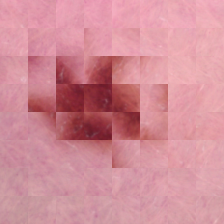}
        \label{subfigure:dermamnist_idx_2_patch_rotation}
    \end{subfigure}

  \vspace{-10pt}
  
    \makebox[\linewidth][c]{%
        \begin{minipage}{\subfigwidth\linewidth}
            \centering\scriptsize\textbf{(d) Original}
        \end{minipage}
        \hspace{0.1cm}
        \begin{minipage}{\subfigwidth\linewidth}
            \centering\scriptsize\textbf{(e) Texture (Bilateral Filter)}
        \end{minipage}
        \hspace{0.1cm}
        \begin{minipage}{\subfigwidth\linewidth}
            \centering\scriptsize\textbf{(f) Texture (Gaussian Blur)}
        \end{minipage}
    }

    \vspace{2pt}

    \begin{subfigure}{\subfigwidth\linewidth}
        \includegraphics[width=\linewidth]{images/visual_examples/dermamnist_idx_2_original.png}
    \label{subfigure:dermamnist_idx_2_original2}
    \end{subfigure}
    \hspace{0.1cm}
    \begin{subfigure}{\subfigwidth\linewidth}
        \includegraphics[width=\linewidth]{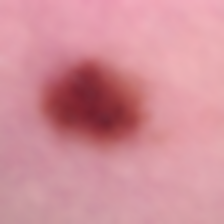}
        \label{subfigure:dermamnist_idx_2_bilateral}
    \end{subfigure}
    \hspace{0.1cm}
    \begin{subfigure}{\subfigwidth\linewidth}
        \includegraphics[width=\linewidth]{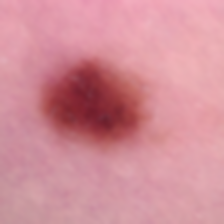}
        \label{subfigure:dermamnist_idx_2_gaussian}
    \end{subfigure}

   \vspace{-10pt}

    \makebox[\linewidth][c]{%
        \begin{minipage}{\subfigwidth\linewidth}
            \centering\scriptsize\textbf{(g) Original}
        \end{minipage}
        \hspace{0.1cm}
        \begin{minipage}{\subfigwidth\linewidth}
            \centering\scriptsize\textbf{(h) Color (Grayscale)}
        \end{minipage}
        \hspace{0.1cm}
        \begin{minipage}{\subfigwidth\linewidth}
            \centering\scriptsize\textbf{(i) Color (Channel Shuffle)}
        \end{minipage}
    }

    \vspace{2pt}

    \begin{subfigure}{\subfigwidth\linewidth}
        \includegraphics[width=\linewidth]{images/visual_examples/dermamnist_idx_2_original.png}
        \label{subfigure:dermamnist_idx_2_original3}
    \end{subfigure}
    \hspace{0.1cm}
    \begin{subfigure}{\subfigwidth\linewidth}
        \includegraphics[width=\linewidth]{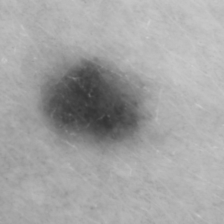}
        \label{subfigure:dermamnist_idx_2_grayscale}
    \end{subfigure}
    \hspace{0.1cm}
    \begin{subfigure}{\subfigwidth\linewidth}
        \includegraphics[width=\linewidth]{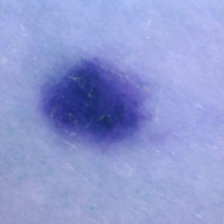}
        \label{subfigure:dermamnist_idx_2_channel_shuffle}
    \end{subfigure}
    \vspace{-5pt}
    \caption{Visual illustration of feature suppression transformations applied to a sample image from the DermaMNIST training set belonging to class melanocytic nevi. \textbf{(a, d, g)} Show the original image. \textbf{(b)} Global shape suppression via Patch Shuffle with grid size 3. \textbf{(c)} Local shape suppression via Patch Rotation with grid size 8. \textbf{(e)} Texture suppression using Bilateral Filtering with \( d = 12 \), \( \sigma_{\text{color}} = 170 \), and \( \sigma_{\text{space}} = 75 \). \textbf{(f)} Texture suppression using Gaussian Blur with kernel size \( k = 11 \) and standard deviation \( \sigma = 2.0 \). \textbf{(h)} Color suppression via grayscale conversion. \textbf{(i)} Color suppression via random channel shuffle.}
    \label{fig:visual_example_5}
\end{figure}

\begin{figure}[t!]
    \def\subfigwidth{.31}
    \centering

    \makebox[\linewidth][c]{%
        \begin{minipage}{\subfigwidth\linewidth}
            \centering\scriptsize\textbf{(a) Original}
        \end{minipage}
        \hspace{0.1cm}
        \begin{minipage}{\subfigwidth\linewidth}
            \centering\scriptsize\textbf{(b) Global Shape (Patch Shuffle)}
        \end{minipage}
        \hspace{0.1cm}
        \begin{minipage}{\subfigwidth\linewidth}
            \centering\scriptsize\textbf{(c) Local Shape (Patch Rotation)}
        \end{minipage}
    }

    \vspace{2pt}

    \begin{subfigure}{\subfigwidth\linewidth}
        \includegraphics[width=\linewidth]{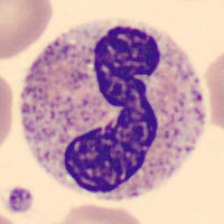}
        \label{subfigure:bloodmnist_idx_3_original1}
    \end{subfigure}
    \hspace{0.1cm}
    \begin{subfigure}{\subfigwidth\linewidth}
        \includegraphics[width=\linewidth]{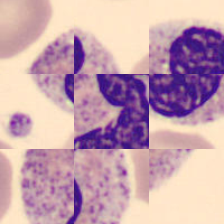}
        \label{subfigure:bloodmnist_idx_3_patch_shuffle}
    \end{subfigure}
    \hspace{0.1cm}
    \begin{subfigure}{\subfigwidth\linewidth}
        \includegraphics[width=\linewidth]{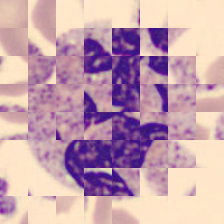}
        \label{subfigure:bloodmnist_idx_3_patch_rotation}
    \end{subfigure}

  \vspace{-10pt}
  
    \makebox[\linewidth][c]{%
        \begin{minipage}{\subfigwidth\linewidth}
            \centering\scriptsize\textbf{(d) Original}
        \end{minipage}
        \hspace{0.1cm}
        \begin{minipage}{\subfigwidth\linewidth}
            \centering\scriptsize\textbf{(e) Texture (Bilateral Filter)}
        \end{minipage}
        \hspace{0.1cm}
        \begin{minipage}{\subfigwidth\linewidth}
            \centering\scriptsize\textbf{(f) Texture (Gaussian Blur)}
        \end{minipage}
    }

    \vspace{2pt}

    \begin{subfigure}{\subfigwidth\linewidth}
        \includegraphics[width=\linewidth]{images/visual_examples/bloodmnist_idx_3_original.png}
    \label{subfigure:bloodmnist_idx_3_original2}
    \end{subfigure}
    \hspace{0.1cm}
    \begin{subfigure}{\subfigwidth\linewidth}
        \includegraphics[width=\linewidth]{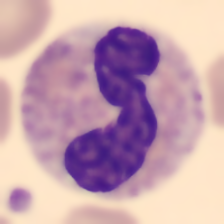}
        \label{subfigure:bloodmnist_idx_3_bilateral}
    \end{subfigure}
    \hspace{0.1cm}
    \begin{subfigure}{\subfigwidth\linewidth}
        \includegraphics[width=\linewidth]{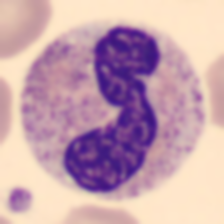}
        \label{subfigure:bloodmnist_idx_3_gaussian}
    \end{subfigure}

   \vspace{-10pt}

    \makebox[\linewidth][c]{%
        \begin{minipage}{\subfigwidth\linewidth}
            \centering\scriptsize\textbf{(g) Original}
        \end{minipage}
        \hspace{0.1cm}
        \begin{minipage}{\subfigwidth\linewidth}
            \centering\scriptsize\textbf{(h) Color (Grayscale)}
        \end{minipage}
        \hspace{0.1cm}
        \begin{minipage}{\subfigwidth\linewidth}
            \centering\scriptsize\textbf{(i) Color (Channel Shuffle)}
        \end{minipage}
    }

    \vspace{2pt}

    \begin{subfigure}{\subfigwidth\linewidth}
        \includegraphics[width=\linewidth]{images/visual_examples/bloodmnist_idx_3_original.png}
        \label{subfigure:bloodmnist_idx_3_original3}
    \end{subfigure}
    \hspace{0.1cm}
    \begin{subfigure}{\subfigwidth\linewidth}
        \includegraphics[width=\linewidth]{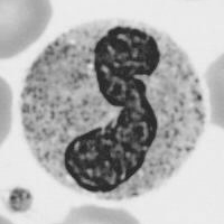}
        \label{subfigure:bloodmnist_idx_3_grayscale}
    \end{subfigure}
    \hspace{0.1cm}
    \begin{subfigure}{\subfigwidth\linewidth}
        \includegraphics[width=\linewidth]{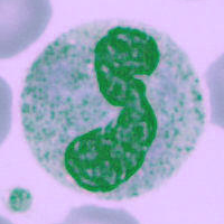}
        \label{subfigure:bloodmnist_idx_3_channel_shuffle}
    \end{subfigure}
    \vspace{-5pt}
    \caption{Visual illustration of feature suppression transformations applied to a sample image from the BloodMNIST train set belonging to the class neutrophil. \textbf{(a, d, g)} Show the original image. \textbf{(b)} Global shape suppression via Patch Shuffle with grid size 3. \textbf{(c)} Local shape suppression via Patch Rotation with grid size 8. \textbf{(e)} Texture suppression using Bilateral Filtering with \( d = 12 \), \( \sigma_{\text{color}} = 170 \), and \( \sigma_{\text{space}} = 75 \). \textbf{(f)} Texture suppression using Gaussian Blur with kernel size \( k = 11 \) and standard deviation \( \sigma = 2.0 \). \textbf{(h)} Color suppression via grayscale conversion. \textbf{(i)} Color suppression via random channel shuffle.}
    \label{fig:visual_example_6}
\end{figure}

Visual examples of the applied feature suppression transformations are provided in \Cref{fig:visual_example_1} and \Cref{fig:visual_example_2}, illustrating images from the entry-level categories dog and bird in the ImageNet validation set. Additional examples for the RS domain are shown in \Cref{fig:visual_example_3} and \Cref{fig:visual_example_4}, corresponding to the classes farmland and beach from the AID dataset. For the MI domain, \Cref{fig:visual_example_5} and \Cref{fig:visual_example_6} present examples from the classes melanocytic nevi in DermaMNIST and neutrophil in BloodMNIST.

\section{Ablation Study for Suppression Effects of Transformations}\label{sec:ablation_supp_effects}

To validate the robustness of our metric-based evaluation, we ablate the kernel size for all texture suppression transformations across the same 800 ImageNet images from the validation set. In this extended analysis, we include Bilateral filtering, Gaussian blur, Non-Local Means Denoising (denoted as NLMeans), Box blur, and Median filtering. For transformations with two parameters (e.g., $\sigma$ and $k$ for Gaussian blur), we vary both parameters along a diagonal correspondence e.g., $(\sigma, k) = (0.66, 5), (1.0, 7), \dots, (2.33, 15)$ for Gaussian blur, and $(\sigma_c, k) = (50, 5), (80, 7), \dots, (200, 15)$ for Bilateral filtering. For Non-Local Means Denoising, we use a slightly offset mapping: $(h, k) = (5, 5), (5, 7), (10, 9), \dots, (25, 15)$ where $k$ is used as template window size.

\begin{figure}[t!]
    \def\subfigwidth{.45}
    \centering

    \makebox[\linewidth][c]{%
        \begin{minipage}{\subfigwidth\linewidth}
            \centering\scriptsize\textbf{(a) Texture Suppression}
        \end{minipage}
        \hspace{0.1cm}
        \begin{minipage}{\subfigwidth\linewidth}
            \centering\scriptsize\textbf{(b) Shape Preservation}
        \end{minipage}
    }

    \vspace{2pt}

    \begin{subfigure}{\subfigwidth\linewidth}
        \includegraphics[width=\linewidth]{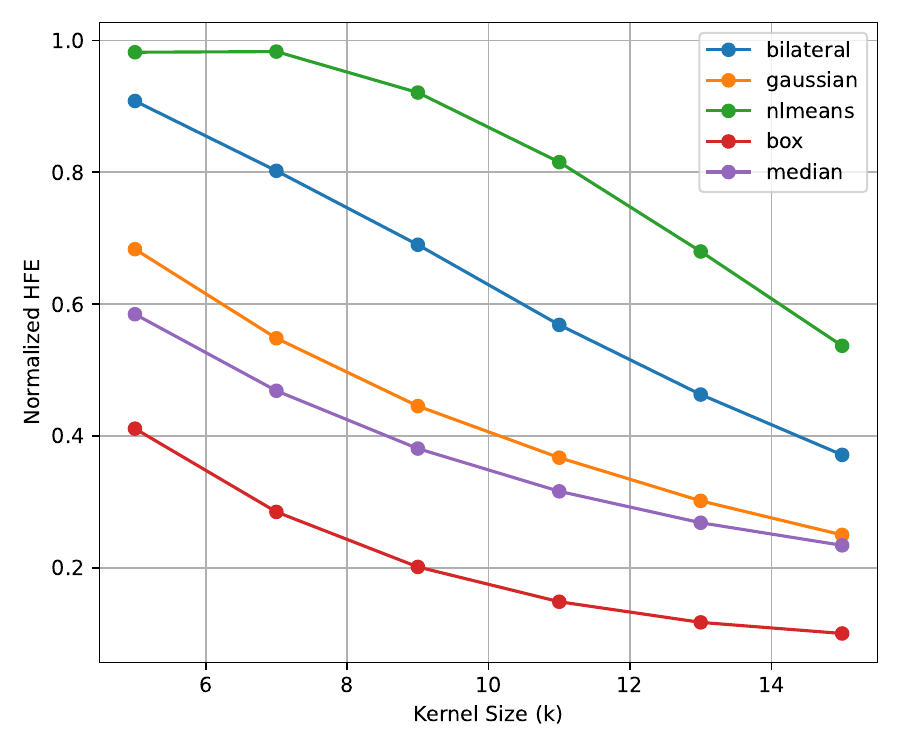}
        \label{subfigure:texture_suppression}
    \end{subfigure}
    \hspace{0.1cm}
    \begin{subfigure}{\subfigwidth\linewidth}
        \includegraphics[width=\linewidth]{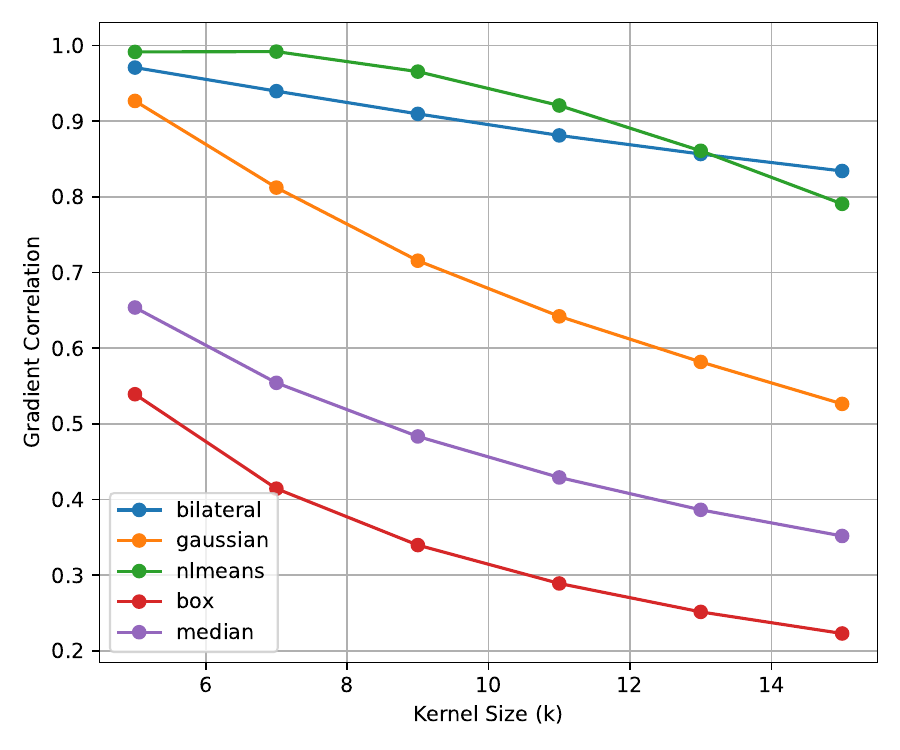}
        \label{subfigure:edge_preservation}
    \end{subfigure}

    \vspace{-5pt}
    \caption{Ablation of kernel parameter for quantitative evaluation of feature suppression transformations. \textbf{(a)} Normalized high-frequency energy quantifies texture removal. \textbf{(b)} Gradient correlation reflects shape preservation.}
    \label{fig:filter_ablation}
    \vspace{-10pt}
\end{figure}

\Cref{fig:filter_ablation} presents the effects of these transformations on normalized high-frequency energy (HFE) and gradient correlation (GC). Bilateral filtering achieves the most favorable trade-off, consistently reducing HFE while preserving edge structure to a high degree across kernel sizes. Gaussian blur suppresses texture even more strongly than Bilateral filtering, but still maintains a reasonable level of shape preservation, making it a competitive alternative when stronger smoothing is required. In contrast, Median and Box blur aggressively reduce texture but at the expense of substantial degradation in edge information. Non-Local Means Denoising preserves structural information well but is comparatively less effective at suppressing texture.

\section{Experimental Details for Experiment Human vs. CNNs}
\label{sec:experimental_details}

\subsection{Participants Instruction}

\textbf{Participant Instructions}

Thank you for participating in our visual perception study. This study investigates how humans recognize objects when certain visual features are suppressed. Please read the following instructions carefully before beginning.

\textbf{General Information}

You will be shown a series of images, each belonging to one of 16 everyday object categories (e.g., cat, car, airplane). In each trial, your task is to classify the object as accurately as possible.

\textbf{Trial Procedure}

Each trial will proceed as follows:

\begin{enumerate}[label=(\arabic*)]
    \item A small black fixation square will appear in the center of the screen for 300 milliseconds. Please focus your gaze on it.
    \item An image will be shown for 200 milliseconds. The image may appear altered (e.g., blurred, gray, or shuffled).
    \item Immediately after the image, a noise mask will appear briefly to reduce visual aftereffects.
\end{enumerate}

\textbf{Response Task}

After each image, you will see a $4 \times 4$ grid of category labels. Click on the label that best matches the object you saw. If you are unsure or could not recognize the object, you may select the “not clear” option. Each image will be shown only once. Please respond based on your first impression.

\textbf{Estimated Duration}

The study will take approximately 35–45 minutes. Please complete it in one sitting and avoid distractions.

\textbf{Participation and Data Protection}

Participation in this study is entirely voluntary. You may stop the study at any time without providing a reason and without any negative consequences. Before beginning the experiment, you will be asked to provide written informed consent. By doing so, you confirm that you understand the nature and purpose of the study and agree to participate. All data collected will be stored in an anonymized form and handled in accordance with institutional data protection policies. No personal identifying information will be published or shared.

By continuing, you confirm that you have read and understood the instructions.

\subsection{Attention Test}

Every 100 trials, an unannounced attention test was administered (in total 7). The participant was informed via a small text box about the attention test, in which the participant was informed that they would see an object of class A in the next image and that they had to click on a different class B to successfully pass it. Only if the attention test was correctly passed, we considered the results as valid.

\subsection{Screenshots}

\begin{figure}[t!]
    \def\subfigwidth{.31}
    \centering

    \makebox[\linewidth][c]{%
        \begin{minipage}{\subfigwidth\linewidth}
            \centering\scriptsize\textbf{(a) Start Screen}
        \end{minipage}
        \hspace{0.1cm}
        \begin{minipage}{\subfigwidth\linewidth}
            \centering\scriptsize\textbf{(b) Fixation Cross}
        \end{minipage}
        \hspace{0.1cm}
        \begin{minipage}{\subfigwidth\linewidth}
            \centering\scriptsize\textbf{(c) Image Stimuli}
        \end{minipage}
    }

    \vspace{2pt}

    \begin{subfigure}{\subfigwidth\linewidth}
        \includegraphics[width=\linewidth]{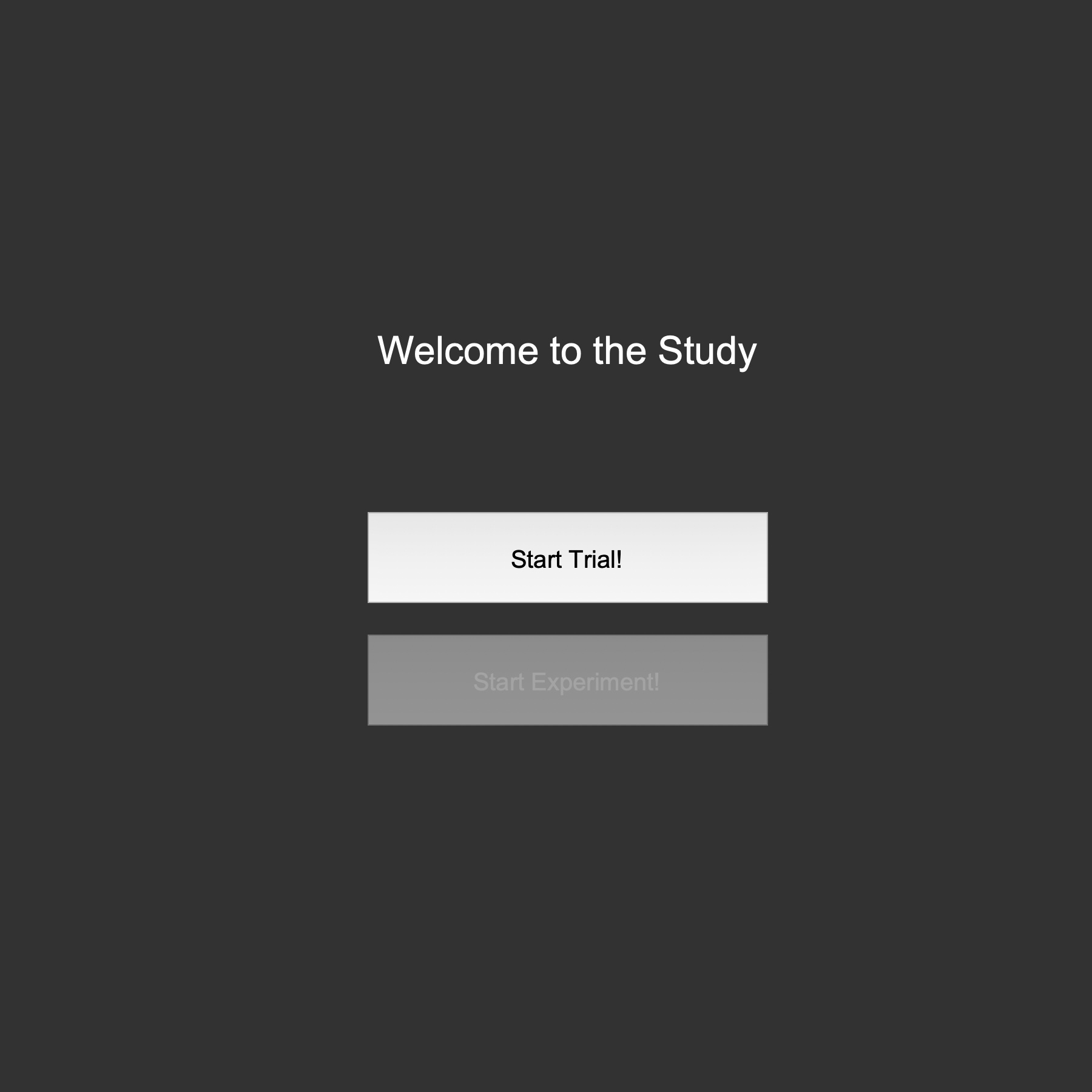}
        \label{subfigure:start_screen}
    \end{subfigure}
    \hspace{0.1cm}
    \begin{subfigure}{\subfigwidth\linewidth}
        \includegraphics[width=\linewidth]{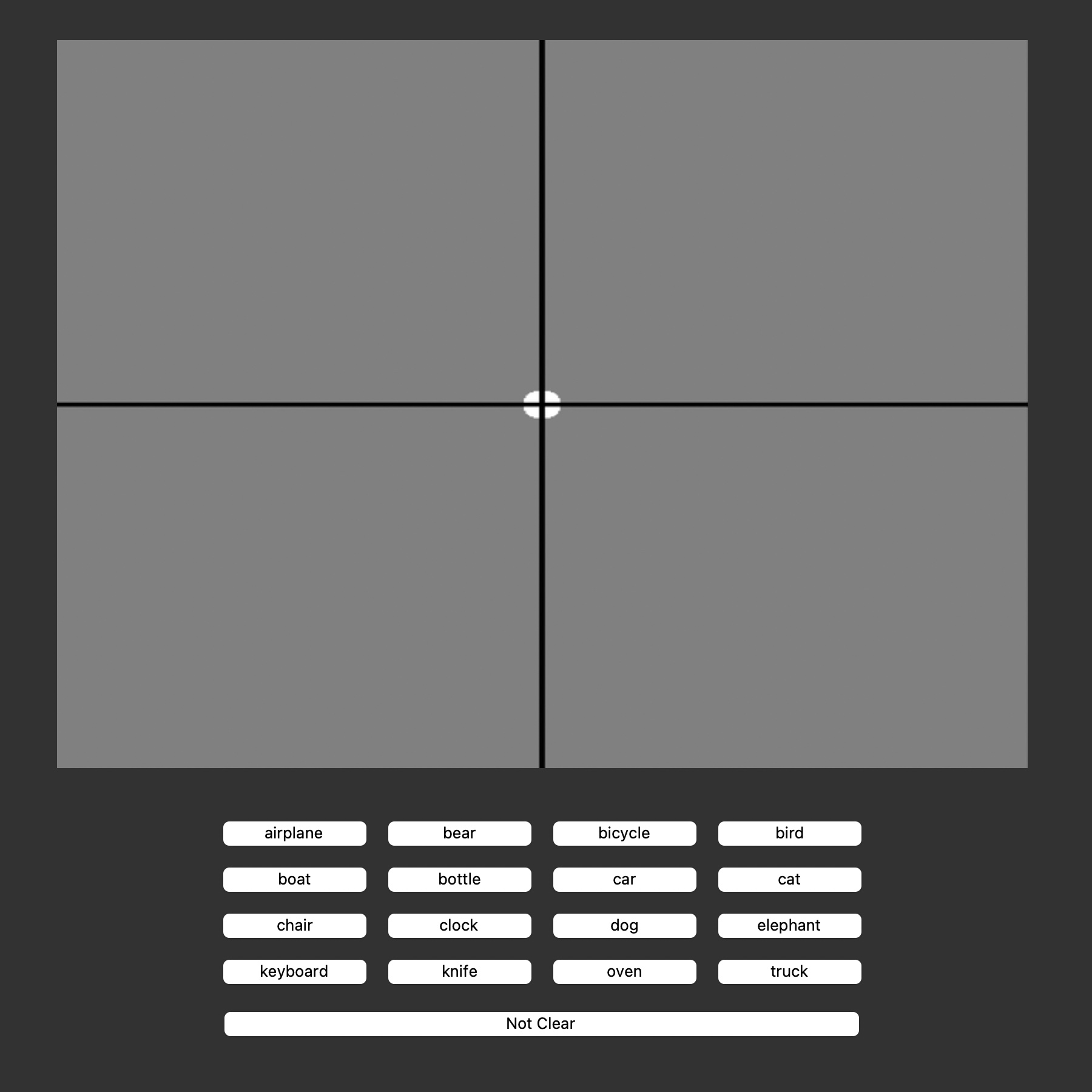}
        \label{subfigure:fixation_cross}
    \end{subfigure}
    \hspace{0.1cm}
    \begin{subfigure}{\subfigwidth\linewidth}
        \includegraphics[width=\linewidth]{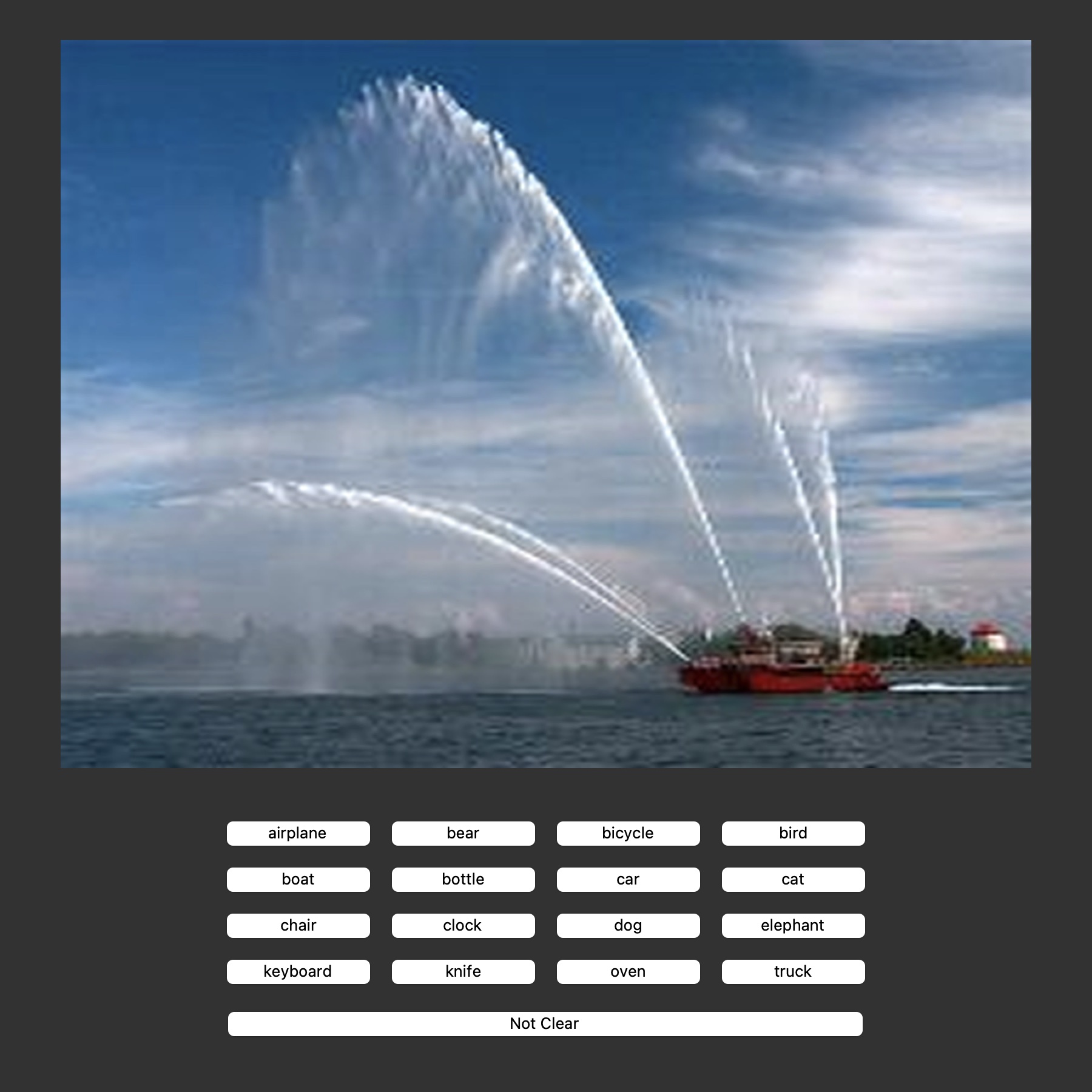}
        \label{subfigure:image_stimuli}
    \end{subfigure}

  \vspace{-10pt}
  
    \makebox[\linewidth][c]{%
        \begin{minipage}{\subfigwidth\linewidth}
            \centering\scriptsize\textbf{(d) Pink Noise}
        \end{minipage}
        \hspace{0.1cm}
        \begin{minipage}{\subfigwidth\linewidth}
            \centering\scriptsize\textbf{(e) Selection Screen}
        \end{minipage}
        \hspace{0.1cm}
        \begin{minipage}{\subfigwidth\linewidth}
            \centering\scriptsize\textbf{(f) Attention Test}
        \end{minipage}
    }

    \vspace{2pt}

    \begin{subfigure}{\subfigwidth\linewidth}
        \includegraphics[width=\linewidth]{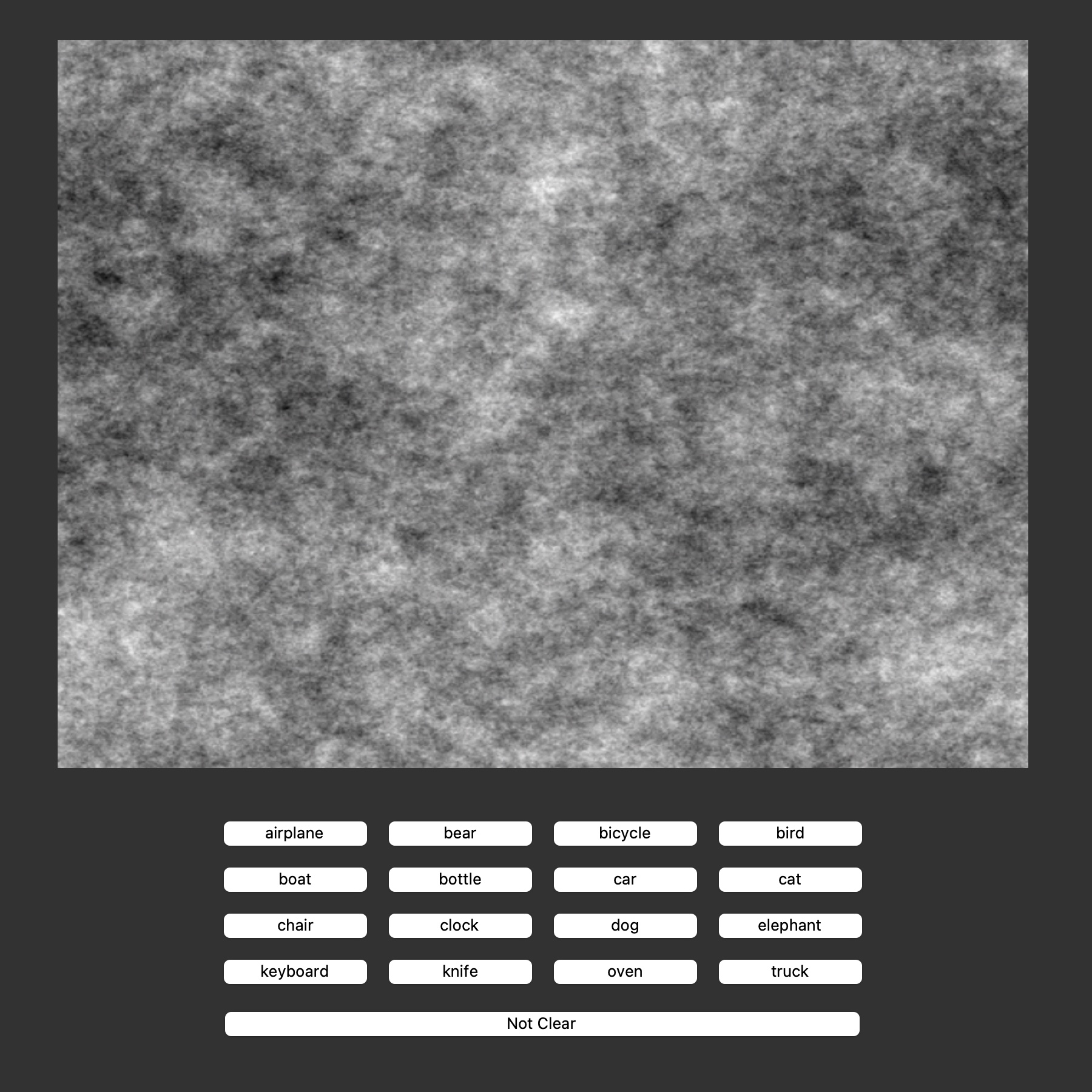}
        \label{subfigure:pink_noise}
    \end{subfigure}
    \hspace{0.1cm}
    \begin{subfigure}{\subfigwidth\linewidth}
        \includegraphics[width=\linewidth]{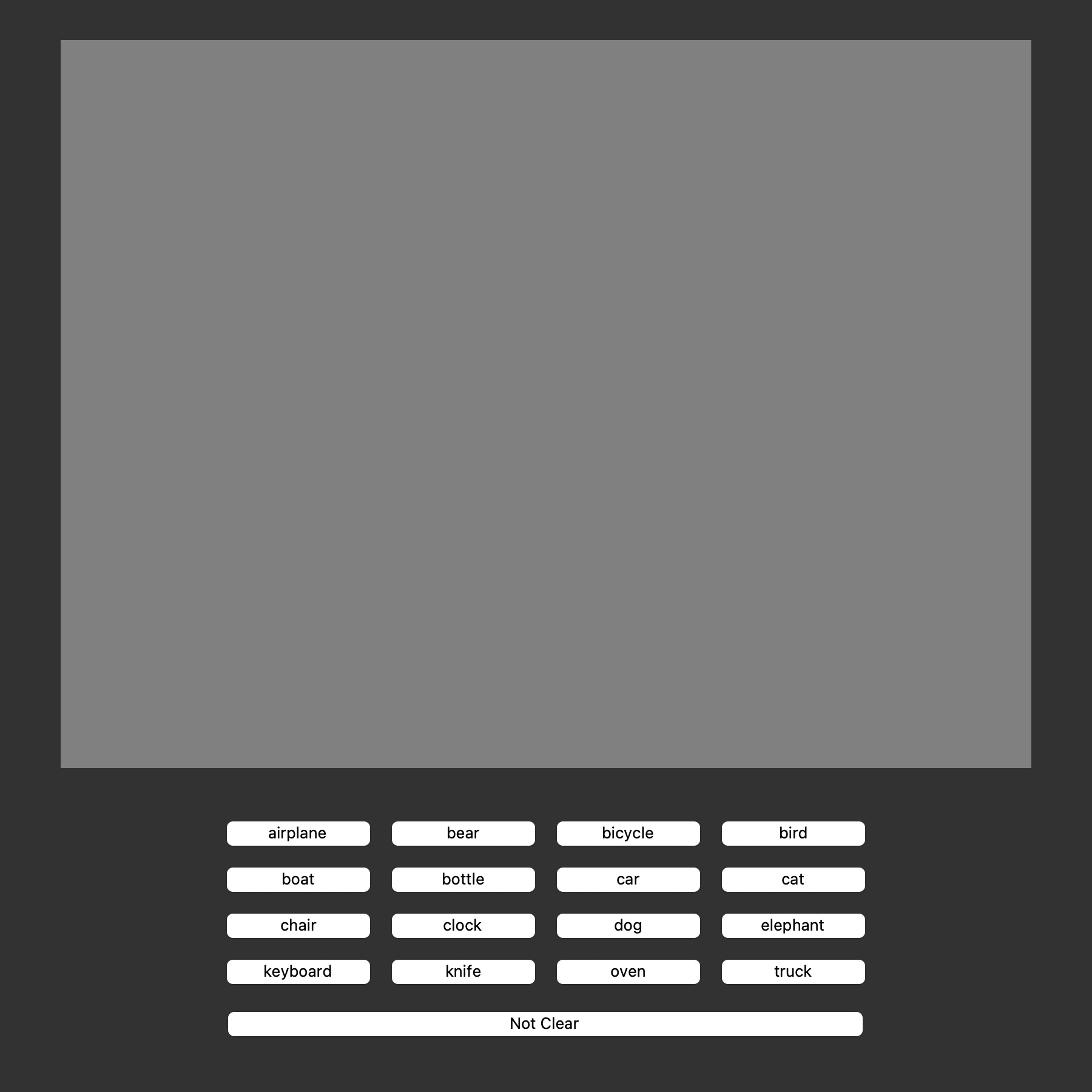}
        \label{subfigure:selection_screen}
    \end{subfigure}
    \hspace{0.1cm}
    \begin{subfigure}{\subfigwidth\linewidth}
        \includegraphics[width=\linewidth]{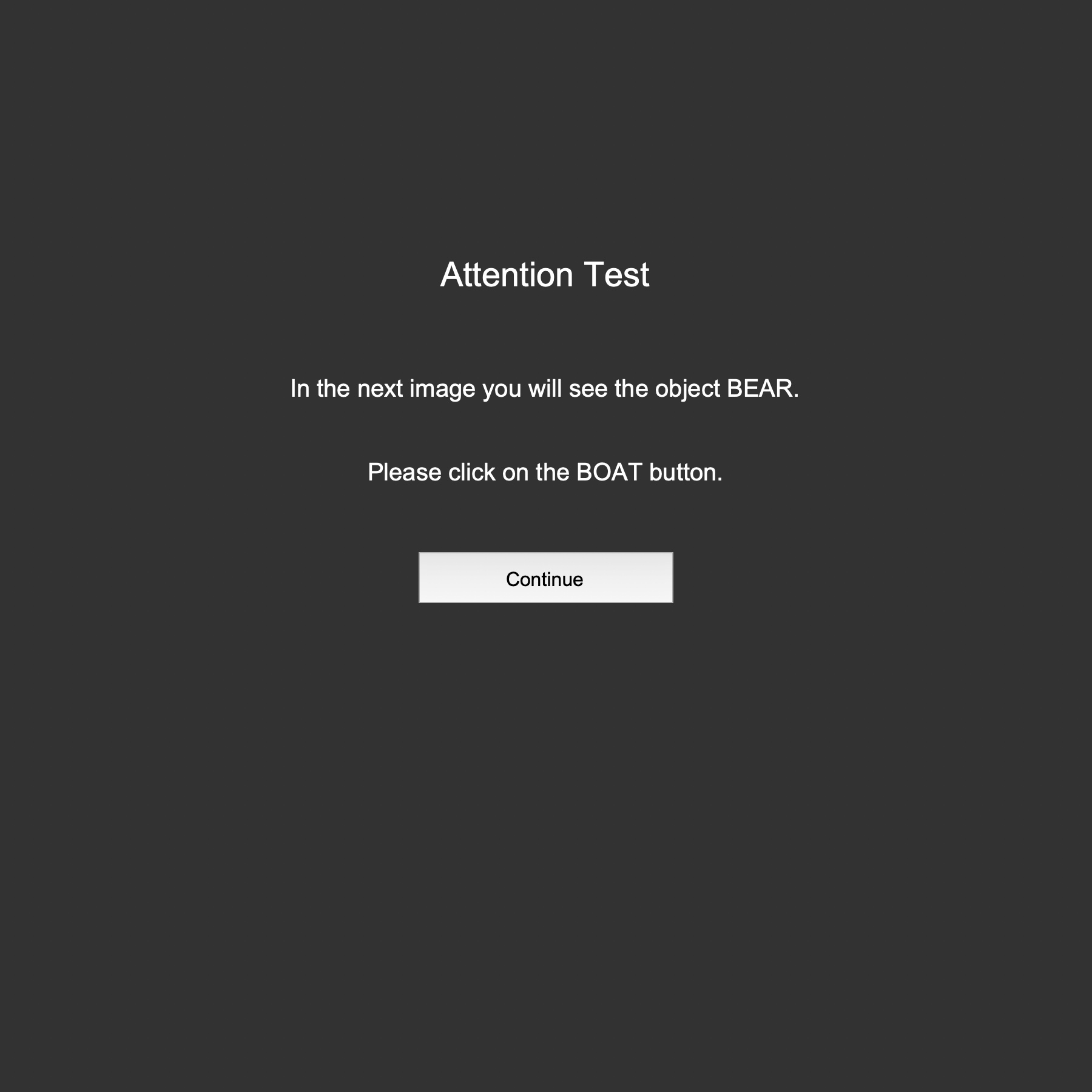}
        \label{subfigure:attention_test}
    \end{subfigure}
    \vspace{-5pt}
    \caption{Screenshots from the human study. \text{}}
    \label{fig:screenshots}
\end{figure}

In \Cref{fig:screenshots}, screenshots of the tool for conducting the human study are shown.

\section{Statistical Significance Test for Experiment Humans vs. CNNs}

Two-sided paired t-tests performed for the experiment in \Cref{sec:human_vs_cnns} reveal statistically significant differences between human and ResNet50-standard performance across all suppression conditions (p < 0.001). Effect sizes are large in all cases (Cohen’s d ranging from 8.69 to 39.83), confirming strong and systematic divergences in feature reliance (see \Cref{tab:t_test_results}). The t-tests are performed using the scipy library \cite{virtanen_scipy_2020}. 

Table \ref{tab:confidence_intervals} reports mean accuracies with 95\% confidence intervals for humans and the ResNet50-standard under each suppression condition, along with the corresponding human–model accuracy differences. To assess inter-subject and inter-model consistency, we further estimate the noise ceiling: the item-wise human noise ceiling is 0.542 $\pm$ 0.055, and the model noise ceiling is 0.646 $\pm$ 0.015.

\begin{table}[h]
\centering
\caption{Paired two-sided \textit{t}-test comparing human vs.\ ResNet50 accuracy under each suppression condition ($n=4$ per group). Cohen’s $d$ quantifies the standardized effect size.}
\label{tab:t_test_results}
\vspace{5pt}
\begin{tabular}{lcccc}
\toprule
\textbf{Suppression Type} & \textbf{\textit{t}-statistic} & \textbf{\textit{p}-value} & \textbf{Cohen’s $d$} & \textbf{Power} \\
\midrule
Global Shape  & 24.76 & 0.0001 & 12.38 & 1.000 \\
Local Shape   & 48.16 & $<$0.0001 & 24.08 & 1.000 \\
Texture       & 79.65 & $<$0.0001 & 39.83 & 1.000 \\
Color         & 17.38 & 0.0004 & 8.69 & 1.000 \\
\bottomrule
\end{tabular}
\end{table}

\begin{table}[h]
\centering
\caption{Human and ResNet50-standard accuracies under different feature suppression conditions. Values denote mean accuracies with 95\% confidence intervals (CI), and the rightmost column reports the human–model accuracy difference.}
\label{tab:confidence_intervals}
\vspace{5pt}
\begin{tabular}{lccc}
\toprule
\textbf{Suppression Type} & \textbf{Human Accuracy (CI)} & \textbf{ResNet Accuracy (CI)} & \textbf{Difference (CI)} \\
\midrule
Global Shape  & 0.965 (±0.0067) & 0.832 (±0.0102) & 0.133 (±0.0121) \\
Local Shape   & 0.763 (±0.0092) & 0.276 (±0.0141) & 0.487 (±0.0168) \\
Texture       & 0.979 (±0.0064) & 0.795 (±0.0073) & 0.184 (±0.0097) \\
Color         & 0.999 (±0.0052) & 0.924 (±0.0035) & 0.075 (±0.0063) \\
\bottomrule
\end{tabular}
\end{table}

\section{Comparison of Softmax Thresholding and Argmax Decision Rules}

In \Cref{tab:argmax_vs_threshold} we compare our heuristic decision rule used in the main experiments, which aggregates subclass probabilities via summed softmax and applies a 0.5 threshold, with a plain argmax rule. Argmax increases absolute accuracy slightly, yet the relative degradation patterns across suppression types and models remain stable, indicating that the main conclusions about feature reliance are robust to the choice of decision rule.

\begin{table}[t]
\centering
\caption{Comparison of softmax thresholding and argmax decision rules. Values report accuracy under each suppression condition and on original images for three representative models.}
\vspace{5pt}

\renewcommand{\tabularxcolumn}[1]{>{\centering\arraybackslash}m{#1}} 
\setlength{\tabcolsep}{4pt} 
\begin{tabularx}{\textwidth}{l*{4}{X} X}
\toprule
\textbf{Model variant} & \textbf{Global Shape} & \textbf{Local Shape} & \textbf{Texture} & \textbf{Color} & \textbf{Original} \\
\midrule
ResNet50-standard (Sum + Softmax > 0.5) & 0.832 & 0.276 & 0.795 & 0.924 & 0.954 \\
ResNet50-standard (Argmax)        & 0.880 & 0.361 & 0.840 & 0.944 & 0.962 \\
\midrule
ResNet50-sota (Sum + Softmax > 0.5)     & 0.943 & 0.618 & 0.867 & 0.948 & 0.931 \\
ResNet50-sota (Argmax)            & 0.949 & 0.665 & 0.904 & 0.965 & 0.940 \\
\midrule
ConvNeXtV2 (Sum + Softmax > 0.5)        & 0.949 & 0.647 & 0.925 & 0.969 & 0.940 \\
ConvNeXtV2 (Argmax)               & 0.979 & 0.748 & 0.956 & 0.984 & 0.944 \\
\bottomrule
\label{tab:argmax_vs_threshold}
\end{tabularx}
\end{table}

\section{Absolute Performance of Models trained on CV, MI and RS}

\Cref{tab:absolute_accuracies} summarizes the maximum validation performance of ResNet50 across all datasets and domains. Results are reported as macro and micro accuracy, separately for models trained from scratch and, where applicable, fine-tuned from ImageNet pretrained weights. For DeepGlobe, we report performance in mean average precision.

\begin{table*}[ht]
\centering
\caption{Maximum validation performance (macro and micro accuracy) of ResNet50 across domains, datasets, and training settings. For DeepGlobe, the performance metric is mean average precision.}
\label{tab:absolute_accuracies}
\setlength{\tabcolsep}{12pt} 
\begin{tabular}{lllccc}
\toprule
\multirow{2}{*}{\textbf{Domain}} &
\multirow{2}{*}{\textbf{Dataset}} &
\multirow{2}{*}{\textbf{Training Type}} &
 \textbf{Accuracy} & \textbf{Accuracy} \\
& & & \textbf{Macro} & \textbf{Micro} \\
\midrule
Computer Vision & Caltech101      & Pretrained   & 0.9523 & 0.9700 \\
Computer Vision & STL10           & Pretrained   & 0.9800 & 0.9800 \\
Computer Vision & OxfordIIITPet   & Pretrained   & 0.9404 & 0.9402 \\
Computer Vision & Flowers102      & Pretrained   & 0.9225 & 0.9225 \\
Computer Vision & ImageNet        & From Scratch & 0.7423 & 0.7423 \\
Computer Vision & Caltech101      & From Scratch & 0.7012 & 0.7972 \\
Computer Vision & STL10           & From Scratch & 0.7800 & 0.7800 \\
Computer Vision & OxfordIIITPet   & From Scratch & 0.6207 & 0.6214 \\
Computer Vision & Flowers102      & From Scratch & 0.5186 & 0.5186 \\
Medical Imaging & BloodMNIST      & From Scratch & 0.9868 & 0.9848 \\
Medical Imaging & DermaMNIST      & From Scratch & 0.5444 & 0.7906 \\
Medical Imaging & PathMNIST       & From Scratch & 0.9983 & 0.9984 \\
Medical Imaging & ChestMNIST      & From Scratch & 0.7048 & 0.7048 \\
Medical Imaging & RetinaMNIST     & From Scratch & 0.4446 & 0.5750 \\
Remote Sensing  & AID             & From Scratch & 0.8812 & 0.8847 \\
Remote Sensing  & PatternNet      & From Scratch & 0.9921 & 0.9921 \\
Remote Sensing  & RSD46WHU        & From Scratch & 0.8123 & 0.8177 \\
Remote Sensing  & UCMerced        & From Scratch & 0.9335 & 0.9335 \\
Remote Sensing  & DeepGlobe       & From Scratch & 0.8857 & 0.9295 \\
\bottomrule
\end{tabular}
\end{table*}

\section{Additional Experiments}
\subsection{Class-wise analysis for ImageNet in Experiment II}

\begin{figure}[t!]
    \def\subfigwidth{.31}
    \centering

    \makebox[\linewidth][c]{%
        \begin{minipage}{\subfigwidth\linewidth}
            \centering\scriptsize\textbf{(a) Local Shape and Texture}
        \end{minipage}
        \hspace{0.1cm}
        \begin{minipage}{\subfigwidth\linewidth}
            \centering\scriptsize\textbf{(b) Local Shape and Color}
        \end{minipage}
        \hspace{0.1cm}
        \begin{minipage}{\subfigwidth\linewidth}
            \centering\scriptsize\textbf{(c) Texture and Color}
        \end{minipage}
    }

    \vspace{2pt}

    \begin{subfigure}{\subfigwidth\linewidth}
        \includegraphics[width=\linewidth]{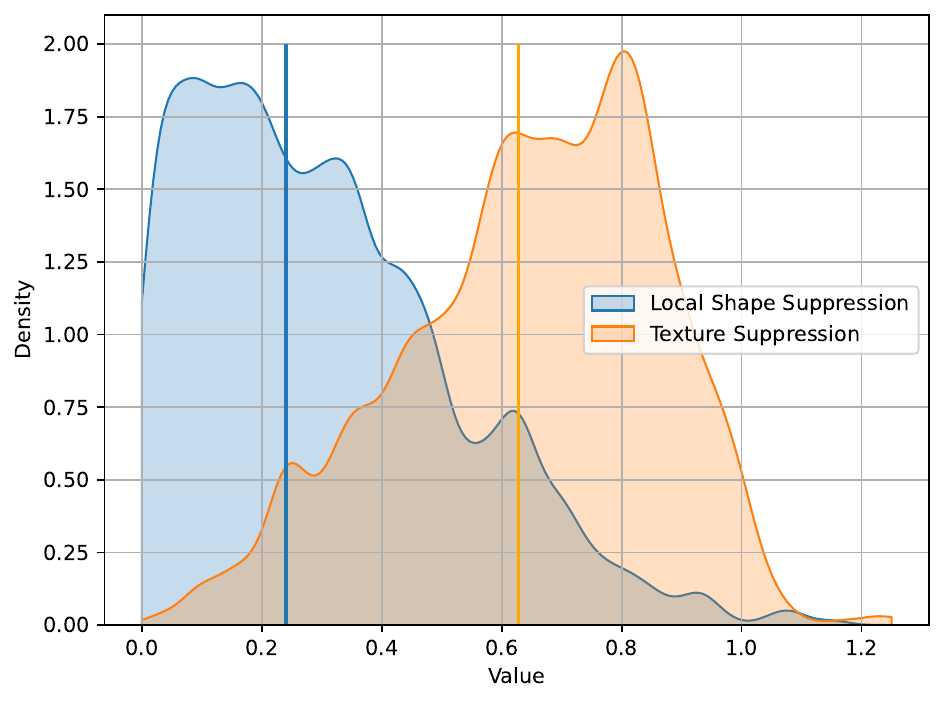}
        \label{subfigure:kde_shape_texture}
    \end{subfigure}
    \hspace{0.1cm}
    \begin{subfigure}{\subfigwidth\linewidth}
        \includegraphics[width=\linewidth]{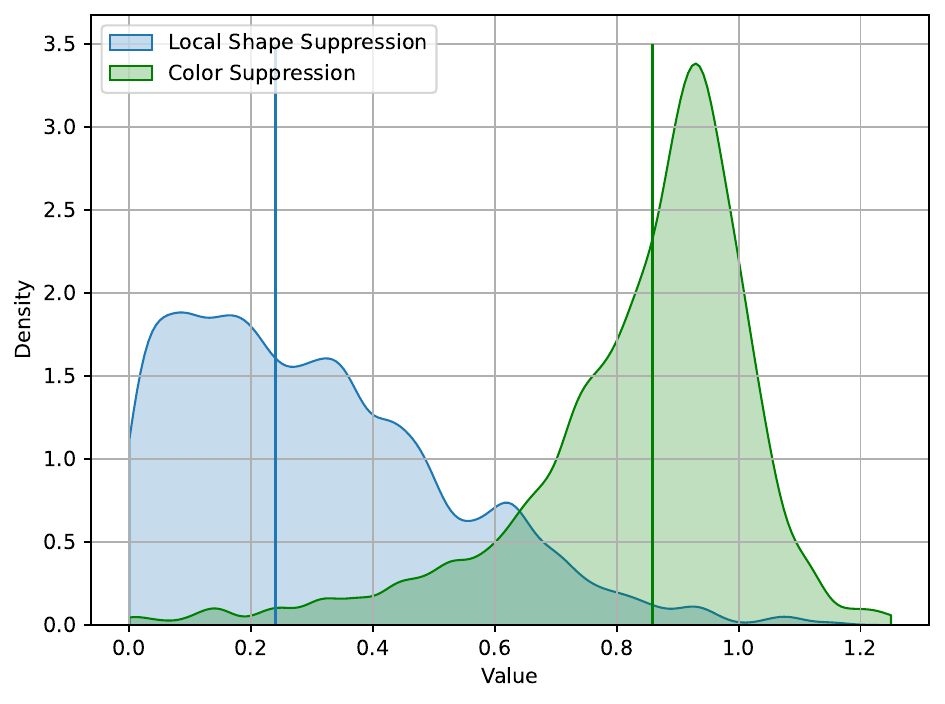}
        \label{subfigure:kde_shape_color}
    \end{subfigure}
    \hspace{0.1cm}
    \begin{subfigure}{\subfigwidth\linewidth}
        \includegraphics[width=\linewidth]{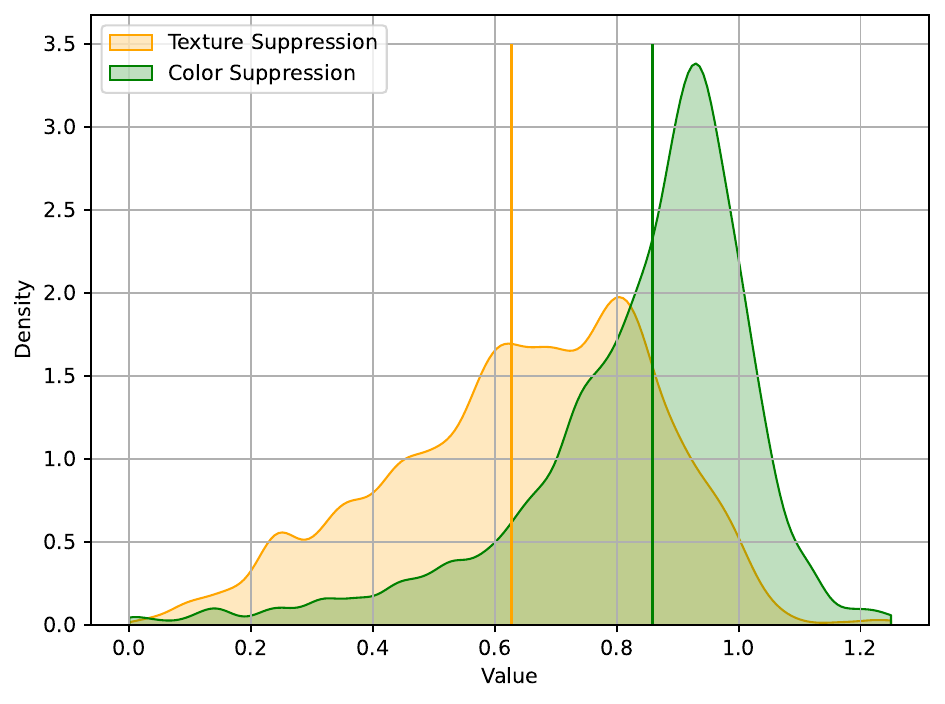}
        \label{subfigure:kde_texture_color}
    \end{subfigure}
    \vspace{-5pt}
    \caption{Kernel density estimates of class-wise relative accuracy under local shape, texture, and color suppression for ImageNet-1K.}
    \label{fig:imagenet_classwise}
    \vspace{-10pt}
\end{figure}

To examine whether the feature reliance patterns observed in Experiment II generalize across categories or are driven by a small subset of classes, we conduct a class-wise analysis on ImageNet-1K. We evaluate performance under local shape suppression (Patch Shuffle, grid size 6), texture suppression (bilateral filter, kernel size 12), and color suppression (grayscale). For each of the 1000 ImageNet classes, we computed relative accuracy under suppression and visualized the distributions using kernel density estimates (KDEs). The class-level distributions in \Cref{fig:imagenet_classwise} confirm the global trend: scores under texture suppression are clearly shifted toward higher values (mean = 0.62) compared to local shape suppression (mean = 0.24), with only modest overlap between the distributions (approximately 25–30\%).

To assess consistency across classes, we computed the percentage of categories following the general ranking of feature importance:

\begin{itemize}
    \item 88.0\% of classes of ImageNet show greater reliance on local shape than on texture,
    \item 94.2\% of classes of ImageNet show greater reliance on local shape than on color,
    \item 77.6\% of classes of ImageNet show greater reliance on texture than on color.
\end{itemize}

These results demonstrate that the observed reliance patterns hold broadly across the dataset rather than being driven by isolated outlier categories. To complement the quantitative analysis, we identified representative outlier classes. Representative outlier classes with unusually low reliance on local shape include \textit{corn}, \textit{Appenzeller (cheese)}, \textit{bookstall}, \textit{spider’s web}, \textit{zebra}, \textit{guacamole}, and \textit{stone wall}. Classes with unusually high texture reliance include \textit{Chesapeake Bay retriever}, \textit{Crotalus cerastes}, \textit{stingray}, \textit{bath towel}, \textit{Alligator mississippiensis}, and \textit{bolete}. Outliers with stronger color reliance include \textit{sunglass}, \textit{tank suit}, \textit{ladle}, \textit{coffeepot}, \textit{chain}, and \textit{tam-tam}.

\subsection{CV Feature Reliance for Models trained from Scratch}

\begin{figure}[t!]
    \def\subfigwidth{.31}
    \centering

    \makebox[\linewidth][c]{%
        \begin{minipage}{\subfigwidth\linewidth}
            \centering\scriptsize\textbf{(a) Shape Suppression}
        \end{minipage}
        \hspace{0.1cm}
        \begin{minipage}{\subfigwidth\linewidth}
            \centering\scriptsize\textbf{(b) Texture Suppression}
        \end{minipage}
        \hspace{0.1cm}
        \begin{minipage}{\subfigwidth\linewidth}
            \centering\scriptsize\textbf{(c) Color Suppression}
        \end{minipage}
    }

    \vspace{2pt}

    \begin{subfigure}{\subfigwidth\linewidth}
        \includegraphics[width=\linewidth]{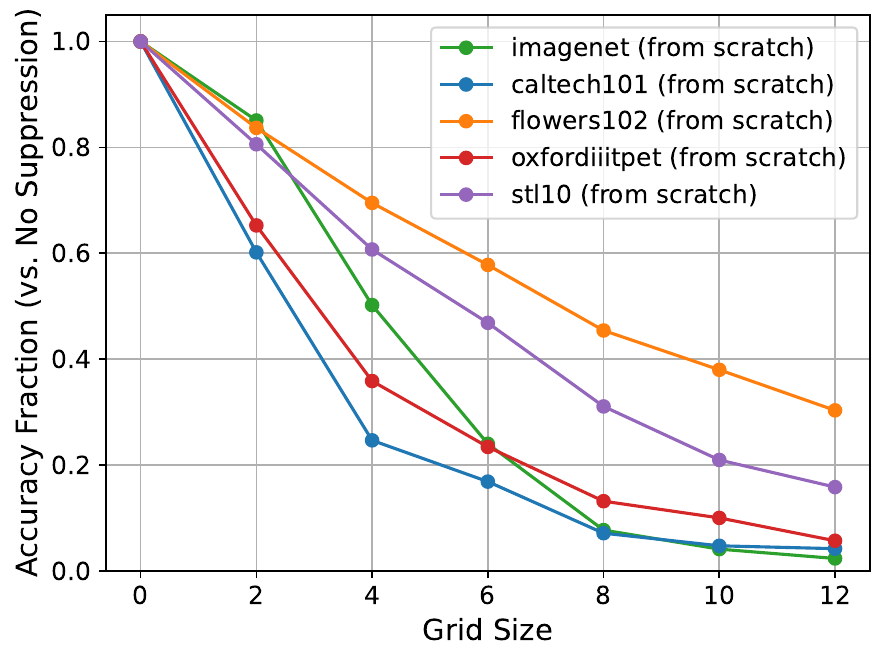}
        \label{subfigure:cv_from_scratch_patch_shuffle}
    \end{subfigure}
    \hspace{0.1cm}
    \begin{subfigure}{\subfigwidth\linewidth}
        \includegraphics[width=\linewidth]{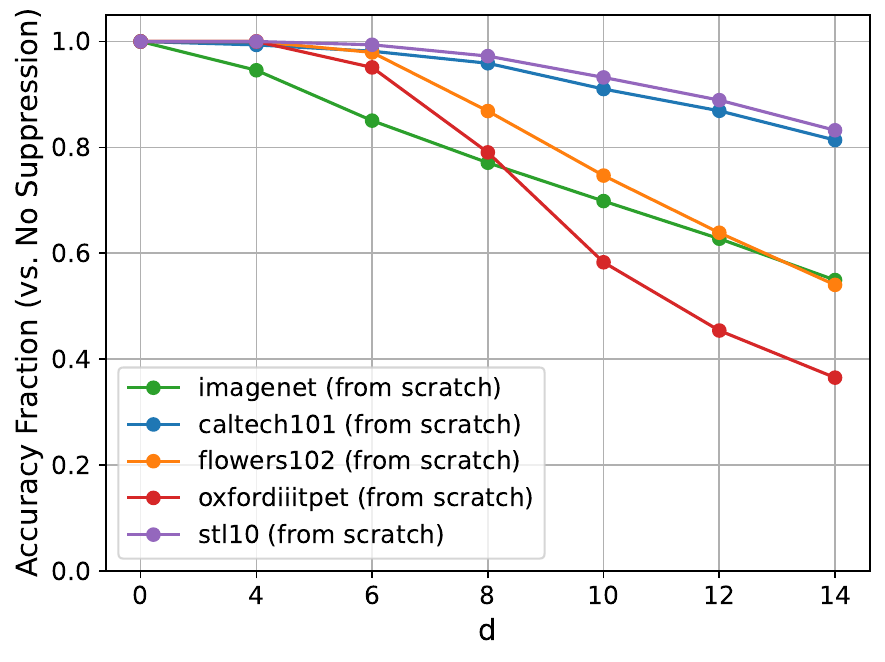}
        \label{subfigure:cv_from_scratch_bilateral}
    \end{subfigure}
    \hspace{0.1cm}
    \begin{subfigure}{\subfigwidth\linewidth}
        \includegraphics[width=\linewidth]{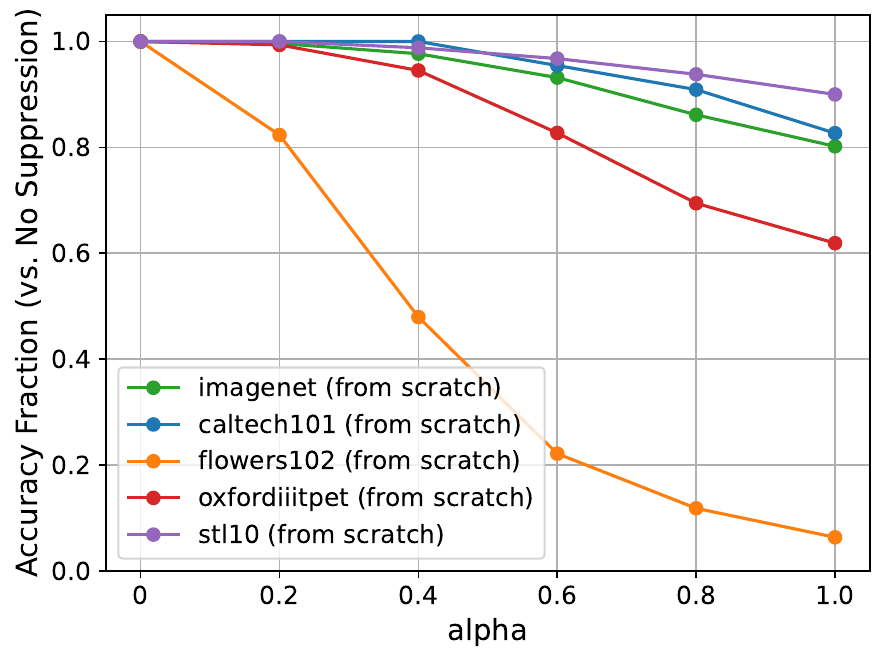}
        \label{subfigure:cv_from_scratch_graysscale}
    \end{subfigure}
    \vspace{-5pt}
    \caption{Feature suppression results on \gls{CV} datasets for a ResNet50 trained from scratch. \textbf{(a)} Shape suppression via Patch Shuffle. \textbf{(b)} Texture suppression via bilateral filtering. \textbf{(c)} Color suppression via grayscale.}
    \label{fig:cv_from_scratch}
\end{figure}

\Cref{fig:cv_from_scratch} shows suppression results for CV datasets using models trained from scratch. Compared to their fine-tuned counterparts, these models exhibit greater sensitivity to color and texture suppression, indicating higher reliance on these features. Interestingly, relative accuracy under light shape suppression (e.g., Patch Shuffle with small grid size) is lower than for fine-tuned models, while performance under strong shape suppression improves in several cases, indicating that these models increasingly use texture or color cues when shape information is heavily degraded. This effect is most pronounced for Flowers102, where the model retains comparatively high accuracy despite aggressive shape perturbation.

\subsection{MI and RS Feature Reliance for ImageNet-pretrained models}

\begin{figure}[t!]
    \def\subfigwidth{.31}
    \centering
  
    \makebox[\linewidth][c]{%
        \begin{minipage}{\subfigwidth\linewidth}
            \centering\scriptsize\textbf{(a) MI Shape Suppression}
        \end{minipage}
        \hspace{0.1cm}
        \begin{minipage}{\subfigwidth\linewidth}
            \centering\scriptsize\textbf{(b) MI Texture Suppression}
        \end{minipage}
        \hspace{0.1cm}
        \begin{minipage}{\subfigwidth\linewidth}
            \centering\scriptsize\textbf{(c) MI Color Suppression}
        \end{minipage}
    }

    \vspace{2pt}

    \begin{subfigure}{\subfigwidth\linewidth}
        \includegraphics[width=\linewidth]{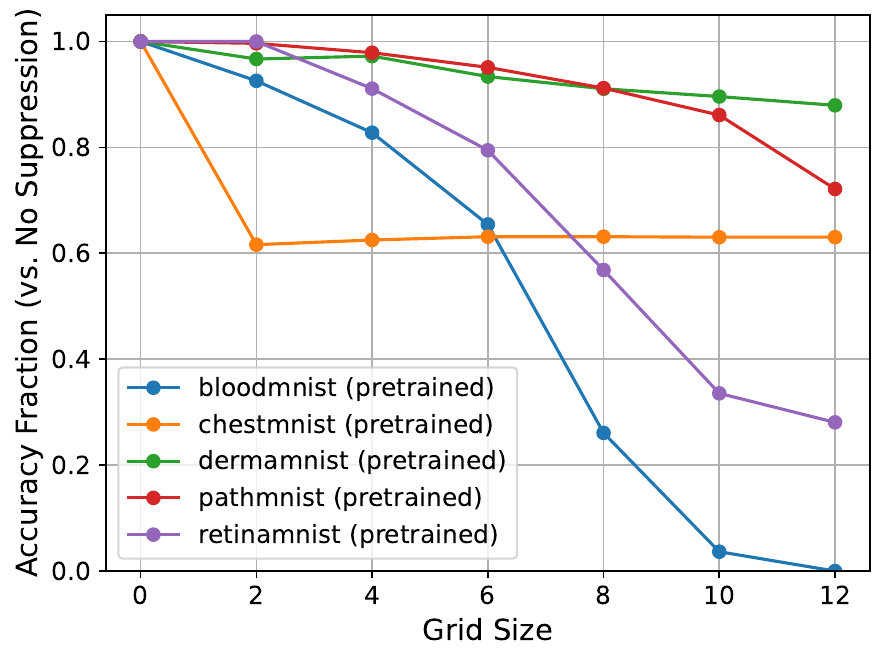}
        \label{subfigure:mi_pre_trained_shape}
    \end{subfigure}
    \hspace{0.1cm}
    \begin{subfigure}{\subfigwidth\linewidth}
        \includegraphics[width=\linewidth]{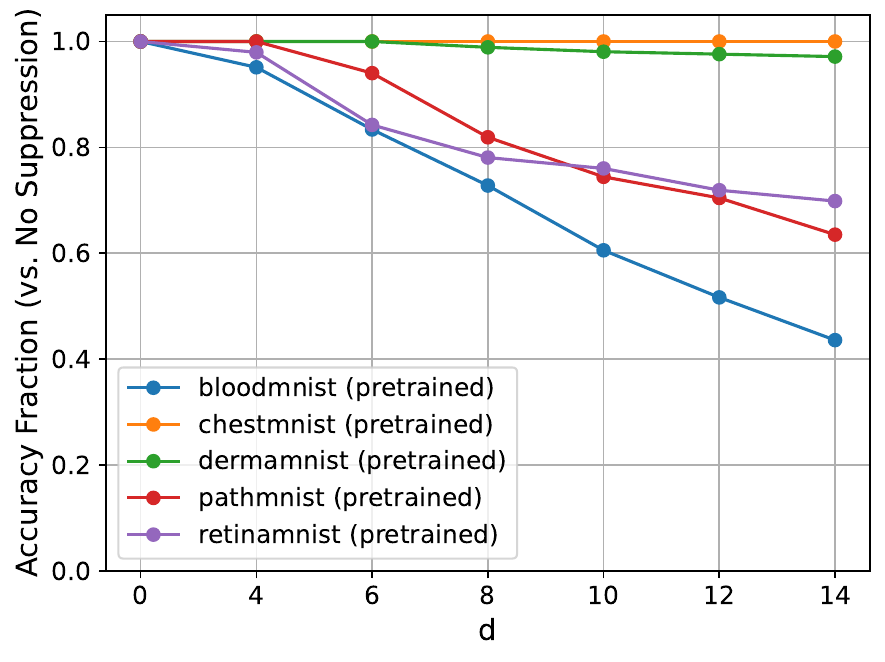}
        \label{subfigure:mi_pre_trained_patch_texture}
    \end{subfigure}
    \hspace{0.1cm}
    \begin{subfigure}{\subfigwidth\linewidth}
        \includegraphics[width=\linewidth]{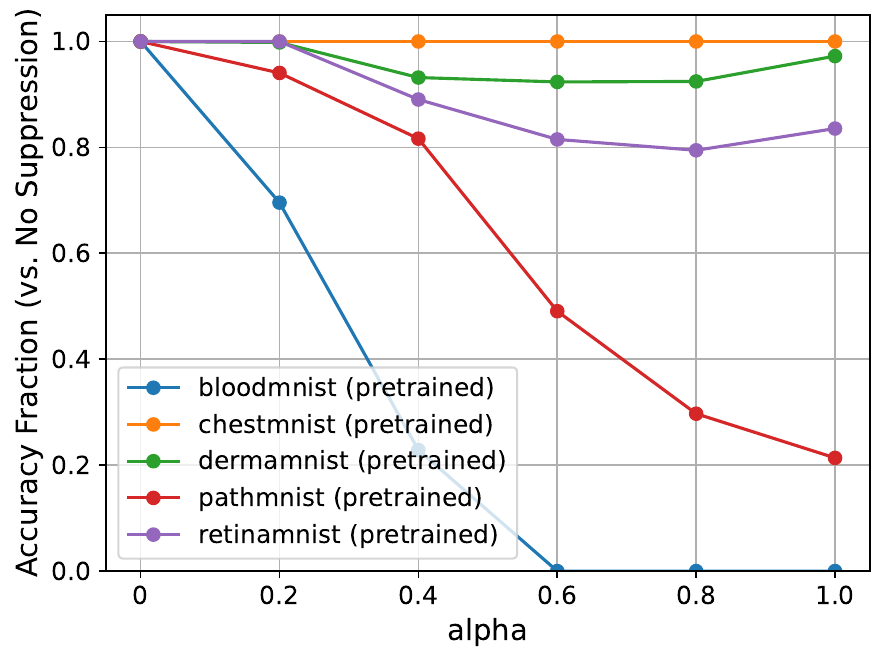}
        \label{subfigure:mi_pre_trained_color}
    \end{subfigure}

   \vspace{-10pt}

    \makebox[\linewidth][c]{%
        \begin{minipage}{\subfigwidth\linewidth}
            \centering\scriptsize\textbf{(d) RS Shape Suppression}
        \end{minipage}
        \hspace{0.1cm}
        \begin{minipage}{\subfigwidth\linewidth}
            \centering\scriptsize\textbf{(e) RS Texture Suppression}
        \end{minipage}
        \hspace{0.1cm}
        \begin{minipage}{\subfigwidth\linewidth}
            \centering\scriptsize\textbf{(f) RS Color Suppression}
        \end{minipage}
    }

    \vspace{2pt}

    \begin{subfigure}{\subfigwidth\linewidth}
        \includegraphics[width=\linewidth]{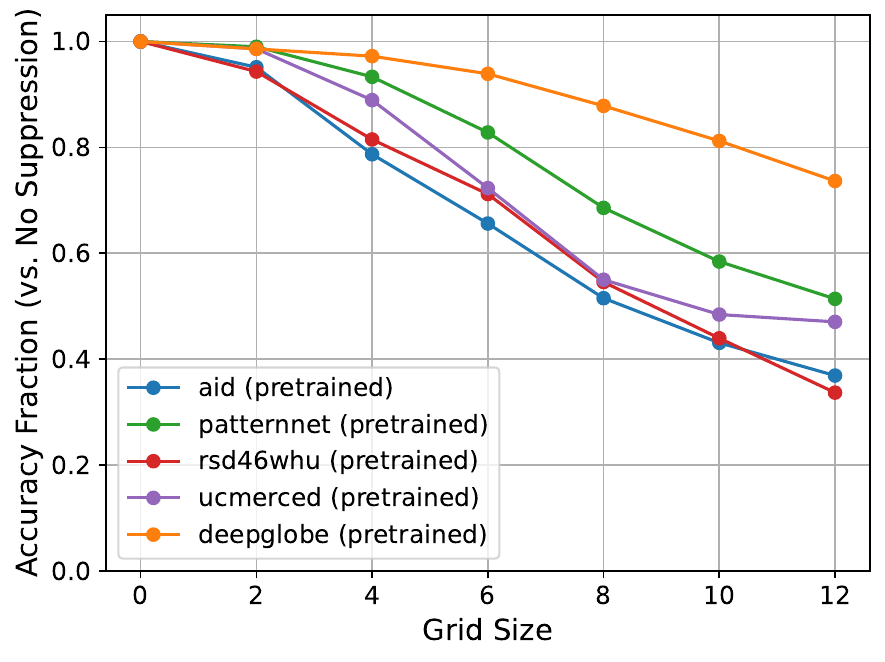}
        \label{subfigure:rs_pretrained_shape}
    \end{subfigure}
    \hspace{0.1cm}
    \begin{subfigure}{\subfigwidth\linewidth}
        \includegraphics[width=\linewidth]{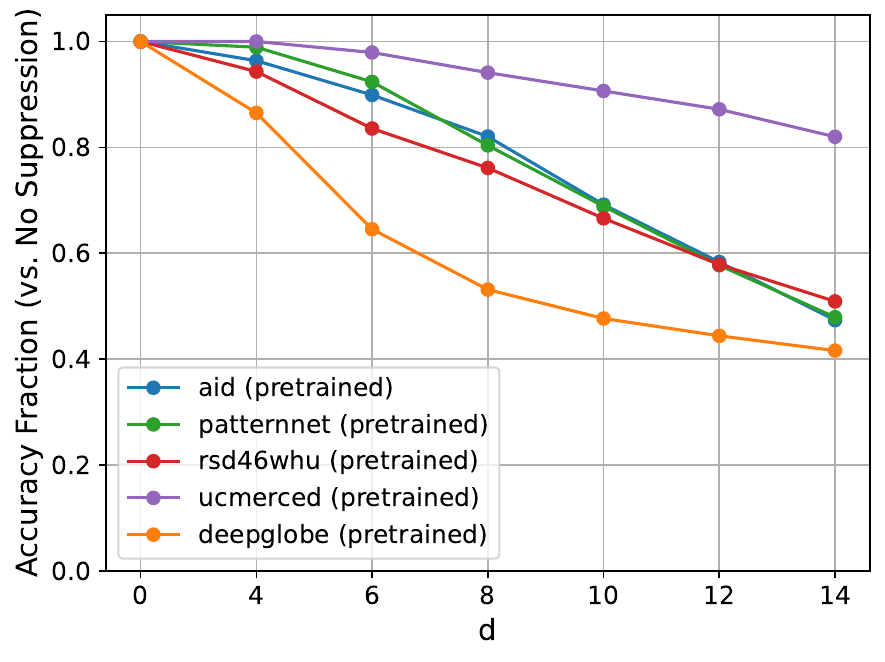}
        \label{subfigure:rs_patch_pretrained_texture}
    \end{subfigure}
    \hspace{0.1cm}
    \begin{subfigure}{\subfigwidth\linewidth}
        \includegraphics[width=\linewidth]{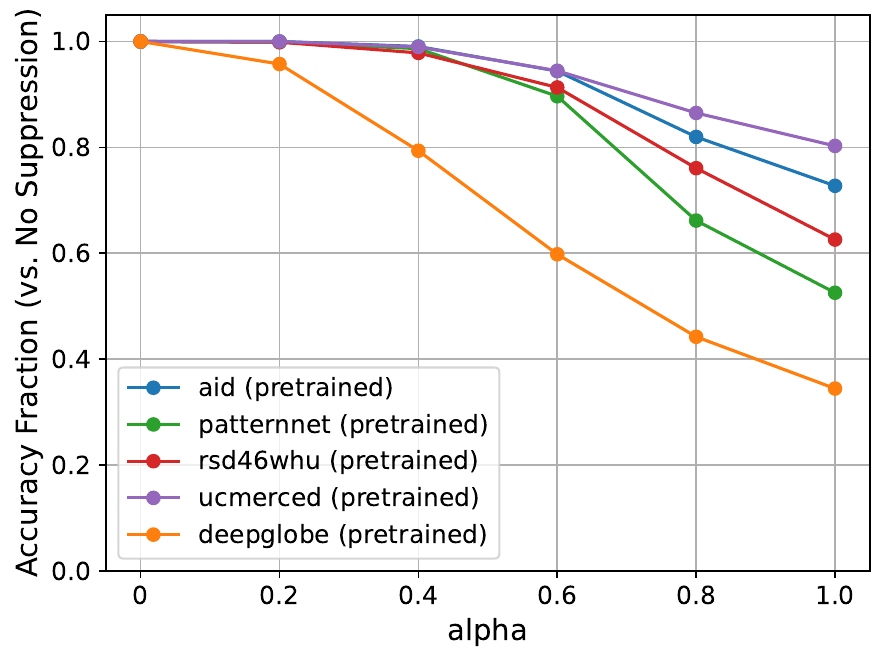}
        \label{subfigure:rs_pretrained_color}
    \end{subfigure}
    \vspace{-5pt}
    \caption{Feature suppression results across MI and RS domains when pre-trained on ImageNet. 
\textbf{Top row (a–c)}: ResNet50 pretrained on ImageNet and fine-tuned on MedMNIST-v2 (medical imaging).  
\textbf{Middle row (d–f)}: ResNet50 pretrained on ImageNet and fine-tuned on high-resolution \gls{RS} datasets. 
Columns correspond to: \textbf{(a, d)} shape suppression (Patch Shuffle), \textbf{(b, e)} texture suppression (Bilateral Filter), and \textbf{(c, f)} color suppression (Grayscale).}
    \label{fig:suppression_grid_mi_rs_finetune}
\end{figure}

\Cref{fig:suppression_grid_mi_rs_finetune} reports suppression results for MI and RS datasets using ImageNet-pretrained ResNet50 backbones fine-tuned on the respective training sets. The experimental setup and suppression protocols mirror those of \Cref{sec:experiment2} in the main paper. Across both domains, the overall reliance profiles remain broadly consistent with the models trained from scratch. However, pretraining introduces systematic shifts. In RS, models exhibit slightly stronger shape reliance and reduced sensitivity to texture and color suppression, in some cases up to 7\% greater degradation under shape suppression and up to 20\% less degradation under texture or color suppression. This aligns with the trends observed in CV, where ImageNet pretraining enhances robustness to non-shape perturbations. In MI, the effects are more heterogeneous. For BloodMNIST, pretraining increases sensitivity to shape suppression, particularly under strong perturbations (e.g., Patch Shuffle with grid size 8), while for PathMNIST it amplifies the impact of texture suppression.

\subsection{Alternative Transformations for Domain-specific Feature Reliance}
\label{sec:experimental_details}

\begin{figure}[t!]
    \def\subfigwidth{.31}
    \centering

    \makebox[\linewidth][c]{%
        \begin{minipage}{\subfigwidth\linewidth}
            \centering\scriptsize\textbf{(a) CV Shape Suppression}
        \end{minipage}
        \hspace{0.1cm}
        \begin{minipage}{\subfigwidth\linewidth}
            \centering\scriptsize\textbf{(b) CV Texture Suppression}
        \end{minipage}
        \hspace{0.1cm}
        \begin{minipage}{\subfigwidth\linewidth}
            \centering\scriptsize\textbf{(c) CV Color Suppression}
        \end{minipage}
    }

    \vspace{2pt}

    \begin{subfigure}{\subfigwidth\linewidth}
        \includegraphics[width=\linewidth]{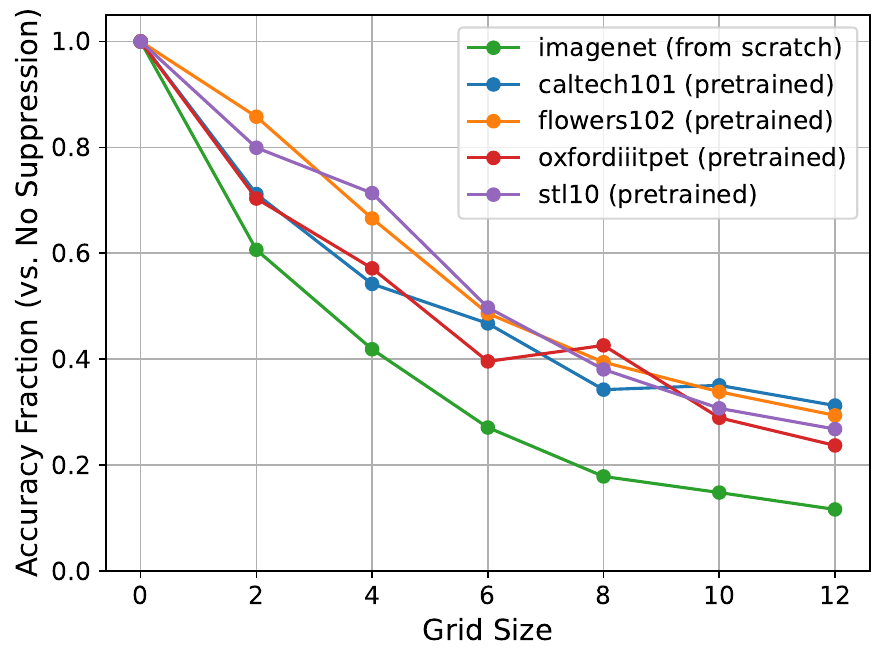}
        \label{subfigure:cv_pretrained_shape_pr}
    \end{subfigure}
    \hspace{0.1cm}
    \begin{subfigure}{\subfigwidth\linewidth}
        \includegraphics[width=\linewidth]{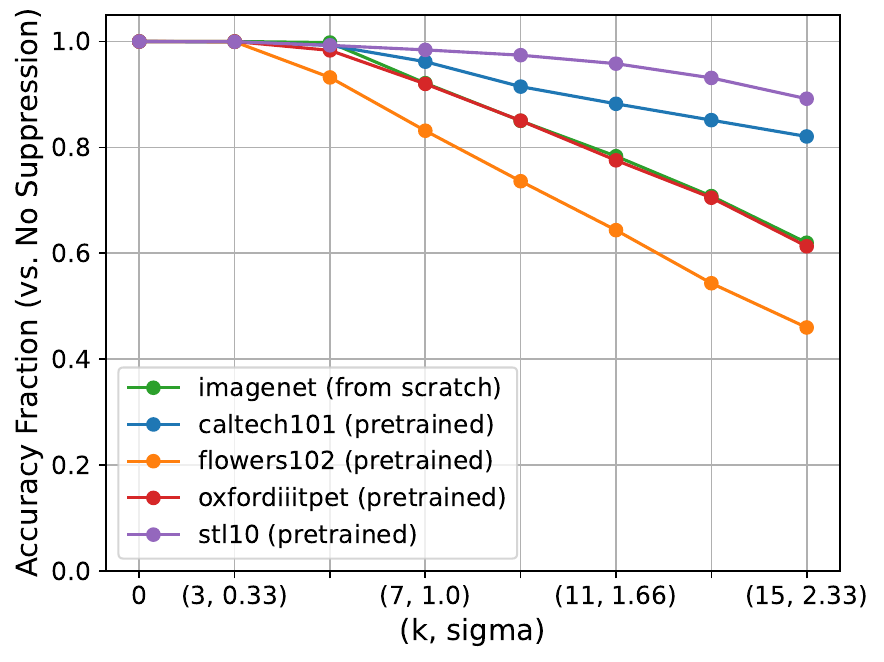}
        \label{subfigure:cv_pretrained_patch_texture_gb}
    \end{subfigure}
    \hspace{0.1cm}
    \begin{subfigure}{\subfigwidth\linewidth}
        \includegraphics[width=\linewidth]{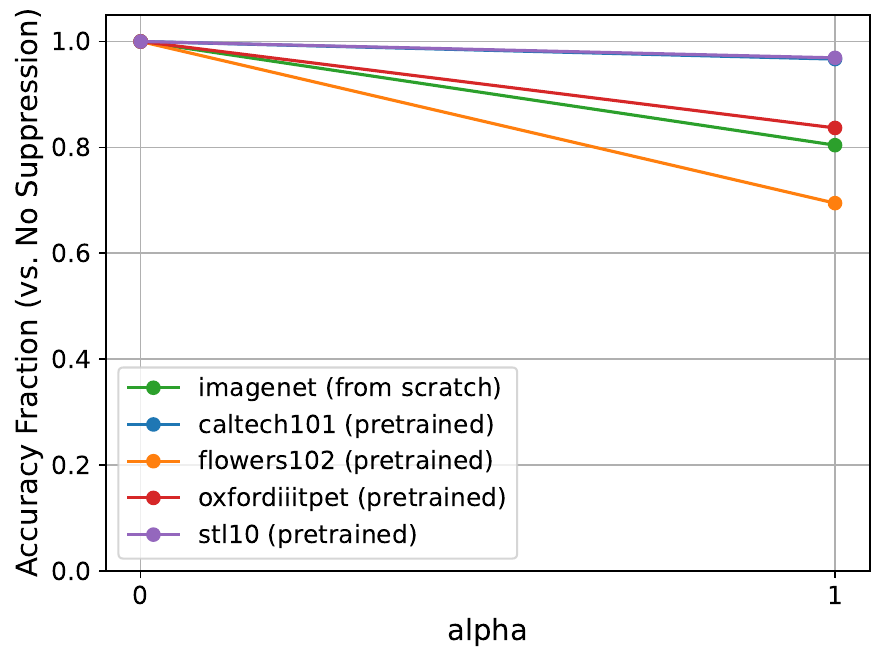}
        \label{subfigure:cv_pretrained_color_cs}
    \end{subfigure}

  \vspace{-10pt}
  
    \makebox[\linewidth][c]{%
        \begin{minipage}{\subfigwidth\linewidth}
            \centering\scriptsize\textbf{(d) MI Shape Suppression}
        \end{minipage}
        \hspace{0.1cm}
        \begin{minipage}{\subfigwidth\linewidth}
            \centering\scriptsize\textbf{(e) MI Texture Suppression}
        \end{minipage}
        \hspace{0.1cm}
        \begin{minipage}{\subfigwidth\linewidth}
            \centering\scriptsize\textbf{(f) MI Color Suppression}
        \end{minipage}
    }

    \vspace{2pt}

    \begin{subfigure}{\subfigwidth\linewidth}
        \includegraphics[width=\linewidth]{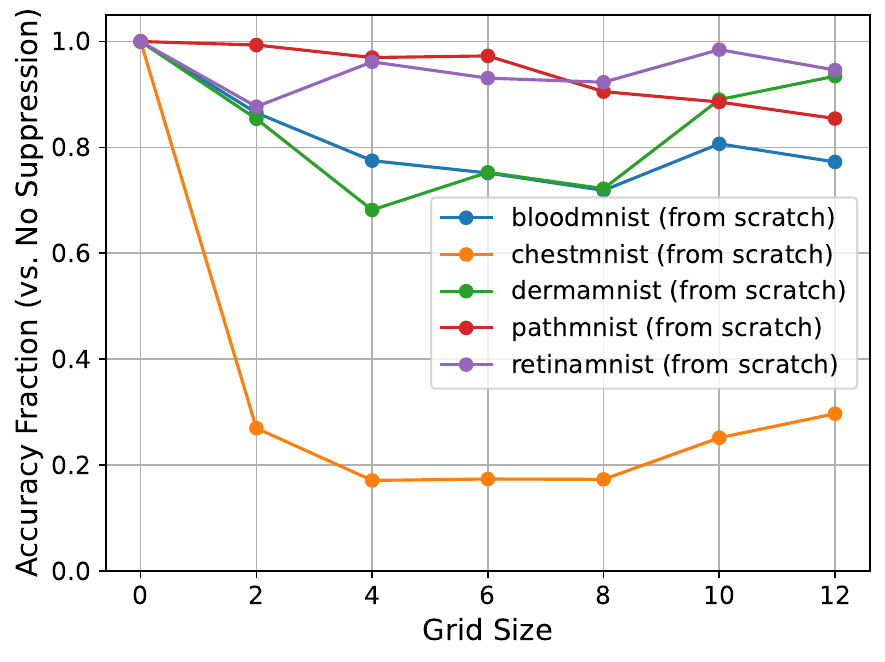}
        \label{subfigure:mi_shape_pr}
    \end{subfigure}
    \hspace{0.1cm}
    \begin{subfigure}{\subfigwidth\linewidth}
        \includegraphics[width=\linewidth]{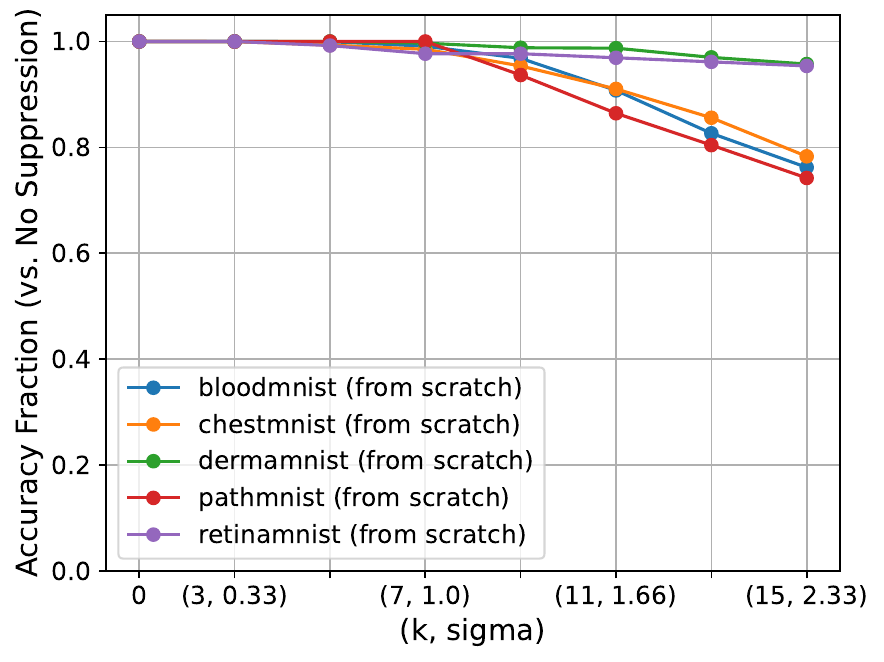}
        \label{subfigure:mi_patch_texture_gb}
    \end{subfigure}
    \hspace{0.1cm}
    \begin{subfigure}{\subfigwidth\linewidth}
        \includegraphics[width=\linewidth]{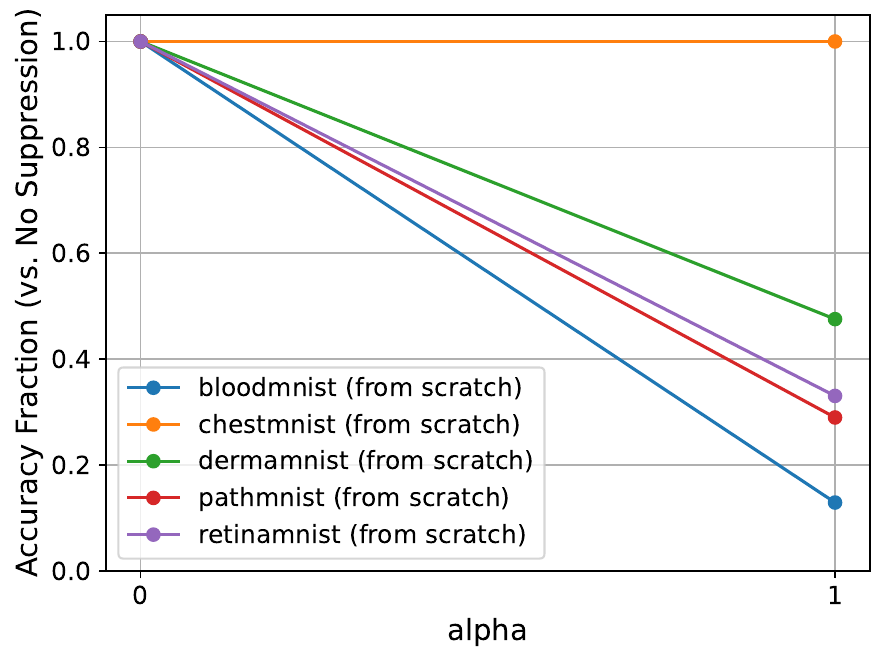}
        \label{subfigure:mi_color_cs}
    \end{subfigure}

   \vspace{-10pt}

    \makebox[\linewidth][c]{%
        \begin{minipage}{\subfigwidth\linewidth}
            \centering\scriptsize\textbf{(g) RS Shape Suppression}
        \end{minipage}
        \hspace{0.1cm}
        \begin{minipage}{\subfigwidth\linewidth}
            \centering\scriptsize\textbf{(h) RS Texture Suppression}
        \end{minipage}
        \hspace{0.1cm}
        \begin{minipage}{\subfigwidth\linewidth}
            \centering\scriptsize\textbf{(i) RS Color Suppression}
        \end{minipage}
    }

    \vspace{2pt}

    \begin{subfigure}{\subfigwidth\linewidth}
        \includegraphics[width=\linewidth]{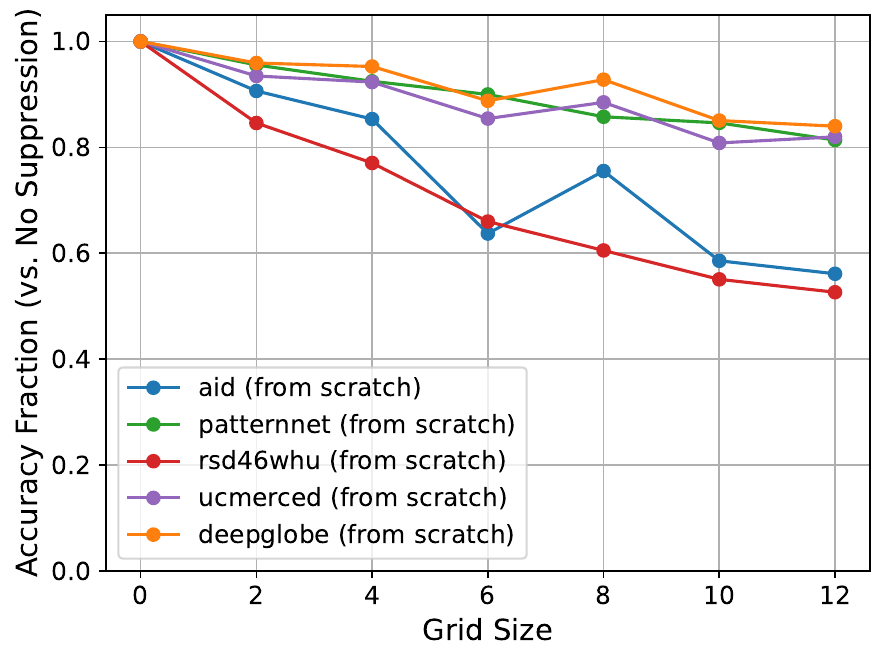}
        \label{subfigure:rs_shape_pr}
    \end{subfigure}
    \hspace{0.1cm}
    \begin{subfigure}{\subfigwidth\linewidth}
        \includegraphics[width=\linewidth]{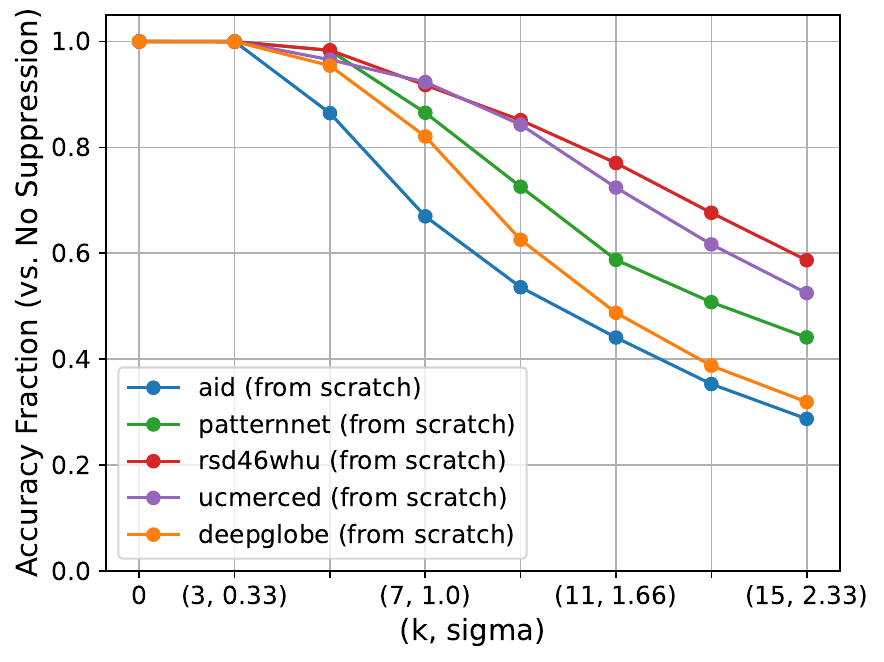}
        \label{subfigure:rs_patch_texture_gb}
    \end{subfigure}
    \hspace{0.1cm}
    \begin{subfigure}{\subfigwidth\linewidth}
        \includegraphics[width=\linewidth]{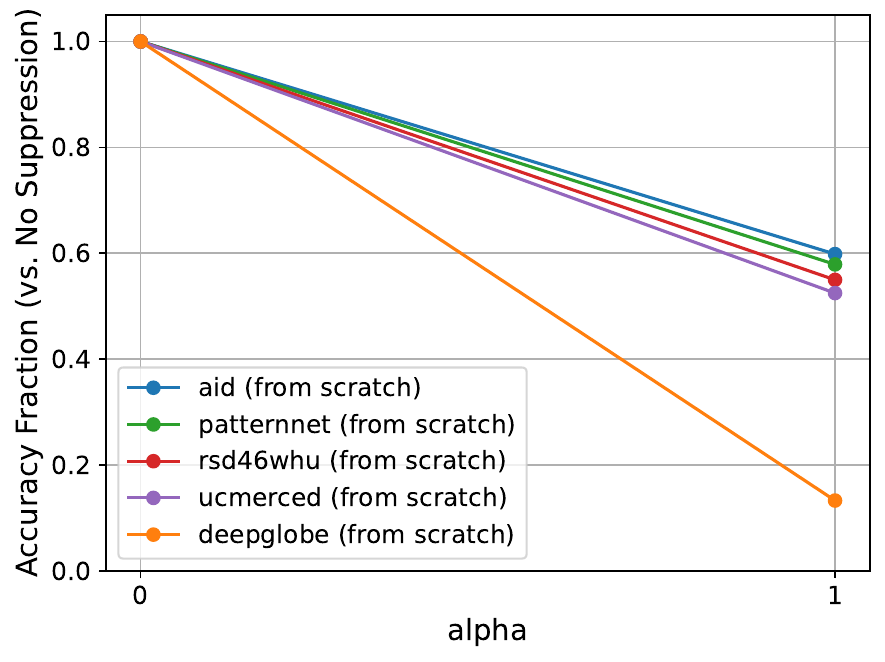}
        \label{subfigure:rs_color_cs}
    \end{subfigure}
    \vspace{-5pt}
    \caption{Feature suppression results across three domains. 
\textbf{Top row (a–c)}: ResNet50 pretrained on ImageNet and fine-tuned on \gls{CV} datasets. 
\textbf{Middle row (d–f)}: ResNet50 trained from scratch on MedMNIST-v2 (medical imaging). 
\textbf{Bottom row (g–i)}: ResNet50 trained from scratch on high-resolution \gls{RS} datasets. 
Columns correspond to: \textbf{(a, d, g)} shape suppression (Patch Rotation), \textbf{(b, e, h)} texture suppression (Gaussian Blur), and \textbf{(c, f, i)} color suppression (Channel Shuffle).}
    \label{fig:suppression_grid_alternative}
\end{figure}

\Cref{fig:suppression_grid_alternative} presents the results using alternative transformations targeting the same feature types: Patch Rotation for shape suppression, Gaussian Blur for texture suppression, and Channel Shuffle for color suppression. Overall, the results strongly correlate with those reported in the main paper using the primary suppression transformations. The only notable deviations occur for DermaMNIST and BloodMNIST, which show increased robustness to shape suppression via Patch Rotation, and for DermaMNIST, which exhibits a stronger decline under color suppression via Channel Shuffle. Across all other datasets and domains, the relative suppression effects remain consistent.

\subsection{Joint Suppression of Multiple Feature Types}\label{sec:joint_suppression}

The potential interdependence between features raises the question of whether the observed reliance patterns persist when multiple features are suppressed simultaneously. To probe this, we conduct an additional experiment on the \gls{CV} datasets when pretrained on ImageNet using joint suppression. We evaluate three combinations: texture and color suppression (preserving only shape), shape and color suppression (preserving only texture), and shape and texture suppression (preserving only color). For each case, we measure relative accuracy across increasing suppression strengths, analogous to the procedure in \Cref{sec:experiment2}.

\begin{figure}[t!]
    \def\subfigwidth{.31}
    \centering

    \makebox[\linewidth][c]{%
        \begin{minipage}{\subfigwidth\linewidth}
            \centering\scriptsize\textbf{(a) Texture and Color Suppression}
        \end{minipage}
        \hspace{0.1cm}
        \begin{minipage}{\subfigwidth\linewidth}
            \centering\scriptsize\textbf{(b) Shape and Texture Suppression}
        \end{minipage}
        \hspace{0.1cm}
        \begin{minipage}{\subfigwidth\linewidth}
            \centering\scriptsize\textbf{(c) Shape and Color Suppression}
        \end{minipage}
    }

    \vspace{2pt}

    \begin{subfigure}{\subfigwidth\linewidth}
        \includegraphics[width=\linewidth]{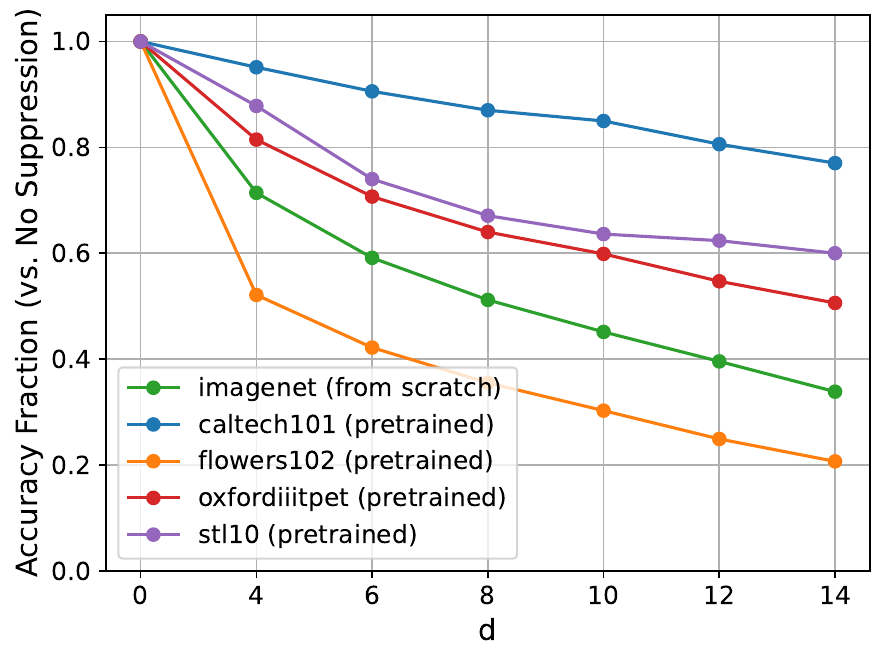}
        \label{subfigure:cv_pretrained_bilateral_grayscale}
    \end{subfigure}
    \hspace{0.1cm}
    \begin{subfigure}{\subfigwidth\linewidth}
        \includegraphics[width=\linewidth]{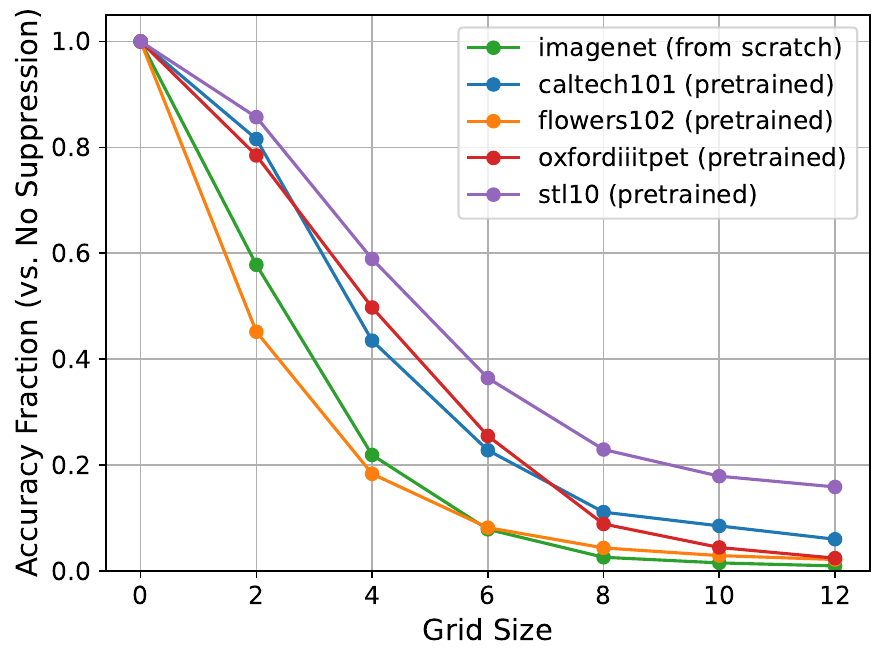}
        \label{subfigure:cv_pretrained_patch_shuffle_grayscale}
    \end{subfigure}
    \hspace{0.1cm}
    \begin{subfigure}{\subfigwidth\linewidth}
        \includegraphics[width=\linewidth]{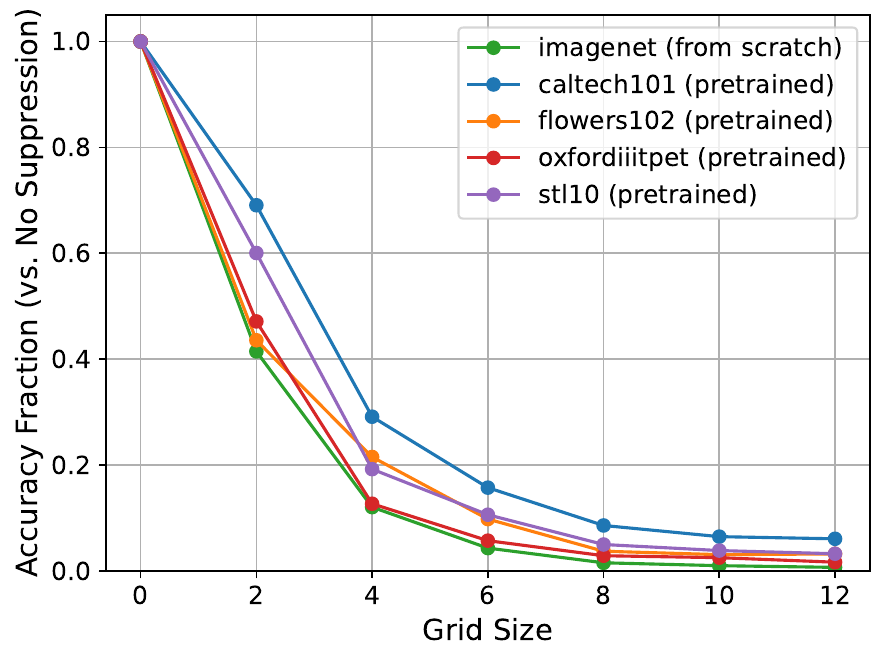}
        \label{subfigure:cv_pretrained_patch_shuffle_bilateral}
    \end{subfigure}
    \vspace{-5pt}
    \caption{Joint feature suppression results for a ResNet50 pretrained on ImageNet and fine-tuned on \gls{CV} datasets. \textbf{(a)} Texture suppression via bilateral filtering and color suppression via grayscale (only shape preserving). \textbf{(b)}  Shape suppression via Patch Shuffle and color suppression via grayscale (only texture preserving). \textbf{(c)} Shape suppression via Patch Shuffle and texture suppression via bilateral filtering (only color preserving).}
    \label{fig:cv_pretrained_two_feature_suppression}
\end{figure}

The results reinforce and extend the single-feature findings. Relative accuracy is highest when only shape is preserved, lower when only texture remains, and almost lost when only color is available (see \Cref{fig:cv_pretrained_two_feature_suppression}). These outcomes suggest that the relative importance of features remains stable even under combined suppression. 

\section{Control Experiment for Block-Edge Artifacts in Patch Shuffle}

\begin{figure*}[h]
    \def\subfigwidth{.29}
    \centering
    \begin{subfigure}{\subfigwidth\linewidth}
        {\includegraphics[width=\linewidth]{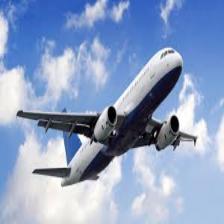}}
        {\caption{}\label{subfigure:block_structure_original}}
    \end{subfigure}
    \hfill
    \begin{subfigure}{\subfigwidth\linewidth}
        {\includegraphics[width=\linewidth]{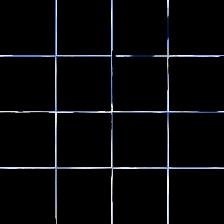}}
        {\caption{}\label{subfigure:only_overlay}}
    \end{subfigure}
    \hfill
    \begin{subfigure}{\subfigwidth\linewidth}
        {\includegraphics[width=\linewidth]{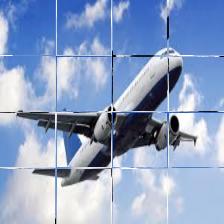}}
        {\caption{}\label{subfigure:block_structure_full_image}}
    \end{subfigure}
    \caption{Example of superimposition of grid-structure for control experiment for block-edge artifacts. \textbf{(a)} Original image. \textbf{(b)} Extracted grid structure from the shuffled image. \textbf{(c)} Image with grid structure as overlay.
    }
    \label{fig:block_structure_example}
    \vspace{-10pt}
\end{figure*}

Patch Shuffle simultaneously disrupts local spatial continuity and introduces artificial block-edge structures. To examine whether the observed performance degradation could be attributed primarily to block artifacts, we design a control condition that isolates the grid structure from the shuffling operation. For grid sizes of 2, 4, and 8, we generate an overlay variant of Patch Shuffle as follows. We first apply Patch Shuffle to an image, then extract the 2-pixel-wide block boundaries (1 pixel on either side). These boundaries are superimposed onto the original unshuffled image, preserving the global and local content while mimicking the block structure characteristic of Patch Shuffle. This procedure introduces visible grid lines without altering the patch arrangement. An example of the procedure is presented in \Cref{fig:block_structure_example}.

It is important to note that even the overlay condition introduces minor shape discontinuities, since the superimposed boundaries can slightly interrupt edge continuity. Consequently, the overlay condition still reflects two effects: (i) the presence of block edges and (ii) minor local shape disruption. Nevertheless, the stronger performance degradation observed under full Patch Shuffle compared to the overlay variant indicates (see \Cref{tab:block_structure_experiment}) that block artifacts alone do not explain the results. Instead, the principal effect arises from the combined disruption of local spatial structure.

\begin{table}[h]
\centering
\caption{Control experiment comparing Patch Shuffle with the grid-overlay variant, which mimics block structures without altering patch arrangement. Results are reported for different models and grid sizes.}
\begin{tabular}{lccc}
\toprule
\textbf{Model} & \textbf{Grid Size} & \textbf{Overlay} & \textbf{Patch Shuffle} \\
\midrule
ResNet50-standard & 2 & 0.950 & 0.921 \\
                  & 4 & 0.809 & 0.548 \\
                  & 8 & 0.540 & 0.069 \\
\midrule
ResNet50-sota     & 2 & 0.983 & 0.980 \\
                  & 4 & 0.870 & 0.837 \\
                  & 8 & 0.554 & 0.344 \\
\midrule
ConvNeXtV2        & 2 & 0.991 & 0.983 \\
                  & 4 & 0.957 & 0.859 \\
                  & 8 & 0.911 & 0.347 \\
\bottomrule
\end{tabular}
\label{tab:block_structure_experiment}
\end{table}

\section{Implementation Details}
\label{sec:implementation_details}

\subsection{Timm Pretraining Hyperparameter}
\label{sec:timm_config}

The following tables provide training hyperparameters for all evaluated CNNs, transformer-based models, and hybrid architectures used in Experiment I. All models, except ResNet50-standard, were obtained as pretrained checkpoints from the \texttt{timm} library and evaluated without further fine-tuning. \gls{CNNs} are listed in \Cref{tab:training_configs_timm_cnns}, transformer and hybrid models in \Cref{tab:training_configs_timm_transformers}, and additional pretraining hyperparameters, if applicable, in \Cref{tab:pretraining_settings_timm}.

We compiled the hyperparameter settings from a combination of official papers, GitHub repositories, HuggingFace model cards, and the \texttt{timm} source code. While we aim to be as faithful as possible, no centralized specification of all training details exists within \texttt{timm}. Accordingly, some entries (e.g., MixUp and CutMix) are marked as \emph{yes} to indicate usage without a reported $\alpha$ value.

\begin{table*}[t]
\centering
\scriptsize
\caption{Training hyperparameters for evaluated CNNs obtained from the timm library.}
\label{tab:training_configs_timm_cnns}
\renewcommand{\arraystretch}{1.1}
\setlength{\tabcolsep}{3pt}
\begin{tabular}{p{2.5cm}p{1.6cm}p{1.6cm}p{1.6cm}p{1.6cm}p{1.6cm}p{1.6cm}}
\toprule
\textbf{Category} & \makecell[l]{\textbf{ResNet50} \\ \textbf{-sota}} & \makecell[l]{\textbf{ConvNeXt} \\ \textbf{Tiny}} & \makecell[l]{\textbf{ConvNeXt-} \\ \textbf{V2-Tiny}} & \makecell[l]{\textbf{EfficientNet} \\ \textbf{-B5}} & \makecell[l]{\textbf{EfficientNet-} \\ \textbf{V2-RW-T}} & \makecell[l]{\textbf{MobileNet} \\ \textbf{V3-Large}} \\
\midrule
Pretraining & – & – & \makecell[l]{IN-22k \\ (FC-MAE)} & \makecell[l]{JFT-300M \\ (Noisy Student)} & – & – \\
Input Resolution & 224×224 & 224×224 & 224×224 & 456×456 & 224×224 & 224×224 \\
Epochs & 600 & 300 & 300 & 350 & 600 & 600 \\
Batch Size & 2048 & 4096 & 1024 & 2048 & 2048 & 2048 \\
Optimizer & LAMB & AdamW & AdamW & RMSProp & RMSProp & RMSProp \\
Decay / $\beta_2$ & – & 0.999 & 0.999 & 0.999 & 0.9 & 0.9 \\
Momentum / $\beta_1$ & – & 0.9 & 0.9 & 0.9 & 0.9 & 0.9 \\
Base LR & 5e-3 & 4e-3 & 8e-4 & 0.256 & 0.18 & 0.18 \\
LR Schedule & Cosine & Cosine & Cosine & RMSProp decay & Step exp decay & Step exp decay \\
Decay Rate & – & – & – & – & 0.988 & 0.988 \\
Warmup Epochs & 5 & 20 & 40 & 5 & 5 & 5 \\
Warmup Schedule & Linear & Linear & Linear & – & – & – \\
Label Smoothing & 0.1 & 0.1 & 0.1 & 0.1 & 0.1 & 0.1 \\
RandAugment & (7, 0.5) & (9, 0.5) & (9, 0.5) & – & (8, 2, 1.0) & (8, 2, 1.0) \\
AutoAugment & – & – & – & yes & – & – \\
Mixup $\alpha$ & 0.2 & 0.8 & 0.8 & 0.2 & 0.2 & 0.2 \\
CutMix $\alpha$ & 1.0 & 1.0 & 1.0 & – & – & – \\
Rand. Erasing $p$ & 0.25 & 0.25 & – & – & 0.35 & 0.35 \\
Dropout & – & – & – & 0.2 & 0.2 & 0.2 \\
Stoch. Depth & 0.05 & 0.1 & – & – & 0.1 & 0.1 \\
Drop Path & – & – & 0.2 & 0.2 & – & – \\
Layer-wise LR Decay & – & 0.65 & 0.9 & – & – & – \\
Weight Init & – & Trunc. Normal & – & Trunc. Normal & – & – \\
Layer Scale Init & – & 1e-6 & – & – & – & – \\
Head Init Scale & – & – & 0.001 & – & – & – \\
EMA & – & 0.9999 & 0.9999 & – & 0.9999 & 0.9999 \\
Loss Function & BCE & CE & CE & CE & CE & CE \\
Mixed Precision & Yes & Yes & Yes & Yes & Yes & Yes \\
Top-1 Accuracy & 80.4\% & 82.1\% & $\sim$83–84\% & 83.6\% & 79.4\% & 75.2\% \\
\bottomrule
\end{tabular}
\end{table*}

\begin{table*}[t]
\centering
\scriptsize
\caption{Training hyperparameters for evaluated transformers and hybrid architectures obtained from the timm library. MixUp and CutMix are marked as \textit{yes} where applied but unspecified.}
\begin{tabular}{p{2.5cm}p{1.6cm}p{1.6cm}p{1.6cm}p{1.6cm}p{1.7cm}}
\toprule
\textbf{Category} & \makecell[l]{\textbf{ConvMixer-}\\\textbf{768/32}} & \textbf{ViT-B/16} & \textbf{DeiT-B} & \textbf{Swin-BP4-W7} & \textbf{CLIP ViT-B/16} \\
\midrule
Pretraining & - & ImageNet-21k & - & - & \makecell[l]{400M image-\\text pairs} \\
Input Resolution & 224×224 & 224×224 & 224×224 & 224×224 & 224×224 \\
Epochs & 150 & 300 & 300 & 300 & 32 \\
Batch Size & 64 & 4096 & 1024 & 1024 & 32768 \\
Optimizer & AdamW & AdamW & AdamW & AdamW & AdamW \\
Decay / $\beta_2$ & 1e-3 ($\epsilon$) & 0.999 & 0.999 & 0.999 & 0.999 \\
Momentum / $\beta_1$ & 0.9 & 0.9 & 0.9 & 0.9 & 0.9 \\
Base Learning Rate & 0.01 & 0.003 & 0.0005 × bs/512 & 0.001 & 5e-4 \\
LR Schedule & OneCycle & Cosine decay & Cosine decay & Cosine decay & Cosine decay \\
Decay Rate & – & – & – & – & – \\
Warmup Epochs & 0 & 5 & 5 & 20 & – (2000 steps) \\
Warmup Schedule & – & – & – & Linear & – \\
Label Smoothing & – & 0.1 & 0.1 & 0.1 & – \\
RandAugment & (9, 0.5) & (9, 0.5) & (9, 0.5) & (9, 0.5) & (9, 0.5) \\
AutoAugment & – & – & – & – & – \\
Mixup & 0.5 & 0.8 & 0.8 & yes & yes \\
CutMix & 0.5 & 1.0 & 1.0 & yes & yes \\
Random Erasing $p$ & 0.25 & 0.25 & 0.25 & 0.25 & yes \\
Dropout & – & 0.1 & 0.1 & – & – \\
Stochastic Depth & – & 0.1 & 0.1 & 0.2 & 0.2 \\
Drop Path & – & – & – & – & – \\
Layer-wise LR Decay & – & – & – & – & – \\
Weight Initialization & – & – & – & – & – \\
Layer Scale Init & – & – & – & – & – \\
Head Init Scale & – & – & – & – & – \\
EMA & – & – & – & – & – \\
Loss Function & CE & CE & CE & CE & \makecell[l]{InfoNCE \\ (Contrastive)} \\
Mixed Precision & Yes & Yes & Yes & Yes & Yes \\
Top-1 Accuracy & $\sim$82.0\% & $\sim$83.0\% & $\sim$83.0\% & $\sim$83.0\% (est.) & $\sim$78.0\% (0-shot) \\
\bottomrule
\end{tabular}
\label{tab:training_configs_timm_transformers}
\end{table*}

\begin{table}[t]
\centering
\scriptsize
\caption{Pretraining hyperparameter for models later fine-tuned on ImageNet-1K.}
\vspace{5pt}
\label{tab:pretraining_settings_timm}
\renewcommand{\arraystretch}{1.1}
\setlength{\tabcolsep}{4pt}
\begin{tabular}{lccc}
\toprule
\textbf{Model} & \textbf{Dataset} & \textbf{Epochs} & \textbf{Pretraining Setup} \\
\midrule
ConvNeXtV2-Tiny & ImageNet-22k & 800–1600 & \makecell[l]{AdamW, LR 1.5e-4, weight decay 0.05, \\ $\beta_1$=0.9, $\beta_2$=0.95, Cosine decay, \\ RandomResizedCrop, warmup 40 epochs.} \\
EfficientNet-B5 & JFT-300M & 800 & \makecell[l]{Noisy Student self-training with teacher on IN-1k, \\ student on JFT-300M with RandAugment, Mixup, Dropout.} \\
\bottomrule
\end{tabular}
\end{table}

\subsection{Training Hyperparameter ResNet50 from Scratch}
\label{sec:resnet50_from_scratch}

\Cref{tab:resnet50_training_configs} summarizes the training hyperparameters for all ResNet50 models (denoted as ResNet50-sota in \Cref{sec:human_vs_cnns}) trained from scratch that we used in our experiments. All models were trained or fine-tuned with minimal regularization to ensure comparability across datasets and domains. Data augmentation was limited to Random Resized Crop (RRC) and Horizontal Flip (HF). RRC was applied with a scale range of (0.3, 1.0), default aspect ratio, and probability 1.0. Through RRC all images were resized to $224 \times 224$. Horizontal flipping was applied with a probability of 0.5. For all computer vision datasets, ImageNet normalization statistics were used. For remote sensing and medical imaging datasets, dataset-specific statistics were computed from the training set. All models were trained using the cross-entropy loss, except for ChestMNIST (binary classification) and DeepGlobe (multi-label classification), which used binary cross-entropy. When using cosine annealing with warm restarts as learning rate scheduler, we set \(T_0 = 10\) epochs, \(\eta_{\text{min}} = 1 \times 10^{-6}\), and \(T_{\text{mult}} = 2\), except for fine-tuning where \(T_{\text{mult}} = 1\). For each dataset, the checkpoint with the highest validation accuracy was selected for subsequent suppression-based evaluation.

\begin{table*}[t]
\centering
\scriptsize
\caption{Training hyperparameters for ResNet50 models across all domains and settings. RRC: RandomResizedCrop, HF: HorizontalFlip. All models were trained or fine-tuned using supervised learning with cross-entropy loss, except for ChestMNIST (binary classification)
and DeepGlobe (multi-label classification), which used binary cross-entropy.}
\label{tab:resnet50_training_configs}
\renewcommand{\arraystretch}{1.1}
\setlength{\tabcolsep}{4pt}
\begin{tabular}{l l l l l l l l l l}
\toprule
\textbf{Dataset} & \textbf{Pretraining} & \textbf{Epochs} & \makecell[l]{\textbf{Batch} \\ \textbf{Size}} & \textbf{Image Size} & \textbf{Optimizer} & \textbf{LR} & \makecell[l]{\textbf{Weight} \\ \textbf{Decay}} & \textbf{LR Schedule} & \textbf{Train Augment} \\
\midrule
\multicolumn{9}{l}{\textit{Computer Vision (From Scratch)}} \\
ImageNet     & -- & 100 & 256 & 224x224 & SGD     & 0.1     & 0.0001 & CosAnnealWR & RRC + HF \\
Flowers102   & -- & 300 & 64  & 224x224 & AdamW   & 0.001   & 0.01   & StepLR (4) & RRC + HF \\
STL10        & -- & 300 & 64  & 224x224 & AdamW   & 0.001   & 0.01   & StepLR (4) & RRC + HF \\
Caltech101   & -- & 300 & 64  & 224x224 & AdamW   & 0.001   & 0.01   & StepLR (4) & RRC + HF \\
OxfordIIITPet& -- & 300 & 64  & 224x224 & AdamW   & 0.001   & 0.01   & StepLR (4) & RRC + HF \\
\midrule
\multicolumn{9}{l}{\textit{Computer Vision (Pretrained)}} \\
Flowers102   & IN-1k & 100 & 64 & 224x224 & AdamW   & 1e-5    & 0.001  & CosAnnealWR & RRC + HF \\
STL10        & IN-1k & 100 & 64 & 224x224 & AdamW   & 1e-5    & 0.001  & CosAnnealWR & RRC + HF \\
OxfordIIITPet& IN-1k & 100 & 64 & 224x224 & AdamW   & 1e-5    & 0.001  & CosAnnealWR & RRC + HF \\
Caltech101   & IN-1k & 100 & 64 & 224x224 & AdamW   & 1e-5    & 0.001  & CosAnnealWR & RRC + HF \\
\midrule
\multicolumn{9}{l}{\textit{Medical Imaging}} \\
BloodMNIST   & -- & 50  & 64 & 224x224 & AdamW   & 0.001   & 1e-5   & StepLR (3)  & RRC + HF \\
ChestMNIST   & -- & 50  & 64 & 224x224 & AdamW   & 0.001   & 1e-5   & StepLR (3)  & RRC + HF \\
DermaMNIST   & -- & 50  & 64 & 224x224 & AdamW   & 0.001   & 1e-5   & StepLR (3)  & RRC + HF \\
PathMNIST    & -- & 50  & 64 & 224x224 & AdamW   & 0.001   & 1e-5   & StepLR (3)  & RRC + HF \\
RetinaMNIST  & -- & 50  & 64 & 224x224 & AdamW   & 0.001   & 1e-5   & StepLR (3)  & RRC + HF \\
\midrule
\multicolumn{9}{l}{\textit{Remote Sensing}} \\
AID          & -- & 80  & 64 & 600x600 & AdamW   & 0.0005  & 0.01   & CosAnnealWR & RRC + HF \\
DeepGlobe    & -- & 80  & 64 & 256x256 & AdamW   & 0.0005  & 0.01   & CosAnnealWR & RRC + HF \\
PatternNet   & -- & 80  & 64 & 256x256 & AdamW   & 0.0005  & 0.01   & CosAnnealWR & RRC + HF \\
RSD46-WHU    & -- & 80  & 64 & 256x256 & AdamW   & 0.0005  & 0.01   & CosAnnealWR & RRC + HF \\
UCMerced     & -- & 80  & 64 & 256x256 & AdamW   & 0.0005  & 0.01   & CosAnnealWR & RRC + HF \\
\bottomrule
\end{tabular}
\end{table*}

\subsection{Computation Resources}
\label{sec:computation_resources}

All experiments were conducted on an internal server equipped with 2× AMD EPYC 9554 64-core processors (256 threads), 6× NVIDIA H100 PCIe GPUs (each with 81 GB memory, CUDA 12.2), and 1.5 TiB of system RAM. The system runs Ubuntu 22.04 with Linux kernel 5.15 and NVIDIA driver version 535.183.01. 

Training times for ResNet50 models varied by dataset. Training on ImageNet took approximately 10 days on a single GPU. For smaller \gls{CV} datasets, training from scratch took 30 minutes for Flowers102, 120 minutes for STL-10 and Caltech101, and 90 minutes for Oxford-IIITPet. Fine-tuning on the same datasets required 10 minutes for Flowers102, 40 minutes for STL-10 and Caltech101, and 30 minutes for Oxford-IIITPet. In the \gls{MI} domain, training durations were 30 minutes for BloodMNIST, 3 hours for ChestMNIST and PathMNIST, 20 minutes for DermaMNIST, and 5 minutes for RetinaMNIST. Training on \gls{RS} datasets took 90 minutes for AID, DeepGlobe, PatternNet, and RSD46-WHU, and 15 minutes for UCMerced.

Evaluation time per model and dataset ranged between 1 and 10 minutes, depending on the suppression condition and dataset size.

\section{Ethics Statement and Risk Assessment}

This study involved a low-risk visual classification task conducted with adult participants. All participants were volunteers, fully informed about the purpose and procedures of the study, and provided written informed consent prior to participation. No vulnerable populations (e.g., children, patients, or individuals with impaired consent capacity) were involved. Participants were not exposed to any physical or emotional risks, high stress levels, or invasive procedures such as fMRI or TMS.

The study design, including all procedures for participant interaction and data handling, was reviewed through the standard ethics assessment protocol of our institution. The responsible ethics committee certified that the study complies with all relevant legal and institutional guidelines. Specifically, the ethics committee confirmed that:

\begin{itemize}
    \item the study involves no foreseeable risk of harm;
    \item participants are not drawn from vulnerable populations;
    \item data privacy is protected in accordance with applicable regulations;
    \item informed consent was obtained from all participants;
    \item the study team bears responsibility for the truthful completion of the ethics review questionnaire and the ethical integrity of the study.
\end{itemize}

As such, the study received approval from the institutional ethics committee, and no ethical concerns were identified.

\end{document}